\documentclass[11pt,a4paper]{article}
\usepackage{times,latexsym}
\usepackage{url}
\usepackage[utf8]{inputenc}
\usepackage[T1]{fontenc}
\usepackage[acceptedWithA]{style/tacl2021v1} 

\usepackage{xspace,mfirstuc,tabulary}

\newif\iftaclinstructions
\taclinstructionsfalse 
\iftaclinstructions
\renewcommand{\confidential}{}
\renewcommand{\anonsubtext}{(No author info supplied here, for consistency with
TACL-submission anonymization requirements)}
\newcommand{\instr}
\fi

\iftaclpubformat 

\else

\fi

\title{\iconresearchqa~\dataset: Evaluating Scholarly Question Answering at Scale\\ Across 75 Fields with Survey-Mined Questions and Rubrics}

\author{
  Li S. Yifei$^*$,
  Allen Chang$^*$,
  Chaitanya Malaviya,
  Mark Yatskar
  \ \\
  University of Pennsylvania
  \\
  \texttt{\{liyifei, cylumn\}@seas.upenn.edu}
  \vspace{1em}
  \\
    {\footnotesize\begin{tabular}{@{}ll@{}}
    \icondata~\,\textbf{Data:} & \href{https://huggingface.co/datasets/realliyifei/ResearchQA}{\texttt{huggingface.co/datasets/realliyifei/ResearchQA}} \\
    \iconcode~\,\textbf{Code:} & \href{https://github.com/realliyifei/ResearchQA}{\texttt{github.com/realliyifei/ResearchQA}} \\
    \iconwebsite~\,\textbf{Website:} & \href{https://cylumn.com/ResearchQA}{\texttt{cylumn.com/ResearchQA}}
    \end{tabular}
    }
}

\date{}

\usepackage{booktabs}
\usepackage{amsmath}
\usepackage{pifont}
\usepackage{makecell}
\usepackage{colortbl}
\usepackage{bm}
\usepackage[most,]{tcolorbox}
\usepackage{multirow}
\usepackage{lipsum}
\usepackage{soul}
\usepackage{colortbl}
\usepackage{float}
\usepackage{bm}
\usepackage{algorithm}
\usepackage{algorithmic}
\usepackage{tcolorbox}
\usepackage{amsfonts}
\usepackage{booktabs}
\usepackage{array}
\usepackage{amsmath}
\usepackage{enumitem}
\usepackage{changepage}
\usepackage{makecell}
\usepackage{placeins}

\usepackage[margin=1in]{geometry}
\definecolor{darkred}{rgb}{0.8, 0, 0}
\definecolor{shade}{rgb}{0.9, 0.9, 0.9}
\definecolor{linkcolor}{HTML}{350d9a}
\definecolor{lightgray}{gray}{0.95}
\definecolor{lavender}{HTML}{7D5DFD}

\usepackage{xspace}

\usepackage{listings}
\newcommand{\dataset}[0]{\textsc{ResearchQA}\xspace}
\newcommand{\query}[0]{$Q$\xspace}
\newcommand{\answer}[0]{$A$\xspace}
\newcommand{\queryinitial}[0]{$Q_\text{initial}$\xspace}
\newcommand{\mockanswer}[0]{$\hat{A}$\xspace}

\newcommand{\queryanswerinitialtuple}[0]{$(Q_\text{initial},\hat{A}_\text{initial})$\xspace}
\newcommand{\datecutoff}[0]{$D$\xspace}
\newcommand{\rubricparametric}[0]{$\mathbf{R}_P$\xspace}
\newcommand{\rubricsurvey}[0]{$\mathbf{R}_S$\xspace}
\newcommand{\rubrichybrid}[0]{$\mathbf{R}_H$\xspace}
\newcommand{\rubricgeneric}[0]
{$\mathbf{R}_G$\xspace}
\newcommand{\rubricitem}[0]{$R_i$\xspace}
\newcommand{\keyword}[1]{\emph{#1}}

\newcommand{\metricensemble}[0]{$\mathrm{J}_\text{ensemble}$\xspace}
\newcommand{\metricdirect}[0]{$\mathrm{J}_\text{direct}$\xspace}

\newcommand{\datatest}[0]{$\mathcal{D}_\text{test}$\xspace}
\newcommand{\dataval}[0]{$\mathcal{D}_\text{validation}$\xspace}

\newcommand{\icondata}[0]{\raisebox{-0.1\height}{\includegraphics[width=1em]{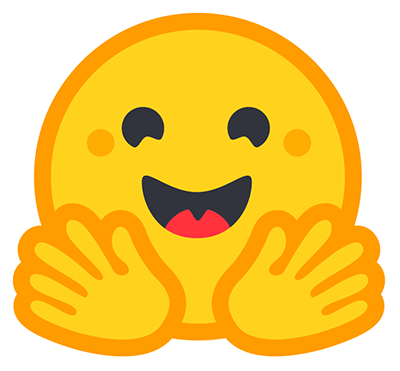}}}
\newcommand{\iconcode}[0]{\raisebox{-0.1\height}{\includegraphics[width=1em]{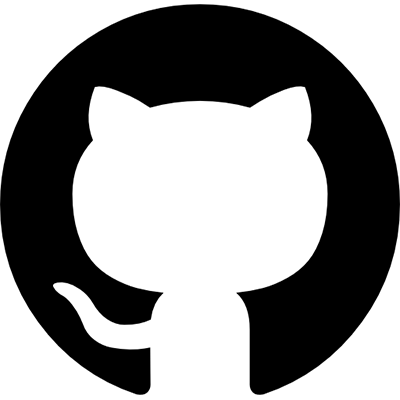}}}
\newcommand{\iconwebsite}[0]{\raisebox{-0.1\height}{\includegraphics[width=1em]{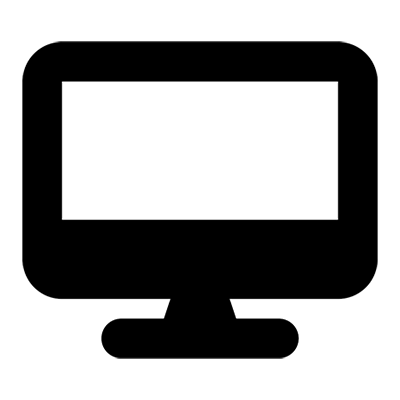}}}
\newcommand{\iconresearchqa}[0]{\smash{\raisebox{-.25\height}{\includegraphics[height=1.5em]{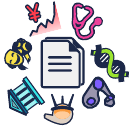}}}}
\newcommand{\iconengineering}[0]{\raisebox{-0.1\height}{\includegraphics[height=1em]{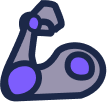}}}
\newcommand{\iconhealth}[0]{\raisebox{-0.1\height}{\includegraphics[height=1em]{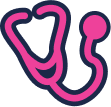}}}
\newcommand{\iconphysical}[0]{\raisebox{-0.1\height}{\includegraphics[height=1em]{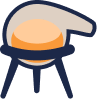}}}
\newcommand{\iconhumanities}[0]{\raisebox{-0.1\height}{\includegraphics[height=1em]{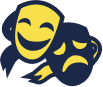}}}
\newcommand{\iconbusiness}[0]{\raisebox{-0.1\height}{\includegraphics[height=1em]{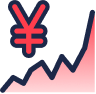}}}
\newcommand{\iconlife}[0]{\raisebox{-0.1\height}{\includegraphics[height=1em]{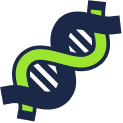}}}
\newcommand{\iconsocial}[0]{\raisebox{-0.1\height}{\includegraphics[height=1em]{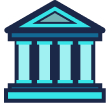}}}

\newcommand{\iconmeta}[0]{\raisebox{-0.1\height}{\includegraphics[height=0.8em]{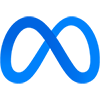}}}
\newcommand{\iconopenai}[0]{\raisebox{-0.1\height}{\includegraphics[height=0.8em]{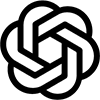}}}
\newcommand{\iconqwen}[0]{\raisebox{-0.1\height}{\includegraphics[height=0.8em]{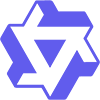}}}
\newcommand{\iconanthropic}[0]{\raisebox{-0.1\height}{\includegraphics[height=0.8em]{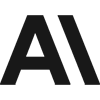}}}
\newcommand{\icongoogle}[0]{\raisebox{-0.1\height}{\includegraphics[height=0.8em]{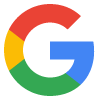}}}
\newcommand{\iconallenai}[0]{\raisebox{-0.1\height}{\includegraphics[height=0.8em]{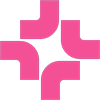}}}
\newcommand{\iconperplexity}[0]{\raisebox{-0.1\height}{\includegraphics[height=0.8em]{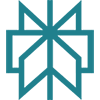}}}

\newcommand{\prometheustwomoe}[0]{\texttt{prometheus-2-8x7b}\xspace}

\newcommand{\gptfourone}[0]{\texttt{gpt-4.1}\xspace}
\newcommand{\gptfouronemini}[0]{\texttt{gpt-4.1-mini}\xspace}
\newcommand{\gptfouro}[0]{\texttt{gpt-4o}\xspace}

\newcommand{\geminitwofivepro}[0]{\texttt{gemini-2.5-pro}\xspace}
\newcommand{\geminitwofiveflash}[0]{\texttt{gemini-2.5-flash}\xspace}

\newcommand{\claudefoursonnet}[0]{\texttt{claude-4-sonnet}\xspace}

\newcommand{\llamathreethreeseventyb}[0]{\texttt{llama-3.3-70b}\xspace}

\newcommand{\qwenthreethirtytwob}[0]{\texttt{qwen-3-32b}\xspace}

\newcommand{\openscholareightb}[0]{\texttt{openscholar-8b}\xspace}
\newcommand{\openscholareightbagentic}[0]{\texttt{openscholar-8b+feedback}\xspace}

\newcommand{\openaiwebsearch}[0]{\texttt{gpt-4o-search-preview}\xspace}
\newcommand{\anthropicwebsearch}[0]{\texttt{claude-4-sonnet+ws}\xspace}
\newcommand{\geminiwebsearch}[0]{\texttt{gemini-2.5-pro+grounding}\xspace}

\newcommand{\sonar}[0]{\texttt{sonar}\xspace}
\newcommand{\sonarreasoning}[0]{\texttt{sonar-reasoning}\xspace}
\newcommand{\sonardeepresearch}[0]{\texttt{sonar-deep-research}\xspace}
\newcommand{\openaideepresearch}[0]{\texttt{o4-mini-deep-research}\xspace}

\newcommand{\startgraymidrule}[0]{
    \specialrule{0.06em}{\aboverulesep}{0pt} 
    \rowcolor{gray!20}
    \rule[-2.8\aboverulesep]{0pt}{\dimexpr\aboverulesep + \normalbaselineskip + \belowrulesep\relax}\null
}
\newcommand{\stopgraymidrule}[0]{
    \specialrule{0.05em}{0pt}{\belowrulesep} 
}
\newcommand{\graybottomrule}[0]{
    \specialrule{\heavyrulewidth}{0pt}{\belowbottomsep} 
}

\newcommand{\statqueries}[0]{21.4K\xspace}
\newcommand{\statqueriestruncated}[0]{21K\xspace}
\newcommand{\statfields}[0]{75\xspace}

\newcommand{\statannotators}[0]{31\xspace}

\newcommand{\statsurveysretrieved}[0]{615K\xspace}
\newcommand{\statsurveysdownloaded}[0]{134K\xspace}
\newcommand{\statsurveyslitreview}[0]{54K\xspace}

\newcommand{\statsectionstotal}[0]{886K\xspace}

\newcommand{\statsectionsmincitation}[0]{319K\xspace}

\usepackage[nameinlink]{cleveref}
\crefformat{section}{#2\S#1#3}
\Crefformat{section}{#2\S#1#3}

\usepackage{tabularx}
\usepackage{longtable}
\usepackage{enumitem}
\definecolor{systembg}{HTML}{daeced}
\definecolor{userbg}{HTML}{EEEEEE}
\definecolor{assistantbg}{HTML}{EEEEEE}
\colorlet{insert}{darkblue}

\newlength{\promptwidth}
\newlength{\promptenummargin}
\newtcolorbox{promptbox}[1]{%
    colback=#1, colframe=userbg,
    boxrule=0pt,
    arc=0pt,
    left=2pt,right=2pt,top=2pt,bottom=2pt,
    breakable,
    width=0.5\textwidth,
    before skip=0pt,
    after skip=0pt
}
\newcommand{\systemprompt}[1]{\noindent\begin{promptbox}{systembg}
        \textbf{System:}~#1%
    \end{promptbox}}
\newcommand{\userprompt}[1]{\noindent\begin{promptbox}{userbg}
        \textbf{User:}~#1%
    \end{promptbox}}

\newcommand{\prompttag}[1]{\hspace{0pt}\texttt{#1}}

\newcommand{\prompt}[2]{\subsection{#1}
    \vspace{-2.5ex}
    {\noindent\\
    \rule[0pt]{0.5\textwidth}{0.7pt}\scriptsize #2
    \noindent\rule[0pt]{0.5\textwidth}{0.6pt}\normalsize}\relax}
\begin{document}

\maketitle
{\let\thefootnote\relax
 \begin{NoHyper}
   \footnotetext{\hspace*{-0.44em}$^*$Equal contribution.}
 \end{NoHyper}}

\begin{abstract}
Evaluating long-form responses to research queries heavily relies on expert annotators, restricting attention to areas like AI where researchers can conveniently enlist colleagues.
Yet, research expertise is abundant: survey articles consolidate knowledge spread across the literature.
We introduce \dataset, a resource for evaluating LLM systems by distilling survey articles from 75 research fields into 21K queries and 160K rubric items.
Queries and rubrics are jointly derived from survey sections, where rubric items list query-specific answer evaluation criteria, i.e., citing papers, making explanations, and describing limitations.
31 Ph.D. annotators in 8 fields judge that 90\% of queries reflect Ph.D. information needs and 87\% of rubric items warrant emphasis of a sentence or longer.
We leverage \dataset to evaluate 18 systems in 7.6K head-to-heads. 
No parametric or retrieval-augmented system we evaluate exceeds 70\% on covering rubric items, and the highest-ranking system shows 75\% coverage.
Error analysis reveals that the highest-ranking system fully addresses less than 11\% of citation rubric items, 48\% of limitation items, and 49\% of comparison items.
We release our data to facilitate more comprehensive multi-field evaluations. 
\end{abstract}

\begin{figure}[!t]
\centering
\includegraphics[width=\columnwidth]{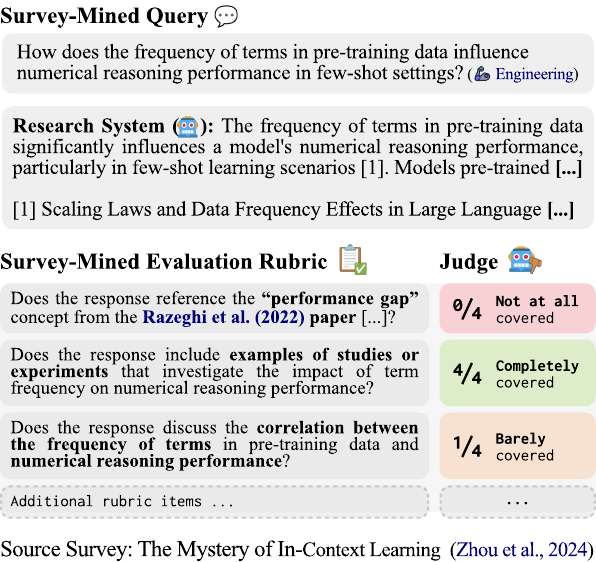}
\vspace{-15pt}
\caption{An example \dataset query and evaluation rubric. The query, mined from \citet{zhou-etal-2024}, instructs a research system to generate a long-form answer. An automatic evaluator creates an absolute measure of answer quality via a rubric with up to 8 items. The first rubric item cites \citet{razeghi2022termfreq}.
}
\label{fig:intro}
\end{figure}

\section{Introduction}

The rapid growth in research literature makes staying informed about advancements in many fields difficult~\cite{price1963little,larsen2010rate}. 
Large language model (LLM) tools, such as deep research systems \cite{geminiGeminiDeep,openaiIntroducingDeep} 
and scientific AI assistants~\cite{ skarlinski2024language,sciencediscovery,si2024llmideas,singh2025ai2}, show potential to address this problem by meeting the information needs of both experts and non-experts.
\textbf{However, evaluating long form answers to research queries is extremely challenging}~\cite{kiwi}.
Several benchmarks have been proposed~(i.a., \citealp{qasa,sciqa,openscholar,sciarena}), but are limited in size and primarily constrained to engineering domains (\Cref{table:statistics}).
Broader evaluation is necessary, but, as yet, has been unachievable because of a lack of affordable availability of appropriate experts.

In this paper, we introduce \textbf{\dataset, a literature-based resource for benchmarking research synthesis systems}.
Our insight is to leverage academic surveys, whose role is to review a research field's foundational questions and synthesize relevant evidence~\cite{kasanishi2023scireviewgen}, for conveniently making comprehensive evaluation of research topics possible.
We operationalize this idea by building a multi-stage pipeline to transform a pool of over 54K academic surveys into queries and answer \emph{evaluation rubrics}~\cite{lin2024wildbench,sawadaarb}.
Each rubric, derived jointly with queries by mining high quality survey sections, lists query-specific evaluation criteria, which may include citing papers, and making comparisons, or describing causal effects, among others (\Cref{fig:intro}).
To demonstrate the effectiveness of literature-based evaluation, we release \dataset, containing  \textbf{\statqueries queries with 160k rubric items, across 75 research fields}, making it the most diverse and comprehensive benchmark of its kind to date~(\Cref{table:statistics,table:field-query-breakdown}).

We validate the quality of \dataset in a multi-field expert evaluation, spanning \statannotators Ph.D. level annotators with expertise from 8 fields.
90\% of queries are judged as reflecting information needs of Ph.D. students and 86\% are stylistically similar to how a Ph.D. student would have expressed that need.
Few queries are considered too open-ended, facilitating the opportunity for criteria-based evaluation over open interpretation.
We also evaluate the quality of rubric items, finding that 87\% of them address concepts that should be covered in at least a sentence by systems. 

We explore whether rubrics can improve LLM-as-a-Judge accuracy.
We sample research queries and ask the same annotators to vote for the higher response quality between two retrieval-augmented systems.
Expert preferences are aggregated using majority voting and find that human accuracy to these labels is 84\%, which approximates an upper bound for LLM-as-a-Judge accuracy.
Proprietary and open-sourced judges that directly predict preference show 71\% and 64\% accuracy respectively, likely because they lack appropriate expert knowledge. 
To bridge this knowledge gap, we score each response from 0 to 4 for each rubric item's \emph{coverage}~(\Cref{fig:intro}).
To our excitement, we find that agreement increases to 74\% and 69\% when coverage and direct predictions are combined in an ensemble LLM-as-a-Judge protocol.

\textbf{With \dataset, we extensively evaluate frontier systems, revealing significant performance gaps between how different inference pipelines address scholarly queries.}
We consider 18 systems across four tiers of test-time inference effort: language models using solely parametric memory, naive retrieval on Google Scholar, production-level retrieval optimized for consumer use, and deep research APIs.
Each system is judged on two expert-validated evaluation metrics: (1) average rubric coverage \%, and (2) Elo score using our ensemble LLM-as-a-Judge.
The results reveal four noteworthy observations:

\emph{All systems struggle to cover rubric items}.
No parametric or retrieval system we evaluate exceeds 70\% coverage.
The highest-performing system we evaluate is Perplexity's deep research~\cite{perplexity2025deepresearch}, with 75.29\% coverage, indicating room for improvement.

\emph{Models selectively benefit from retrieval}.
When only relying on parametric memory, Claude-Sonnet-4~\cite{claude} ranks lower than Gemini-2.5-Pro~\cite{geminitwofive} (30\% win rate), but when both models use retrieval Claude-Sonnet-4 ranks higher (75\% win rate).

\emph{Retrieval optimization is critical}.
We observe that a naive retrieval implementation is often insufficient to improve coverage.
On average, the same models differ by 7.6\% coverage between naive and production retrieval.

\emph{Deep research systems are significantly advantaged for scholarly queries}.
Strikingly, Perplexity's deep research obtains 82\% win rate over the next best system, revealing substantial gaps between the helpfulness of research-oriented tools.

Finally, we conduct an error breakdown of unaddressed rubric items, revealing that systems can improve across many criteria: common error cases include citing key works (under-addressed in 89\% of cases), describing limitations (52\% of cases), and making comparisons (51\% of cases).
\newcommand{\pipelinemodel}[0]{$\bm{\mathcal{M}}$\xspace}

\newlength{\benchmarkcitationlength}
\setlength{\benchmarkcitationlength}{2.3cm}
\newcommand{\benchmarkcitation}[2]{#1~{#2}}
\newcommand{\subfields}[2]{
    {#1}~\,{(#2)}}
\newcommand{\markdesc}[2]{
    {#1}~{\scriptsize #2}
}
\definecolor{limegreen}{rgb}{0.396, 0.704, 0.136}
\definecolor{salmon}{rgb}{0.980, 0.302, 0.347}
\newcommand{\cmark}{\color{limegreen} \ding{51}\color{black}}
\newcommand{\xmark}{\color{salmon}\ding{55}\color{black}\hspace{1.5pt}\null}

\begin{table*}
  \centering
  \footnotesize
  \setlength{\tabcolsep}{4.8pt}
  \begin{tabular}{lrrrrrr}
    \toprule
        \makecell[l]{
            \textbf{Scholarly QA} \\
            \textbf{Benchmark}
        } & 
        \makecell[r]{
            \textbf{\# Scholarly} \\
            \textbf{Queries}
        } & 
        \makecell[r]{
            \textbf{\# Scholarly Fields \&} \\
            \textbf{Research Domains}
        } &
        \makecell[r]{
            \textbf{Abstractive} \\
            \textbf{Eval. Format}
        } &
        \makecell[r]{
            \textbf{Multi-Doc} \\
            \textbf{Reasoning}
        } &
        \makecell[r]{
            \textbf{Evaluation} \\
            \textbf{Rubrics}
        } &
        \makecell[r]{
            \textbf{Auto-} \\
            \textbf{Generated}
        }
        \\
    \midrule
        \benchmarkcitation{QASPER}{(1)} &
        5.0K &
        \subfields{1}{NLP $\leftarrow$ \iconengineering} &
        \markdesc{\xmark}{Extractive} & \xmark & \xmark & \xmark
    \\
        \benchmarkcitation{QASA}{(2)} & 
        1.8K & 
        \subfields{1}{AI $\leftarrow$ \iconengineering} &
        \markdesc{\xmark}{Extractive} & \xmark & \xmark & \xmark
    \\
        \benchmarkcitation{PubMedQA}{(3)} & 
        1.0K & 
        \subfields{---}{\iconhealth} &
        \markdesc{\xmark}{Yes/No} & \xmark & \xmark & \cmark
    \\
        \benchmarkcitation{SciQA}{(4)} &
        2.5K &
        \subfields{1}{CS $\leftarrow$ \iconengineering} &
        \markdesc{\xmark}{Extractive} & \xmark & \xmark & \cmark
    \\
        \benchmarkcitation{KIWI}{(5)} &
        0.2K &
        \subfields{1}{NLP $\leftarrow$ \iconengineering} &
        \markdesc{\cmark}{Abstractive} & \cmark & \xmark & \xmark
    \\
        \benchmarkcitation{SciDQA}{(6)} &
        2.9K &
        \subfields{1}{AI $\leftarrow$ \iconengineering} &
        \markdesc{\xmark}{Extractive} & \cmark & \xmark & \cmark
    \\
        \benchmarkcitation{SciQAG}{(7)} &
        188.0K &
        \subfields{20}{C\&MS $\leftarrow$ \iconphysical} &
        \markdesc{\xmark}{Extractive} & \xmark & \xmark & \cmark
    \\
        \benchmarkcitation{ScholarQABench}{(8)} &
        3.0K &
        \subfields{---}{\iconhealth \iconengineering \iconphysical} &
        \markdesc{\cmark}{Abstractive} & \cmark & \cmark & \xmark
    \\
        \benchmarkcitation{SciArena$^\dagger$}{(9)} &
        8.2K &
        \subfields{---}{\iconhealth \iconlife \iconengineering \iconphysical \iconsocial \iconhumanities \iconbusiness} &
        \markdesc{\cmark}{Abstractive} & \cmark & \xmark & \xmark
    \\
    \startgraymidrule
\benchmarkcitation{\dataset}{(Ours)} & 
        \statqueries & 
        \subfields{\statfields}{\iconhealth \iconlife \iconengineering \iconphysical \iconsocial \iconhumanities \iconbusiness} &
        \markdesc{\cmark}{Abstractive} & \cmark & \cmark & \cmark
    \\
    \graybottomrule
  \end{tabular}
  \caption{Comparison of \dataset to related benchmarks. Icons: \iconhealth~Health Sciences \& Medicine; \iconlife~Life \& Earth Sciences; \iconengineering~Engineering \& CS; \iconphysical~Physical Sciences; \iconsocial~Social Sciences; \iconhumanities~Humanities; \iconbusiness~Economics.
  (1)~\citet{qasper};
  (2)~\citet{qasa};
  (3)~\citet{pubmedqa};
  (4)~\citet{sciqa};
  (5)~\citet{kiwi};
  (6)~\citet{scidqa};
  (7)~\citet{sciqag};
  (8)~\citet{openscholar};
  (9)~\citet{sciarena}
  ($^\dagger$Concurrent work).
  Scholarly fields are marked as blank ``---'' when field descriptions are incomplete or metadata are missing.
  }
  \label{table:statistics}
  \vspace{-3px}
\end{table*}

\section{\dataset: Mining Queries and Rubrics from Academic Surveys} \label{sec:dataset}
\dataset is a large-scale research question answering dataset, consisting of \textbf{\statqueries queries across \statfields research fields}.
Research fields span 7 domains: \iconhealth~Health Sciences \& Medicine (7.5K queries), \iconlife~Life \& Earth Sciences (4.9K), \iconengineering~Engineering \& Computer Science (4.7K), \iconphysical~Physical Sciences (2.5K), \iconsocial~Social Sciences (1.4K), \iconhumanities~Humanities (362), and \iconbusiness~Economics (55). 
\Cref{table:statistics} shows that \dataset contains similar desiderata of related benchmarks, while expanding on queries and research field diversity.
\Cref{table:field-query-breakdown} shows the number of queries for each field.

\begin{figure*}[!t]
\centering
\includegraphics[width=0.85\textwidth]{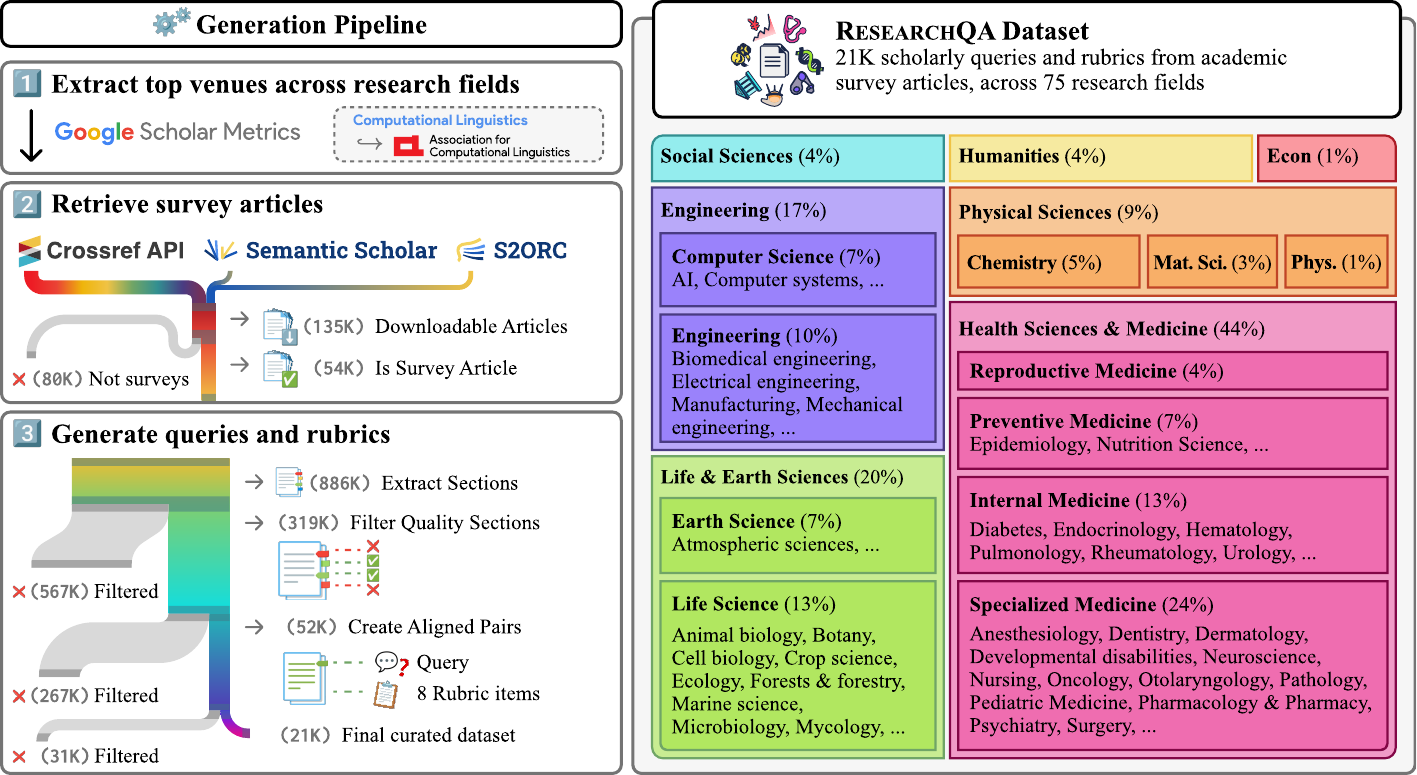}
\vspace{-5pt}
\caption{
\textbf{(Left) \dataset generation stages:} We identify top-20 venues from each field in Google Scholar, retrieve survey articles from available databases, and generate queries and rubrics from survey sections.
Throughout generation, we employ appropriate filtering mechanisms to ensure data quality.
\textbf{(Right) \dataset test split field distribution:} Queries in the test split span 75 research fields from 7 domains, with high representation in Health Sciences \& Medicine, Life \& Earth Sciences, and Engineering.}
\label{fig:setup}
\end{figure*}

To build \dataset, we create a multi-stage pipeline to generate queries and rubrics from survey articles (\Cref{fig:setup}).
We describe the selection of top publication venues in \Cref{subsec:pipeline-one} and survey articles in \Cref{subsec:pipeline-two}. 
We present query generation in \Cref{subsec:pipeline-three} and rubric generation in \Cref{subsec:pipeline-four}.
Finally, we describe dataset splits in  \Cref{subsec:dataset-splits}.
Further details about pipeline implementation are in \Cref{appendix:dataset_building}.
To balance the cost and performance of pipeline creation, we select \pipelinemodel$\leftarrow$ \gptfouronemini whenever an LLM is used to generate or filter data.
We discuss model alternatives in \Cref{sec:pipeline-model-analysis} and present a cost breakdown of different model \pipelinemodel alternatives for the data pipeline in \Cref{table:model_costs}.

\subsection{Extract Top Venues from Research Fields}\label{subsec:pipeline-one}

To focus on high-quality surveys, we identify the top-20 publishing venues ranked by h5-index for each field in Google Scholar.
Fields on Google Scholar can be overly specific (e.g., Wood Science \& Technology) or general (e.g., Health \& Medical Sciences (general)), so we manually redistribute from 257 to 94 fields. 
This process results in 660 venues that are used to retrieve survey articles.

\subsection{Extract Survey Articles}\label{subsec:pipeline-two}

We retrieve survey articles using keyword search on three sources: Crossref,\footnote{\scriptsize \url{https://www.crossref.org/}} Semantic Scholar,\footnote{\scriptsize \url{https://www.semanticscholar.org}} and S2ORC~\cite{s2orc}.
In total, 615K candidate articles are returned, 134K of which are downloadable full-text articles, and 54K which are automatically classified to represent a true survey article versus merely containing survey-related keyword(s) in the title.

\paragraph{Article search.}
For each of 660 unique venues from Google Scholar, we retrieve publications with title keywords:
\keyword{survey}, 
\keyword{literature review}, 
\keyword{a review}, 
\keyword{an overview}, and
\keyword{meta-analysis}.

\paragraph{Article filters.}
To optimize precision in our search, we apply filters to remove erroneous articles (e.g., \keyword{survey} can be used in the context of field observations, not literature reviews). 
We prompt \pipelinemodel to classify articles as literature reviews (F1$=$.80, Prec$=$.87), removing those judged not to be a literature review, i.e., work aimed at synthesizing and reviewing existing literature. 

\subsection{Generate Queries from Survey Content}\label{subsec:pipeline-three}

We identify medium-length survey sections with multiple citations, then generate questions and reference answers using grouped sentences from each section.
Queries are filtered by keywords and \pipelinemodel to ensure they are standalone and without excessive variability in appropriate answers (F1=.86, Prec=.75). 
In total, \statsectionsmincitation of \statsectionstotal sections are used, yielding \statqueriestruncated queries after filtering.

\paragraph{Survey section filters.}
Sections are removed if section titles suggest they are not part of the main body (e.g.,
\keyword{abstract}).
Next, we remove sections too short to be informative ($<$~3 sentences or $<$~800 characters), too long ($>$~300K characters), or lacking citations ($<$~3 inline citations).

\paragraph{Query and reference answer generation.}
We few-shot prompt \pipelinemodel to extract hierarchical summaries~\cite{christensen2014hierarchical} from section content.
Each summary consists of tree-based structure of questions with supporting sentences found in the section content~\cite{benz2017questions,wu2023qudeval}, helping to generate queries that integrate multiple sources of supporting information~(\Cref{fig:question_tree}). 
We prompt \pipelinemodel with examples, using section content and the summary to generate a query and reference answer.
Prompts instruct that queries must be supported by at least 3 sentences and reference answers summarize supporting sentences without introducing new evidence.
Like~\citet{jansen2025matter}, each query is paired with a knowledge cut-off date of the source article.

\begin{figure*}[!tbp]
\centering
\includegraphics[width=\textwidth]{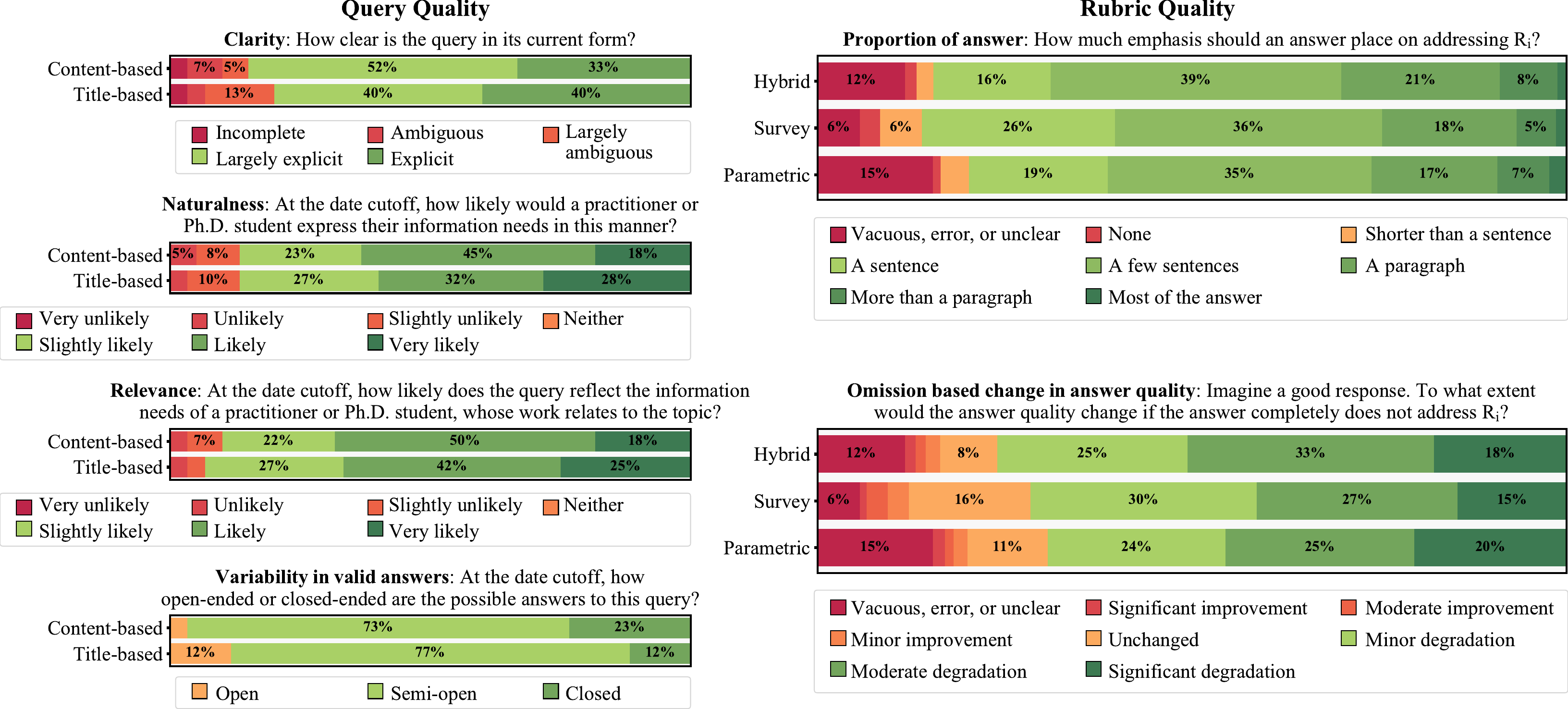}
\vspace{-16pt}
\caption{\dataset query and rubric quality ratings by 31 Ph.D. level experts.}
\label{fig:query-rubric-quality}
\end{figure*}

\paragraph{Query filters.}
We aim for queries that are (1)~\emph{standalone}, i.e., understandable by experts without extra decontextualization~\cite{choi2021decontextualization}; and (2)~\emph{low in answer variability}, i.e., different experts are likely to provide similar ratings so that model responses can be more easily scored. 
To enforce these criteria, we use \pipelinemodel to score standaloneness and answer variability of each query on a scale from 1 to 10.
Queries scoring $<$~7 for standaloneness or $>$~4 for answer variability are removed.\footnote{Thresholds validated by NLP experts on initial dev. data. }
Further, we discard queries that have keywords indicating context dependence (e.g., \keyword{the paper}, \keyword{this study}).
We then remove queries with reference answers that are too short ($<$~800 characters).
Finally, we re-apply \pipelinemodel to remove queries that are misassigned to a field, which can occur when mining queries from multi-field venues.

\subsection{Rubric Generation}
\label{subsec:pipeline-four}
Rubrics $\mathbf{R}=\{R_i\}_{i=1}^K$ consist of $K$ automatically generated rubric items $R_i$ that each evaluate aspects of answer quality.
Below, we present the design choices and method for rubric creation.

\paragraph{Desiderata.}
Rubrics are distilled from survey articles, adopting limitations from the articles themselves.
For example, citations and analysis in the survey article can be subjective.
Consequently, there may be gaps in evaluation criteria, e.g., an important missing $R_i$, and not all rubric items may be important.
We address these potential limitations through two design choices. 
(1) We generate and evaluate three rubric types: 
survey rubrics (\rubricsurvey), parametric rubrics (\rubricparametric), and hybrid rubrics that merge the two (\rubrichybrid $\subset$ (\rubricsurvey $\cup$ \rubricparametric)), covering obvious gaps in rubrics~\cite{wadhwa2025evalagent}.
(2) To create \rubrichybrid, we deduplicate, rerank, and remove hallucinations from the union set \rubricsurvey $\cup$ \rubricparametric, which intends to remove unimportant or flawed rubric items.
The top-8 hybrid rubric items are retained, leaving $\sim$7.5 rubric items on average.

\paragraph{Rubric item generation.}
We few-shot prompt \pipelinemodel, using question-answer pairs from ScholarQABench~\cite{openscholar}, to generate rubric items.
\rubricsurvey is conditioned on both the query and reference answer; \rubricparametric is conditioned on only the query.
Each rubric item is created from one of three prompts designed to increase diversity of rubric item topics: information-based, depth-based, and citation-based items.
Information-based items ask for specific statements, findings, opinions, or comparisons.
Depth-based items ask for elaboration or explanation.
Citation-based items ask whether answers cite a specific article.
To create \rubricsurvey and \rubricparametric with diverse items, we sample 4 information-based items, 2 depth-based items, and 2 citation-based items.

\paragraph{Hybrid rubric construction.}
Using \pipelinemodel, we deduplicate repeating rubric items that represent identical criteria and rerank rubric items based on their importance for answer evaluation.
To remove hallucinations, we remove rubric items referencing papers that cannot be automatically matched to a Google Scholar article.
Post-filters, about 61\% of hybrid rubric items come from \rubricsurvey.

\paragraph{Baseline rubric construction.}
We additionally compare against a generic rubric (\rubricgeneric) that does not take query specifics into account.
Rubric items are sampled from ScholarQABench~\cite{openscholar} and SciArena~\cite{sciarena} metrics: Correctness, Citations, Coverage, Relevance, Organization, Usefulness, Attribution, and Examples.
\Cref{table:generic-rubric} presents the full list of rubric items.

\subsection{Dataset statistics and splits}
\label{subsec:dataset-splits}

We create train ($\mathcal{D}_\text{train}$), validation (\dataval), and test (\datatest) splits.
\datatest (3.7K queries) samples 50 queries from each field with $\geq$~50 queries (\statfields of 94 fields), so that each field is sufficiently represented.
\dataval (703 queries) samples up to 10 queries from remaining fields (74 of \statfields), and $\mathcal{D}_\text{train}$ is made up by the remaining 16.9K queries, which are for supporting the community in developing and tuning research systems.
\begin{table}[t]
\footnotesize
\centering
\setlength{\tabcolsep}{3pt}
\renewcommand{\arraystretch}{0.98}

\begin{tabular}{c l r r}
\toprule
\textbf{Section} & \textbf{Annotation Type} & \textbf{PA} \\
\midrule
\multirow{4}{*}{\Cref{subsec:val-query}} & Clarity & .72\\
& Naturalness & .73\\
& Relevance & .83\\
& Variability in Valid Answers & .88\\
\midrule
\multirow{2}{*}{\Cref{subsec:val-rubric}} & Proportion of Answer & .91\\
 & Omission Based Change in Answer Quality & .94\\
\midrule
\multirow{2}{*}{\Cref{subsec:val-judgements}} & Rubric Coverage & .73\\
& Pairwise Preference & .69\\
\bottomrule
\end{tabular}

\caption{Inter-annotator agreement across experts.}
\label{table:agreement}
\end{table}

\section{Expert Validation}
\label{sec:expert-eval}

This section  presents the method and results of the expert validation study.
We collect expert annotations for three judgement types: query quality (\Cref{subsec:val-query}), rubric quality (\Cref{subsec:val-rubric}), and scoring and preferences on system outputs (\Cref{subsec:val-judgements}).
Judgements are made on Likert-scale and multi-category labels, visualized in \Cref{fig:query-rubric-quality}, and are binarized by meaning for downstream analysis.

\paragraph{Annotator recruitment.}
We recruited annotators via Ph.D. email lists.
45 Ph.D. students and 1 postdoctoral staff registered to participate in annotations, and \statannotators ultimately completed the annotation task.
Annotators' expertise span diverse fields: 
Natural Language Processing (15~experts), Computer Vision (6), Biomedicine (4), Linguistics (3), Physics (2), Genetics (1), Economics (1), and Psychology (1).
Annotators were compensated \$25 per hour of annotation.

\paragraph{Annotation agreement.}
We measure inter-annotator agreement (IAA) as binary pairwise agreement (PA) in \Cref{table:agreement}.
While tasks can be subjective, IAA is similar to prior work exploring expert judgements for scientific QA (e.g., pairwise preference is .69 vs .70 PA in \citet{openscholar}).

\subsection{Query Quality}
\label{subsec:val-query}

\paragraph{Setup.}
Queries are judged on clarity, naturalness, relevance, and variability in valid answers.
We compare queries from our pipeline (content-based) against an ablated pipeline that generates solely from survey titles (title-based), controlling for comparisons on the same query topic.
This analysis isolates the effect of pipeline components used to enhance query generation and filtering.

\paragraph{Label binarization.}
We binarize labels by positive and negative meaning.
Clarity merges \emph{Largely explicit} and \emph{Explicit}; Naturalness merges \emph{Slightly likely} to \emph{Very likely}; Relevance merges \emph{Slightly likely} to \emph{Very likely}; and Variability in valid answers merges \emph{Semi-open} and \emph{Closed}.

\paragraph{Survey content yields unambiguous queries, eliciting answers without too much variability.}
Experts rate $\sim$85\% of content-based queries as \keyword{Largely explicit} or \keyword{Explicit}, indicating that queries generated from our pipeline have mostly specific and easy interpretations.
Likely, content-generated queries have specific interpretations due to grounding in a reference text.
Additionally, content-based queries have more \keyword{Semi-open} or \keyword{Closed-ended} queries (96\% versus 89\%), which supports valid answers that can be directly compared.
By contrast, \keyword{Open-ended} queries might elicit a large number of possible answers.
For example, \emph{``What distinguishes \underline{robustness} under \underline{distribution shift} from \underline{domain adaptation} and \underline{transfer learning} in NLP?''} can yield answers that experts find difficult to rank and compare.

\paragraph{Queries support the information needs of researchers and are naturally expressed.}
Queries generated from both full and ablated pipelines have similar relevance and naturalness.
94-90\% are rated as \emph{Slightly likely} or stronger to reflect researchers' information needs, and 86\% of queries \emph{Slightly likely} or stronger to be expressed in that way by a researcher.
These ratings also indicate room for improvement, because queries rarely are rated as \emph{Very likely} to be relevant or naturally expressed.

\subsection{Rubric Quality}
\label{subsec:val-rubric}
\paragraph{Setup.}
Rubric items are judged on the number of sentences that should be used to address the item in an answer and how its omission affects answer quality (\Cref{fig:query-rubric-quality}).
Additionally, annotators can flag a number of errors of a rubric item \rubricitem: 
(1) \rubricitem is difficult to judge; 
(2) \rubricitem is unclear; 
(3) \rubricitem is an empty or unspecific assessment (e.g., it is a mere rephrasing of the query);\footnote{A rule-based system is also employed to detect unspecific rubric items, where rubric items need to include at least one substantive word not present in the original query.} 
(4) \rubricitem contains an error (e.g., it cites a non-existent paper).
The presence of any of these flags voids the quality of the rubric item.
We compare among hybrid, survey, and parametric rubrics.

\paragraph{Label binarization.}
In ``Proportion of answer measurement'', we merge \emph{None} and \emph{Shorter than a sentence}.
In ``Omission based change in answer quality'' we merge \emph{Significant improvement} to \emph{Minor improvement}.

\paragraph{Hybrid rubrics are likely to contain criteria worth describing in multiple sentences and improve answer quality.}
Across rubric types, rubric items should be described in \keyword{A sentence} or more 84-86\% of the time and their omission would degrade answer quality 69-74\% of the time.
Hybrid rubrics items are rated to be greatly important at a higher frequency, specifically those causing \emph{Moderate degradation} or \emph{Significant degradation} to answer quality when removed. 

\paragraph{Hybrid rubrics contain few non-existent papers and vacuous restatements of the query.}
Parametric rubrics have the highest rate of \keyword{vacuous, error, or unclear} items (15\%).
Experts note that parametric rubric items contain nonexistent paper titles, causing statements about their coverage in answers to be unanswerable or ambiguous.

\subsection{System Output Preferences and Evaluation Protocol Validation}
\label{subsec:val-judgements}

\paragraph{Setup.}
We collect expert judgements on head-to-head answers (\Cref{fig:eval_interface}) to perform a meta-evaluation on LLMs using rubrics to approximate expert judgements.
Experts provide two types of annotations:
(1) Rubric coverage: Experts rate how well each answer covers each rubric item on a 5-point scale (0 = \keyword{Not at all}, 4 = \keyword{Completely}).
(2) Pairwise preference: Experts compare two answers side-by-side (in random order) and select: \keyword{Left} is better, \keyword{Right} is better, \keyword{Tie}, or \keyword{Both bad}.
Majority voting determines the final label. 

Answers are generated from \gptfouronemini and \geminitwofiveflash using their providers' embedding models\footnote{\texttt{\scriptsize text-embedding-3-large}, \texttt{\scriptsize text-embedding-004}} to retrieve relevant passages as input context.
We retrieve the top-20 arXiv papers via Google search on the given query and constrain search by the appropriate date cutoff.
Papers are chunked into 1000 character passages, and the top-20 passages ranked by embedding similarity score are input as context.

\paragraph{Label binarization.}
Rubric coverage merges (0 = \emph{Not at all}, 1 = \emph{Barely}) and (2 = \emph{Moderately}, 3 = \emph{Mostly}, 4 = \emph{Completely}).
For pairwise preference, we drop \emph{Tie} and \emph{Both bad}, keeping only direction labels, i.e., \emph{Left} or \emph{Right}.
PA are 0.73 and 0.69 respectively.
Because agreement for pairwise preference is low, we collect up to 4 expert judgements and keep only those that had a majority voted direction label (89.7\% of cases with direction judgements).
Human annotators with direction label preferences have 84\% accuracy agreement with the majority vote label.

\paragraph{Evaluator LLMs.}
Open-source evaluators (\prometheustwomoe; \citealp{prometheus-two}) and proprietary LLMs (\claudefoursonnet and \gptfouronemini) are validated as automatic evaluators.
For each evaluator, we compute agreement to expert annotations on rubric coverage and pairwise preference, which represent absolute and relative measures of answer quality respectively.

We prompt an LLM to produce rubric coverage in one call, inputting a system answer $A$ and rubric $\mathbf{R}=\{R_i\}^K_{i=1}$ to quantify the number of rubric items covered in $A$ on a 5-point scale, $\mathrm{Coverage}:A\times\mathbf{R}\rightarrow\{0,1,2,3,4\}^K$.
The prompt describes the ends of the scale with labels from the user study (0 = \keyword{Not at all}, 4 = \keyword{Completely}).

We compare two types of automatic judges to predict pairwise preference.
The \textbf{Direct Judge} (\metricdirect) is prompted with the query and two system answers and asked ``Which response is better?''.
Both answer orderings are evaluated to reduce positional bias~\cite{shi2024judging}.
However, direct comparison may lack appropriate knowledge useful to predict expert preferences.
To address this gap, we introduce \textbf{Ensemble Judge} (\metricensemble), which uses both \metricdirect and $\mathrm{Coverage}$ to predict preferred answers: the answer with a larger sum of $\mathrm{Coverage}$ (0-4 scale) and a full 4 points for each \metricdirect comparison, is marked as the preferred answer.
Formally, this sum can be expressed as:
\begin{equation}
    4(\mathbb{I}_A + \mathbb{I}'_A) + \textstyle\sum_{c \in \mathrm{Coverage}(A, \mathbf{R})} c
\end{equation}
where $\mathbb{I}_A$ indicates 1 if $A$ is preferred by \metricdirect and 0 otherwise;
$\mathbb{I}'_A$ implements \metricdirect with answers input in the reverse order.
\metricensemble~prefers the answer with the larger sum or outputs a tie.

\begin{figure}[!t]
\centering
\includegraphics[width=\columnwidth]{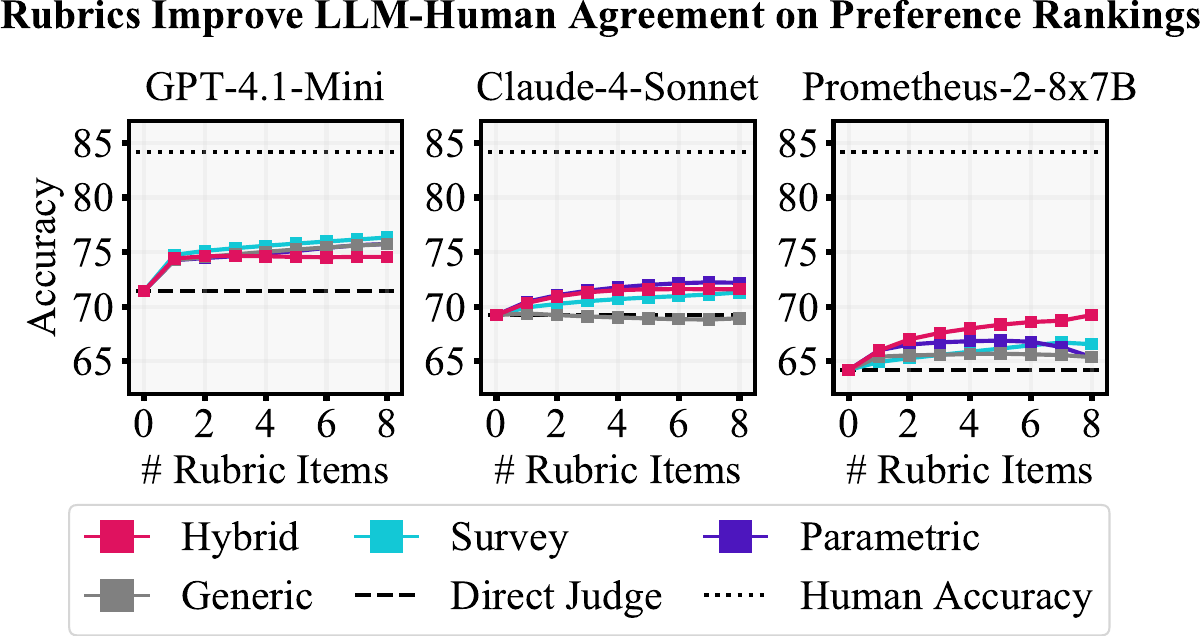}
\vspace{-20pt}
\caption{A comparison of how much rubrics can aid different evaluators in making predictions that agree with plurality human labels (y-axis) as a function of rubric size (x-axis). 
All direct judges benefit from integration of rubrics through the hybrid judge, substantially reducing their disagreement with human experts.
}
\vspace{-10pt}
\label{fig:auto-eval-reliability}
\end{figure}

\paragraph{LLM-Human agreement on Rubric Coverage.}
All evaluator LLMs show fair Pearson correlation with expert annotations on rubric coverage: \prometheustwomoe at .48, and both \claudefoursonnet and \gptfouronemini at .63.
Averaged over our annotated data, experts and \gptfouronemini differ on $\mathrm{Coverage}$ by only 0.11. 
On the other hand, \gptfouronemini tends to make more extreme predictions of $\mathrm{Coverage}$ for individual samples. 
Experts rate a rubric items \keyword{Completely} covered 12.3\% of the time while \gptfouronemini does so 26.0\%. 
While a suitable judge on average, conclusions with \gptfouronemini may overestimate the frequency of some coverage values at the extremes. 

\paragraph{LLM-Human agreement on Pairwise Preference.}
We present the preference ranking accuracies of evaluator LLMs with respect to majority labels in \Cref{fig:auto-eval-reliability}.
Each graph presents a base direct judge and the results of ensembling with the four types of rubrics we consider.
In cases where judges predict ties, we assign partial credit.\footnote{ We have 3 cases of partial credit: (1) for the direct judge, if $\mathbb{I}_A \neq \mathbb{I}'_A$ , 50\%, (2) if $\mathbb{I}_A$ or $\mathbb{I}'_A$ is correct and the other reports tie, 75\%, and (3) if \metricensemble~reports tie, 50\%. }
Human-Human accuracy agreement indicates an upper bound of 84\% for LLM-Human agreement.
Ensemble judges (\metricensemble) are consistently better estimators of expert preferences than direct judges (\metricdirect), achieving up to $\sim$75\% accuracy. 
Notably, rubrics can decrease the LLM-Human and Human-Human agreement gap from 12.7\% to 9.6\%, corresponding to a 24\% relative reduction. 
In general, query-specific rubrics match or outperform generic rubrics.
Hybrid rubrics demonstrate high performance gain across all systems, allowing \prometheustwomoe to match the direct judge performance of \claudefoursonnet.
As visualized in \Cref{fig:auto-eval-by-field}, hybrid rubrics enhance evaluation in 7 of 8 fields, where they are the best-performing rubric in 4 of 8.
The performance of parametric and generic rubrics depend on the strength of the evaluator: in strong evaluators, they can match hybrid rubrics, but they yield little to no benefit in open evaluators.

\newcommand{\coverage}[3]{
    #1
}
\newcommand{\coverageall}[4]{
    #1 $\pm$ {#4}
}

\begin{table*}[!t]
  \centering
  \scriptsize
  \setlength{\tabcolsep}{2pt}
  \begin{tabular}{@{}lrrrrrrrrrrcrc}
    \toprule
        \multirow[t]{2}{*}{\raisebox{-0.5\aboverulesep}{\textbf{System}}} &
        \hspace{0.2em} &
        
        \multicolumn{8}{c}{\textbf{Coverage~\%~\raisebox{.1em}{$\uparrow$}}} &
        \hspace{0.2em} &
        
        \multirow{2}{*}{\raisebox{-3\aboverulesep}{\makecell[c]{\textbf{Leaderboard} \\ \textbf{\phantom{$\uparrow$}~Score~\raisebox{.1em}{$\uparrow$}}}}} & 
        \hspace{0.2em} &
        
        \multirow{2}{*}{\raisebox{-3\aboverulesep}{\makecell[c]{\textbf{Avg Length} \\ \textbf{(Words)}}}}\\
        \cmidrule{3-10}
        &&
        \makecell[l]{All domains} &
        \makecell[l]{\iconhealth} &
        \makecell[l]{\iconlife} & 
        \makecell[l]{\iconengineering} & 
        \makecell[l]{\iconphysical} & 
        \makecell[l]{\iconsocial} &
        \makecell[l]{\iconhumanities} & 
        \makecell[l]{\iconbusiness} & & & & \\
    \midrule
        {\scriptsize Parametric} \\
        \cmidrule(r){1-1}
{\scriptsize\iconmeta~\llamathreethreeseventyb                }&&  \coverageall{53.42}{52.9}{53.9}{0.26} & \coverage{51.82}{51.1}{52.6} & \coverage{54.21}{53.2}{55.3}  & \coverage{54.89}{53.7}{56.2} & \coverage{55.74}{54.4}{57.3} & \coverage{53.91}{51.2}{56.3} & \coverage{53.22}{50.6}{55.6} & \coverage{58.10}{53.2}{62.4} &&\phantom{0}617 ± 13&&167.4\\
{\scriptsize\iconanthropic~\claudefoursonnet                  }&&  \coverageall{64.31}{63.7}{64.9}{0.31} & \coverage{62.92}{62.0}{63.9} & \coverage{64.96}{63.5}{66.3}  & \coverage{66.71}{65.2}{68.2} & \coverage{67.08}{65.2}{68.9} & \coverage{63.33}{59.7}{66.9} & \coverage{59.85}{56.5}{63.3} & \coverage{66.17}{61.0}{71.4} && \underline{1099} ± \underline{09} &&226.5\\
{\scriptsize\iconopenai~\gptfourone                           }&&  \coverageall{65.43}{65.0}{65.9}{0.25} & \coverage{63.98}{63.2}{64.8} & \coverage{66.84}{65.8}{67.9}  & \coverage{66.66}{65.5}{67.8} & \coverage{67.38}{66.0}{68.8} & \coverage{65.45}{62.8}{68.0} & \coverage{63.48}{60.9}{65.8} & \coverage{68.46}{63.5}{73.4} &&1080 ± 09&&241.5\\
{\scriptsize\iconqwen~\qwenthreethirtytwob                    }&&  \coverageall{\underline{66.64}}{66.1}{67.2}{\underline{0.26}} & \coverage{\underline{65.13}}{64.3}{65.9} & \coverage{\underline{67.76}}{66.6}{68.9}  & \coverage{\underline{68.25}}{67.1}{69.4} & \coverage{\underline{69.24}}{67.8}{70.8} & \coverage{\underline{65.96}}{63.3}{68.4} & \coverage{\underline{64.32}}{61.9}{66.7} & \coverage{\underline{69.62}}{64.9}{74.3} &&1038 ± 09&&219.3\\
{\scriptsize\icongoogle~\geminitwofivepro                     }&&  \coverageall{\textbf{68.84}}{68.3}{69.3}{\textbf{0.25}} & \coverage{\textbf{67.42}}{66.6}{68.2} & \coverage{\textbf{70.20}}{69.2}{71.2}  & \coverage{\textbf{69.82}}{68.6}{71.0} & \coverage{\textbf{71.86}}{70.5}{73.2} & \coverage{\textbf{68.28}}{65.2}{71.1} & \coverage{\textbf{65.83}}{63.3}{68.2} & \coverage{\textbf{72.06}}{67.2}{77.0} && \textbf{1244} ± \textbf{10} &&267.1\\
    \cmidrule(r){1-1}
        {\scriptsize Retrieval (Naive)} \\
        \cmidrule(r){1-1}
{\scriptsize\iconallenai~\openscholareightb                   }$^{i}$&&  \coverageall{54.71}{54.1}{55.2}{0.28} & \coverage{54.08}{53.2}{55.0} & \coverage{56.15}{55.0}{57.3}  & \coverage{54.76}{53.3}{56.1} & \coverage{54.98}{53.5}{56.5} & \coverage{54.67}{52.0}{57.2} & \coverage{52.69}{49.8}{55.5} & \coverage{57.46}{51.9}{62.6} &&\phantom{0}478 ± 17 && 499.9\\
{\scriptsize\icongoogle~\geminitwofivepro                     }$^{ii}$&&  \coverageall{59.92}{59.3}{60.5}{0.30} & \coverage{58.73}{57.8}{59.7} & \coverage{61.46}{60.2}{62.7}  & \coverage{61.13}{59.7}{62.5} & \coverage{61.72}{59.9}{63.4} & \coverage{57.79}{54.6}{60.9} & \coverage{56.71}{53.8}{59.6} & \coverage{63.96}{59.1}{69.1} &&\phantom{0}945 ± 10 && 270.4\\
{\scriptsize\iconqwen~\qwenthreethirtytwob                    }$^{iii}$&&  \coverageall{60.90}{60.2}{61.5}{0.32} & \coverage{57.62}{56.6}{58.6} & \coverage{62.93}{61.7}{64.2}  & \coverage{\underline{64.25}}{62.9}{65.5} & \coverage{\underline{65.58}}{64.0}{67.2} & \coverage{60.49}{57.5}{63.2} & \coverage{\underline{60.23}}{57.4}{63.0} & \coverage{\underline{65.82}}{61.1}{70.9} && \underline{1011} ± \underline{09} && 265.5\\
{\scriptsize\iconanthropic~\claudefoursonnet                  }$^{iv}$&&  \coverageall{\underline{62.50}}{61.9}{63.1}{\underline{0.32}} & \coverage{\underline{61.94}}{61.0}{62.8} & \coverage{\underline{63.48}}{62.3}{64.7}  & \coverage{62.97}{61.7}{64.3} & \coverage{64.01}{62.1}{65.7} & \coverage{\underline{61.67}}{58.5}{64.5} & \coverage{58.05}{55.1}{60.6} & \coverage{65.74}{60.4}{71.6} && \phantom{0}972 ± 09 && 238.4\\
{\scriptsize\iconopenai~\gptfourone                           }$^{v}$&&  \coverageall{\textbf{64.80}}{64.2}{65.4}{\textbf{0.29}} & \coverage{\textbf{63.69}}{62.8}{64.5} & \coverage{\textbf{66.65}}{65.5}{67.8}  & \coverage{\textbf{65.11}}{63.8}{66.4} & \coverage{\textbf{66.72}}{65.1}{68.3} & \coverage{\textbf{64.25}}{61.2}{67.0} & \coverage{\textbf{62.09}}{59.2}{64.6} & \coverage{\textbf{66.33}}{61.4}{71.3} && \textbf{1020} ± \textbf{09} && 263.6\\
    \cmidrule(r){1-1}
        {\scriptsize Retrieval (Production)} \\
        \cmidrule(r){1-1}
{\scriptsize\iconperplexity~\sonar                            }&&  \coverageall{58.61}{58.1}{59.2}{0.29} & \coverage{56.61}{55.8}{57.5} & \coverage{60.55}{59.3}{61.7}  & \coverage{59.43}{58.1}{60.9} & \coverage{61.62}{59.9}{63.3} & \coverage{59.97}{56.9}{62.6} & \coverage{57.48}{54.8}{60.2} & \coverage{62.80}{57.6}{68.1} && \phantom{0}862 ± 10 && 242.2\\
{\scriptsize\iconallenai~\openscholareightbagentic            }&&  \coverageall{58.72}{58.1}{59.4}{0.32} & \coverage{57.77}{56.8}{58.6} & \coverage{59.96}{58.7}{61.2}  & \coverage{58.62}{57.2}{60.0} & \coverage{61.48}{59.8}{63.2} & \coverage{57.27}{54.0}{60.2} & \coverage{57.29}{54.5}{60.1} & \coverage{62.48}{56.8}{67.8} && \phantom{0}769 ± 12 && 788.8\\
{\scriptsize\iconperplexity~\sonarreasoning                   }&&  \coverageall{64.33}{63.7}{65.0}{0.31} & \coverage{62.73}{61.8}{63.6} & \coverage{66.00}{64.8}{67.3}  & \coverage{65.19}{63.8}{66.6} & \coverage{68.11}{66.4}{69.7} & \coverage{62.68}{59.2}{65.7} & \coverage{61.76}{59.1}{64.3} & \coverage{\underline{67.49}}{62.0}{72.7} && \underline{1115} ± \underline{10} && 280.5\\
{\scriptsize\iconopenai~\openaiwebsearch                      }&&  \coverageall{65.98}{65.5}{66.5}{0.27} & \coverage{65.52}{64.7}{66.3} & \coverage{68.21}{67.1}{69.3}  & \coverage{65.01}{63.7}{66.3} & \coverage{66.60}{65.2}{68.0} & \coverage{66.07}{63.6}{68.3} & \coverage{62.62}{59.9}{65.5} & \coverage{65.63}{61.2}{70.3} && \phantom{0}992 ± 09 && 255.0\\
{\scriptsize\icongoogle~\geminiwebsearch                      }&&  \coverageall{\underline{68.51}}{68.0}{69.0}{\underline{0.25}} & \coverage{\underline{67.38}}{66.6}{68.1} & \coverage{\underline{70.02}}{69.0}{71.0}  & \coverage{\textbf{68.76}}{67.5}{70.0} & \coverage{\textbf{70.99}}{69.5}{72.4} & \coverage{\underline{68.09}}{65.7}{70.2} & \coverage{\textbf{65.98}}{63.6}{68.3} & \coverage{\textbf{71.21}}{66.4}{75.9} && \phantom{0}960 ± 09 && 278.5\\
{\scriptsize\iconanthropic~\anthropicwebsearch                }&&  \coverageall{\textbf{69.18}}{68.7}{69.7}{\textbf{0.26}} & \coverage{\textbf{69.54}}{68.8}{70.3} & \coverage{\textbf{70.49}}{69.5}{71.5}  & \coverage{\underline{67.59}}{66.3}{68.9} & \coverage{\underline{70.28}}{68.8}{71.6} & \coverage{\textbf{68.14}}{65.2}{70.6} & \coverage{\underline{64.70}}{62.1}{67.2} & \coverage{67.13}{61.4}{72.5} && \textbf{1149} ± \textbf{10} && 327.8\\
    \cmidrule(r){1-1}
        {\scriptsize Deep Research} \\
        \cmidrule(r){1-1}
{\scriptsize\iconopenai~\openaideepresearch                   }$^{\dagger}$&&  \coverageall{\underline{72.69}}{71.6}{73.7}{\underline{0.54}} & \coverage{\underline{74.02}}{71.7}{76.1} & \coverage{\underline{73.58}}{71.1}{76.0}  & \coverage{\underline{70.57}}{67.3}{73.5} & \coverage{\underline{74.04}}{71.6}{76.5} & \coverage{\underline{73.25}}{70.7}{75.7} & \coverage{\underline{68.99}}{66.1}{71.7} & \coverage{\underline{74.54}}{70.3}{78.8} && \underline{1145} ± \underline{10} && 271.6\\
{\scriptsize\iconperplexity~\sonardeepresearch                }&&  \coverageall{\textbf{75.29}}{74.8}{75.8}{\textbf{0.26}} & \coverage{\textbf{75.01}}{74.3}{75.8} & \coverage{\textbf{76.31}}{75.2}{77.3}  & \coverage{\textbf{74.48}}{73.2}{75.7} & \coverage{\textbf{76.77}}{75.3}{78.3} & \coverage{\textbf{75.34}}{72.8}{77.8} & \coverage{\textbf{72.47}}{70.0}{75.0} & \coverage{\textbf{78.01}}{73.7}{82.0} && \textbf{1505} ± \textbf{17} && 267.3\\
    \bottomrule
  \end{tabular}
  \caption{Performances of LLM systems in a pairwise tournament across domains 
  (\iconhealth~Health Sciences \& Medicine; \iconlife~Life \& Earth Sciences; \iconengineering~Engineering \& CS; \iconphysical~Physical Sciences; \iconsocial~Social Sciences; \iconhumanities~Humanities; \iconbusiness~Economics.)
  Retrieval embedding models belong to corresponding system providers:
  {$^{i}$\texttt{openscholar-retrieve} and \texttt{openscholar-reranker}};
  {$^{ii}$\texttt{text-embedding-004}}; 
  {$^{iii}$\texttt{gte-qwen-2-7b-instruct}};
  {$^{iv}$\texttt{voyage-3-large}}; 
  {$^{v}$\texttt{text-embedding-3-large}}.
  $^\dagger$This system costs $\sim$\$1.15 per query, so statistics are only computed on answers sampled for tournament battles ($\sim$20\%).
  }
  \label{table:eval}
\end{table*}
\section{LLM Systems Evaluation Setup}
\label{sec:eval-setup}
To benchmark top systems used for scholarly inquiry on \dataset, we evaluate 18 parametric, retrieval augmented, and deep research systems in both open-source and proprietary families (\Cref{table:eval}) on \datatest.
We perform all analyses with the best-performing rubrics (hybrid rubrics), evaluator LLM (\gptfouronemini), and protocol (\metricensemble).
We describe the task setup in \Cref{subsec:eval-metrics} and tournament setup in \Cref{subsec:tournament-details}.

\subsection{Evaluation Task and Metrics}
\label{subsec:eval-metrics}

\paragraph{Task.}
We consider the following task: systems are input a query~\query and generate a citation-supported answer~\answer, constrainting on date~\datecutoff and a response length $L=\,$250 words.
A sample answer is visualized in  \Cref{fig:parametric_vs_rag}.
This task design conservatively guards against potential biases from information recency, multi-turn clarification, and length biases~\cite{rlhflengthbiases}, trading off task naturalness.
We explore more natural setups in \cref{subsec:results-additional}, finding that \datecutoff can be presently discarded with low likelihood of bias.
Further, we remove $L$, allowing longer responses that may result from multi-turn interaction or no length specification.

\paragraph{Coverage \%.}
We compute average  $\mathrm{Coverage}$ over answers and divide by 4, normalizing the resulting percentage from 0\% to 100\%.

\paragraph{Leaderboard score.}
Consistent with Chatbot Arena~\cite{chatbotarena}, we use the Bradley-Terry model~\cite{bradley-terry-model} for pairwise battles judged by \metricensemble, which is equivalent to the Elo equation~\cite{elo}.
In case of ties, each system wins half the match.
We report the median score from a 1000-iteration bootstrap and its standard deviation.

\subsection{Tournament Details}
\label{subsec:tournament-details}

We explore systems at a gradation of engineering effort: parametric, naive retrieval, production retrieval, and deep research.

\paragraph{Parametric systems.}
We directly generate answers using default configs of API providers (OpenRouter\footnote{\scriptsize \url{https://openrouter.ai/}} for open-source models).

\paragraph{Naive retrieval systems.}
These systems represent retrieval with minimal engineering effort: language models conditioned on the top-20 retrieved passages.
Each language model is paired with an embedding model from the same provider.
We retrieve papers from a date-constrained Google Scholar search.
For each query, up to 50 papers are retrieved: the top-20 using the query as search field, and an additional 10 for each of 3 related keywords generated by \gptfouronemini, a search method used in \citet{openscholar}.
Papers corresponding to distilled survey papers are removed to prevent unfair evaluation advantages due to data leakage.
Each paper is chunked into 1000-token passages with 200 overlapping tokens of subsequent chunks, and the top-20 passages by embedding similarity score are used as context, optionally reranking when a cross-encoder reranker is available.
We generate answers using the same decoding configurations as in the parametric setting and employ greedy decoding for OpenScholar.

\paragraph{Production retrieval systems.}
These systems represent retrieval with advanced engineering effort, e.g., tool use, feedback, or refinement.
We use default implementations in their respective repositories or APIs without modification.

\paragraph{Deep research systems.}
At the time of writing, two commercial deep research systems have API access: OpenAI and Perplexity deep research.
We use both APIs without modification.

\paragraph{Battle construction.}
We construct a total of 7.6K battles, where each domain consists of a minimum of 1K battles and up to as many queries represented in the domain (1.8K for Health Sciences \& Medicine).
Each battle is constructed such that models are sampled uniformly ($\sim$800 battles per model) and all model matchups are uniformly represented in each domain, ensuring connectivity of the comparison network.

\newcommand{\diffannot}[2]{#1 & \tiny{#2}}
\newcommand{\smallfont}[1]{\fontsize{6.5}{6}\selectfont #1}

\begin{table}[!t]
\centering
\scriptsize
\setlength{\tabcolsep}{1.5pt}
\renewcommand{\arraystretch}{0.98}

\begin{tabular}{l ccc ccc}
\toprule
\textbf{Systems} &
\multicolumn{3}{c}{\textbf{Avg Length}} &
\multicolumn{3}{c}{\textbf{Coverage \%}} \\
\cmidrule(lr){2-4} \cmidrule(lr){5-7}
& $L$=250 & $\neg L$ & $\Delta$\% &
$L$=250 & $\neg L$ & $\Delta$ \\
\midrule
{\scriptsize Parametric} \\
\cmidrule(r){1-1}
\smallfont{\qwenthreethirtytwob}     & 219.5 & \diffannot{\phantom{0}176.5}{\phantom{0}$-$19\%} & 66.8 & \diffannot{64.0}{$-$2.8} \\
\smallfont{\geminitwofivepro}        & 268.5 & \diffannot{\phantom{0}241.4}{\phantom{0}$-$10\%} & 69.9 & \diffannot{67.5}{$-$2.5} \\
\midrule
{\scriptsize Retrieval (Naive)} \\
\cmidrule(r){1-1}
\smallfont{\claudefoursonnet}         & 238.9 & \diffannot{\phantom{0}299.5}{\phantom{0}$+$25\%}  & 63.6 & \diffannot{62.8}{$-$0.8} \\
\smallfont{\gptfourone}            & 264.0 & \diffannot{\phantom{0}246.1}{\phantom{00}$-$6\%}  & 65.0 & \diffannot{63.7}{$-$1.3} \\
\midrule
{\scriptsize Deep Research} \\
\cmidrule(r){1-1}
\smallfont{\openaideepresearch}      & 268.0 & \diffannot{\phantom{0}595.0}{$+$122\%} & 72.2 & \diffannot{78.9}{$+$6.7} \\
\smallfont{\sonardeepresearch}  & 267.2 & \diffannot{1431.0}{$+$435\%} & 76.2 & \diffannot{85.3}{$+$9.1} \\
\bottomrule
\end{tabular}
\caption{Removing the $L$=250 words length guidance prompt can affect average answer length (words).
Longer answers tend to score higher coverage, because coverage is a recall-based measure.
Analysis is performed on a 225 query subset of \datatest (3 per field).}
\label{table:coverage-vs-length}
\end{table}

\begin{table*}[!t]
\centering
\scriptsize
\setlength{\tabcolsep}{3.5pt}
\begin{tabular}{@{\hspace{0.5em}}lllrr@{\hspace{0.5em}}}
\toprule
\textbf{Item Type} & \textbf{Description} & \textbf{Example} & \textbf{Frequency \%} & \textbf{Error \%} \\
\midrule
Citation     & \scriptsize $X$ is cited &
\makecell[l]{
Does the response cite Kvaskoff et al. (2015) (title: [...])\\ that links endometriosis with elevated cardiovascular risk?}
& 8.3 & 89.3 \\
\midrule
Limitation   & \scriptsize Limitations of $X$ are mentioned &
\makecell[l]{
Does the response address limitations of CTC in detecting\\ 
small polyps and flat adenomas?}
& 2.7 & 52.4 \\
\midrule
Comparison   & \scriptsize $X$ and $Y$ are compared &
\makecell[l]{
Does the response compare reaction rates before and after\\
catalyst saturation occurs?}
& 14.2 & 51.9 \\
\midrule
Example      & \scriptsize Examples of $X$ are mentioned &
\makecell[l]{
Does the response include forage species (e.g., legumes, \\ 
chicory) affecting lamb meat’s fatty acid profile?}
& 11.2 & 46.8 \\
\midrule
Impact       & \scriptsize Cause or impact of $X$ is mentioned &
\makecell[l]{
Does the response mention the preservation of the anterior\\ 
cruciate ligament (ACL) as a benefit of UKA?}
& 15.5 & 46.3 \\
\midrule
Other        & \scriptsize None of the above &
\makecell[l]{
Does the response discuss METEOR’s fragmentation \\
penalty and its role in evaluating word order?}
& 48.1 & 43.6 \\
\bottomrule
\end{tabular}
\caption{
Error rates for different rubric types, measured as the percentage of items not rated as \emph{Completely} covered by the best-performing system (\sonardeepresearch).
Each rubric type provides a description (with $X$ representing a concept, method, or paper being evaluated) along with an illustrated example.
}
\label{table:perplexity-errors}
\end{table*}

\section{LLM Systems Evaluation Results} \label{sec:disc}

\subsection{Response Results}

\paragraph{Results by systems.}
All systems struggle to fully cover rubric items, with no parametric or retrieval augmented systems we evaluated exceeding 70\% coverage (\Cref{table:eval}).
The best-performing system \sonardeepresearch reaches not much more, obtaining 75\% coverage.
These results demonstrate headroom for further improvement.
Win-rates of different systems indicate large performance gaps: systems designed for research synthesis have large advantages over others.
In fact, \sonardeepresearch is estimated to have an 82\% win rate over the next highest rated system, \geminitwofivepro.

We visualize sample responses from each system category in \Cref{fig:parametric_vs_rag} and \Cref{fig:rag_vs_dr}, presenting a case study to show qualitative differences.
Here, the question is: ``What strategies are recent research efforts employing to reduce the wall-clock training time for BERT models? (Date: 2020-10-02)''.
The parametric response covers sensible methods for training optimization (\emph{mixed-precision training}, \emph{gradient accumulation}) addressing content mentioned in rubrics.
However, the response lacks detail about each method.
Naive retrieval locates and explains in detail \citet{trainbigcompress}, describing that \emph{the most effective recent strategies involve training larger models for fewer steps}.
The response hardly addresses prototypical explanations to the query, scoring lower coverage on the rubric.
We further compare production retrieval and deep research for the query, ``How does substrate temperature influence the crystal orientation and quality of AlN (002) thin films during sputtering? (Date: 2020-10-19)''.
While the production retrieval response appears sensible, the deep research answer mentions the same explanations (\emph{atomic mobility}) while uniquely discussing further aspects of temperature influences (\emph{stress evolution}) in detail (\emph{<300\textdegree C [increases] residual stress up to 1.2GPa}).
In general, we find that deep research systems cover prototypical explanations while also dense in breadth and depth, demonstrating several advantages over systems from other categories.

\paragraph{Results by rubric type.}
To identify areas for improvement, we break down coverage of rubric items addressing different types of evaluation criteria.
Each rubric item is lemmatized, using NLTK~\cite{nltk}, and categorized using rules to 6 rubric types (Macro F1$=$.92, without ``Other'').
In \Cref{table:perplexity-errors}, we show examples and the distribution of \sonardeepresearch errors. 
We report the proportion of rubric item types that are not \keyword{Completely} covered: citation-based items can be improved the most (89\%), followed by describing important limitations (52\%) and making comparisons (52\%).
Error breakdowns of all other systems are shown in \Cref{fig:systems_rubric_radar,fig:systems_rubric_heatmap}.

\paragraph{Results by domains.}
Differences between domains are within $\pm$6\ Coverage \% (\Cref{table:eval}).
While differences are small, Health Sciences and Humanities rank as the 2 lowest performing domains: in 10/18 systems for Health Sciences and 18/18 for Humanities.
These trends are suggestive that evaluations focused on Engineering and Physical Sciences, which rank as top-performing domains and are also heavily represented in prior evaluations (\Cref{table:statistics}), provide an incomplete analysis of performance across research domains.

\subsection{Effects of Information Recency, Survey Leakage, and Answer Length}\label{subsec:results-additional}

In this section, we explore biases on Coverage \% due to citing recent information, survey retrieval leakage, and constrained answer lengths.

\begin{figure}[!t]
\centering
\includegraphics[width=\columnwidth]{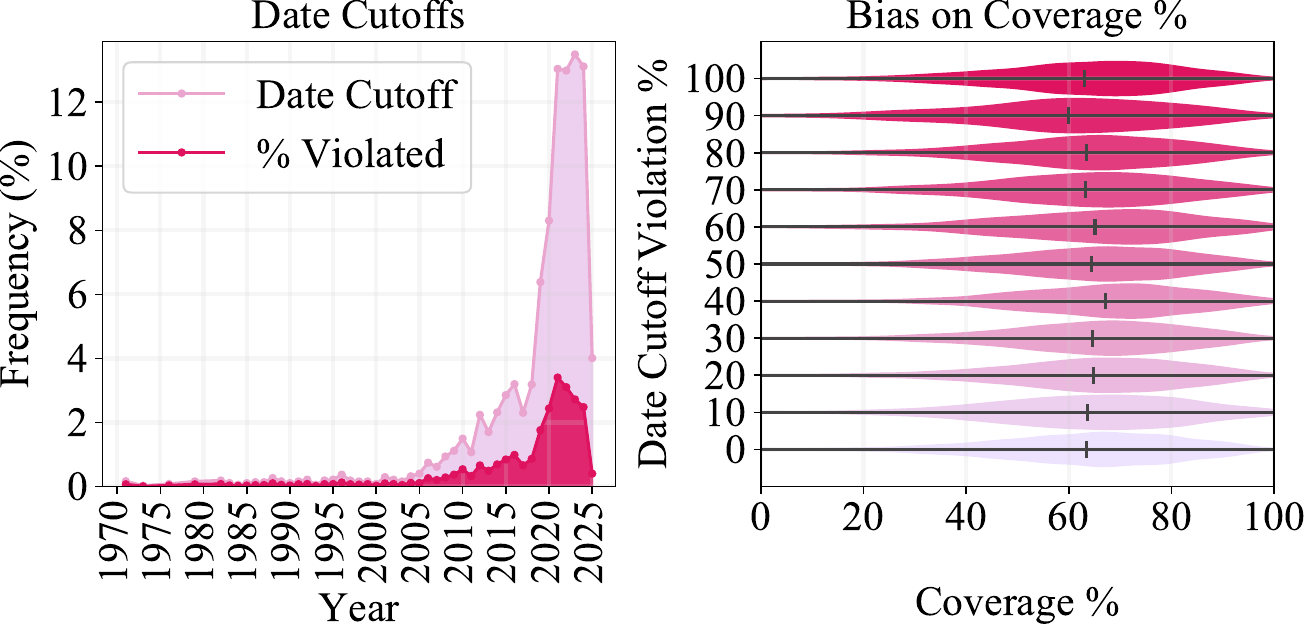}
\vspace{-25pt}
\caption{Rubrics are recent, mostly originating from surveys in the past decade. When cited sources violate date cutoff years, there is little bias on Coverage \%.}
\label{fig:date-violations}
\vspace{-10pt}
\end{figure}

\paragraph{Information recency analysis.}
We examine whether relying on sources published after the distilled survey articles unfairly lowers Coverage \%.
For each answer, we count the \% of cited sources that violate the instructed date cutoff (survey article publication year).
Sources that violate the date cutoff make up $\sim$30\% of the citations (\Cref{fig:date-violations}), suggesting systems do not adhere to date cutoffs stated in instruction text.
Despite high frequency of date violations, we observe that recent sources do not positively or negatively affect Coverage \% in aggregate: Coverage \% is on average $\pm1.3$\% relative to mean coverage when there are no date violations.
While current effects of date violations are small, we recommend monitoring possible information recency biases in future evaluations.

\paragraph{Leakage analysis.}
By distilling rubrics from downloadable survey papers, systems may gain evaluation advantages when retrieving the same survey papers to generate answers, i.e., \textit{leakage}.
To mitigate leakage from affecting evaluation, we restrict source papers from appearing in search retrievals in systems where appropriate control is possible.
We additionally perform post-hoc leakage analysis, because many production systems do not have the option to restrict search retrieval for answer generation.
Leakage is detected by checking whether the survey titles appears in the reference section of the answer.\footnote{To address noise in paper titles, we lowercase text, remove special characters, and search for 6-gram matches.}
We compare Coverage \% in subsets of answers where leakage is and is not detected for top-performing production systems in \Cref{table:agentic-leakage} and all systems in \Cref{table:agentic-leakage-full}.

Leakage advantages are small and inconsistent: in 13 systems exhibiting leakage, Coverage \% increased for answers with leakage in 8/13 systems, where Coverage \% actually decreased in the other 5/13 systems.
In aggregate, we observe increases of $\sim$1.1\% when distilled surveys are retrieved to generate answers.
These small differences assuage our concerns of leakage advantages in current systems, but evaluations of future systems should continue verifying that leakage advantages are minimal and restrict source surveys from search results whenever possible.

\begin{figure}[!t]
\centering
\includegraphics[width=0.98\columnwidth]{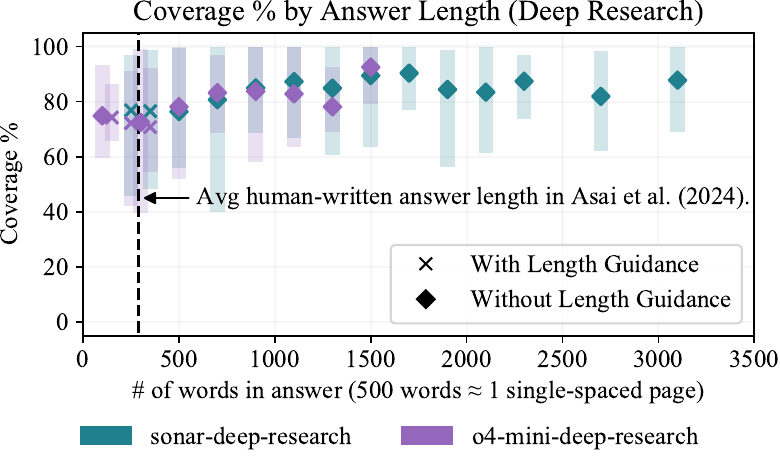}
\vspace{-10pt}
\caption{Coverage \% increases with answer length, up until $\sim$2K words (about 4 pages of text).}
\label{fig:perplexity-full-length}
\end{figure}

\begin{table}[t]
\centering
\scriptsize
\setlength{\tabcolsep}{2pt}
\renewcommand{\arraystretch}{0.98}
\newcommand{\pair}[2]{#1\,/\,#2} 

\begin{tabular}{l r r r r}
\toprule
\textbf{System} & \textbf{L\%} & \multicolumn{3}{c}{\textbf{Coverage \%}} \\
\cmidrule{3-5}
&& $\neg$~Leaked & Leaked & $\Delta$\\
\midrule
\anthropicwebsearch  & 30 & \textbf{71.27} & \diffannot{68.30}{$-$3.0} \\
\geminiwebsearch     & 2  & 66.91 & \diffannot{\textbf{68.54}}{$+$1.6} \\
\openaideepresearch  & 28 & \textbf{74.82} & \diffannot{71.85}{$-$3.0} \\
\sonardeepresearch   & 21 & 74.46 & \diffannot{\textbf{75.51}}{$+$1.1}\\
\bottomrule
\end{tabular}

\caption{Production systems often cite the related survey as a source, at 20-30\% leakage (L\%). 
However, Coverage \% roughly stays the same with leakage (Leaked) and without leakage ($\neg$~Leaked).}
\label{table:agentic-leakage}
\end{table}

\paragraph{Unconstrained answer length analysis.}
We remove the 250 word constraint from the task instruction text to further explore how Coverage~\%, a recall-based metric, increases when we do not control for answer length during evaluations.
We select from the top performers for parametric, naive retrieval, and deep research systems by Coverage~\% and additionally generate answers for 3 queries per field (225 total queries).
Average statistics are shown in \Cref{table:coverage-vs-length}, and trends for deep research systems are visualized in \Cref{fig:perplexity-full-length} and for the remaining systems in \Cref{fig:coverage-vs-length}.

Unconstrained length analyses reveal that many systems naturally generate answers of close to 250 words, with no substantial differences to answer length or Coverage \% (\Cref{fig:coverage-vs-length}).
However, deep research systems tend to generate longer answers, up to 3K words (about 6 pages of text) per answer.
By contrast, expert-written answers lengths for similar topics average 289 words in \citet{openscholar}.
Coverage \% increases with answer length up to $\sim$2K words (about 4 pages of text), as shown in~\Cref{table:coverage-vs-length}, plateauing at 85\%  for \sonardeepresearch and 79\% for \openaideepresearch.
These trends suggest that deep research systems are able to convey more helpful information while trading off concise and focused answers.

\section{Related Work}\label{sec:relwork}
We fit into a broad body of work trying to improve benchmarking for LLMs. This includes 
multi-domain works~\cite{hendrycks2020measuring,wang2025mmlu}, 
challenge sets~\cite{rein2024gpqa,phan2025humanity,wolfson2025monaco} 
and expert annotations~\cite{expertqa,malaviya2025dolomites}. 
Model development for survey creation has used survey data material~\cite{goldfarb2020scalelitreview,kasanishi2023scireviewgen,agarwal2024litllm,wang2024autosurvey}. 
Below we outline the key differences between \dataset and others.

\paragraph{Manually crafted scholarly benchmarks.}
Manual curation of benchmarks for scholarly QA has encountered practical challenges for creating large, diverse, and complex datasets.  
QASPER~\cite{qasper} and QASA~\cite{qasa} limit their focus on queries that can be answered within a single paper.
KIWI~\cite{kiwi} is built from questions derived by researchers on related work sections, and ScholarQABench~\cite{openscholar} recruited researchers to write questions from scratch.
Both efforts have multi-document queries but are smaller than \dataset because of the challenges of recruiting researchers.
Concurrently, SciArena~\cite{sciarena} leverages community contributions to collect queries and preferences; however, it remains smaller in size and coverage than \dataset and lacks evaluation rubrics.

\paragraph{Auto-generated scholarly benchmarks.}
Automatically generated scholarly QA benchmarks trade-off scale for complexity and naturalness.
DeepScholar-Bench~\cite{patel2025deepscholar} uses 63 CS papers to compare machine-written and human-written related works.
SciQA~\cite{sciqa} is template generated and focuses on questions generated from knowledge graphs.
PubMedQA~\cite{pubmedqa} generates queries from the abstracts of PubMed articles, but limits to yes or no questions.
SciDQA~\cite{scidqa} extracts 188k queries asked during peer review on OpenReview but limits queries to extractions about the paper being reviewed.
\dataset achieves more abstractive queries and evaluation materials by focusing on surveys but is smaller than SciDQA. 

\paragraph{Evaluating long-form answers and rubrics.}
A central problem for long-form answer evaluation is a large space of possible correct answers.
Reference answers can be difficult to use, and token based measures like ROUGE~\cite{lin2004rouge} are gameable~\cite{krishna2021hurdles}.
Recent efforts rely on direct evaluator LLMs~\cite{openscholar, alpacaeval} but may inherit self-preference~\cite{llm-judge-self-bias-panickssery,llm-judge-self-bias-watoka}, length biases~\cite{alpacaeval} and may be inaccurate for research queries.
\dataset shows these judges can be effective.
\emph{Rubric-based evaluations} that decompose judgment into nuanced criteria~\cite{sawadaarb, g-eval} are promising alternatives to direct evaluator LLMs.
Manually curated rubrics are often task-specific and small scale~\cite{openscholar, starace2025paperbench, qin2024infobench}. 
WildBench~\cite{lin2024wildbench} creates query-specific checklist rubrics from parametric memory at scale. 
EvalAgent~\cite{wadhwa2025evalagent} discovers query-specific rubrics with online search. 
Our rubrics leverage discovery from surveys, aligning them to expert-written material.
\section{Conclusions}

We introduce \dataset, a resource distilled from survey articles for large scale and multi-field evaluations of research synthesis.
We leverage \dataset to benchmark 18 systems showing each has headroom for improvement.
The highest-ranking deep research system, which achieves 82\% win rate over the next system, only fully addresses fewer than 11\% of items addressing citations, 48\% of items describing limitations, and 49\% of items asking about comparisons.
\dataset can be expanded with newly written surveys as to increase coverage of new topics.
\section*{Limitations}

While \dataset reduces reliance on experts, expert involvement is not eliminated.
We depend on experts and expert-written articles to build and validate \dataset. 
To the extent that we can, we recruit experts to validate data, but not all fields are equally validated.
Additionally, open access to research articles is limited, hindering exhaustive coverage of the literature.
Rubrics distilled from these articles may not fully cover important criteria; we leave synthesis of larger and comprehensive rubrics for future work.
Many recently released deep research systems involve limited access; their omission can distort leaderboard rankings and obfuscate important failure modes.
\section*{Acknowledgements}
This research was developed with funding from the Defense Advanced Research Projects Agency's (DARPA) SciFy program (Agreement No. HR00112520300) and the National Science Foundation Graduate Research Fellowship under Grant No. DGE-2236662.
The views expressed are those of the authors and do not reflect the official policy or position of the sponsors.
We would like to thank the UPenn NLP group and Material Science and Engineering, Biomedical Engineering, Computer and Information Science, Genetics, Linguistics, Physics and Astronomy, Criminology, and Psychology departments for generously participating in our annotation.
We thank Zack Ives for aiding university wide recruitment.  

\bibliographystyle{style/acl_natbib}
\bibliography{refs}

\clearpage
\newpage
\appendix
\section{Additional Details of Dataset Pipeline}
\label{appendix:dataset_building}

Below, we describe the data pipeline (\Cref{sec:dataset}) details.

\subsection{Query generation from survey articles}

\paragraph{Curating a list of top publication venues for 257 research fields.}
We identify the top-20 publishing venues for each of 257 research fields listed on Google Scholar Metrics (total venues $=$ 4634).
The aggregate counts are presented in \Cref{table:field-venues}.

\paragraph{Retrieving survey articles from each publication venue.}
We retrieve articles from three datastores: Crossref, the Semantic Scholar API, and S2ORC~\cite{s2orc}.
Articles are queried with the following keywords:
\keyword{survey}, 
\keyword{literature review}, 
\keyword{a review}, 
\keyword{an overview}, and
\keyword{meta-analysis}.
The article retrieval yields \statsurveysretrieved article weblinks, where \statsurveysdownloaded are downloadable full-text articles.

\paragraph{Removing non-survey articles.}
We use \pipelinemodel to classify between actual literature reviews and articles that mention a keyword like \keyword{survey} in its title.
Classifier metrics are F1$=$.80, Prec$=$.87, Rec$=$.75 on an author-annotated validation set.
The final yield is \statsurveyslitreview survey articles across 254 research fields (3 research fields do not yield survey articles using our method).

\paragraph{Selecting sections from survey articles.}
We apply a sequence of filters to select sections (total $=$ \statsectionstotal, yield $=$ \statsectionsmincitation) from survey articles:
\begin{itemize}
\item \textbf{Title passes keyword blacklist}: We ignore abstract, introduction, or other summary sections, removing any sections that contain the following words: \keyword{question},
\keyword{survey},
\keyword{abstract},
\keyword{introduction},
\keyword{contribution},
\keyword{related},
\keyword{result},
\keyword{discussion},
\keyword{conclusion},
\keyword{limitation},
\keyword{appendix},
\keyword{appendices},
\keyword{appendixes},
\keyword{supplementary},
\keyword{supplemental},
\keyword{supplement},
\keyword{material},
\keyword{acknowledgement},
\keyword{future},
\keyword{direction},
\keyword{summary},
\keyword{suggestion},
\keyword{table},
\keyword{tab.},
\keyword{tbl.},
\keyword{figure},
\keyword{fig.}, and
\keyword{plot}.

\item \textbf{Section length is not too short ($\geq$~3 sentences, $\geq$~800 characters) and not too long ($\leq$~300K characters)}: 
We apply a basic length filter to ensure that queries are generated from substantial, but focused, sections.

\item \textbf{Minimum number of in-text citations ($\geq$~3)}: 
We select sections that are well grounded in the literature, where each section must have 3 or more citations.
\end{itemize}

\begin{table}[!t]
\footnotesize
\centering
\setlength{\tabcolsep}{4.5pt}
\begin{tabular}{lrr}
\toprule
\textbf{Field} & \textbf{Fields} & \textbf{Venues} \\
\midrule
Business, Economics, \& Management      & 16 & 285 \\
Chemical \& Material Sciences           & 17 & 318 \\
Physics \& Mathematics                  & 21 & 370 \\
Humanities, Literature, \& Arts         & 26 & 481 \\
Life Sciences                           & 30 & 509 \\
Social Sciences                         & 31 & 526 \\
Engineering \& Computer Science         & 50 & 925 \\
Health \& Medical Sciences              & 66 & 1220 \\
\midrule
Total & 257 & 4634\\
\bottomrule
\end{tabular}
\caption{Aggregated numbers for top-20 publishing venues in Google Scholar categories.}
\label{table:field-venues}
\end{table}

\newcommand{\subdomain}[2]{\textbf{\emph{#1}} & \textbf{\emph{#2}}\\[1.417mm]%
}

\begin{table*}[!t]
\centering
\scriptsize

\begin{minipage}[t]{0.5\textwidth}
\vspace{0pt}
\centering
\begin{tabularx}{\linewidth}{Xr}
\toprule
\textbf{\footnotesize Research Domain and Field} & \textbf{\footnotesize \# Queries} \\
\startgraymidrule
\textbf{Health Sciences \& Medicine}~\iconhealth & \textbf{7454}\\
\stopgraymidrule

\subdomain{Specialized Medicine}{3968}
Surgery & 673 \\
Pharmacology \& Pharmacy & 630 \\
Dentistry & 380 \\
Oncology & 369 \\
Veterinary Medicine & 332 \\
Pediatric Medicine & 294 \\
Emergency Medicine & 232 \\
Psychiatry & 208 \\
Dermatology & 189 \\
Ophthalmology \& Optometry & 144 \\
Developmental Disabilities & 138 \\
Anesthesiology & 137 \\
Otolaryngology & 101 \\
\makecell[l]{Nuclear Medicine, Radiotherapy\\
\hspace{1em}\& Molecular Imaging} & 
89 \\
Neuroscience & 52 \\

\cmidrule(lr){1-2}
\subdomain{Preventive Medicine}{2151}
Nutrition Science & 1410 \\
Physical Education \& Sports Medicine & 327 \\
Epidemiology & 156 \\
Alternative \& Traditional Medicine & 140 \\
Tropical Medicine \& Parasitology & 118 \\

\cmidrule(lr){1-2}
\subdomain{Internal Medicine \& Chronic Diseases}{1070}
Toxicology & 196 \\
Communicable Diseases & 182 \\
Endocrinology & 124 \\
Diabetes & 118 \\
Hematology & 110 \\
Urology \& Nephrology & 84 \\
Pulmonology & 76 \\
Rheumatology & 76 \\
Gastroenterology \& Hepatology & 54 \\
Vascular Medicine & 50 \\

\cmidrule(lr){1-2}
\subdomain{Reproductive Medicine}{265}
Reproductive Health & 95 \\
Gynecology \& Obstetrics & 91 \\
Pregnancy \& Childbirth & 79 \\

\startgraymidrule
\textbf{Life \& Earth Sciences}~\iconlife & \textbf{4928}\\
\stopgraymidrule

\subdomain{Life Sciences}{3528}
Agronomy \& Crop Science & 780 \\
Forests \& Forestry & 674 \\
Biochemistry & 588 \\
Microbiology & 413 \\
Botany & 294 \\
Animal Biology \& Behavior & 224 \\
Cell Biology & 190 \\
Marine Sciences \& Fisheries & 188 \\
Mycology & 94 \\
Ecology & 83 \\

\bottomrule
\end{tabularx}
\end{minipage}%
\hfill
\begin{minipage}[t]{0.5\textwidth}
\vspace{0pt}
\centering
\begin{tabularx}{\linewidth}{Xr}
\toprule
\textbf{\footnotesize Research Domain and Field (continued)} & \textbf{\footnotesize \# Queries} \\
\midrule

\subdomain{Earth Sciences}{1400}
Sustainable Energy & 466 \\
Hydrology \& Water Resources & 400 \\
Geology & 358 \\
Geochemistry \& Mineralogy & 107 \\
Atmospheric Sciences & 69 \\

\startgraymidrule
\textbf{Engineering \& Computer Science}~\iconengineering & \textbf{4676}\\
\stopgraymidrule

\subdomain{Engineering}{3417}
Materials Engineering & 1944 \\
Manufacturing \& Machinery & 329 \\
Electrical Engineering & 274 \\
Biomedical Engineering & 266 \\
Architecture & 240 \\
Environmental \& Geological Engineering & 194 \\
Ocean \& Marine Engineering & 88 \\
Mechanical Engineering & 82 \\

\cmidrule(lr){1-2}
\subdomain{Computer Science}{1259}
Artificial Intelligence & 509 \\
Signal Processing & 274 \\
Natural Language Processing & 230 \\
Computer Vision \& Pattern Recognition & 193 \\
Computing Systems & 53 \\

\startgraymidrule
\textbf{Physical Sciences}~\iconphysical & \textbf{2544}\\
\stopgraymidrule

\subdomain{Material Sciences}{1320}
Polymers \& Plastics & 1094 \\
Composite Materials & 226 \\
\cmidrule(lr){1-2}
\subdomain{Chemistry}{1153}
Analytical Chemistry & 749 \\
Chemical Kinetics \& Catalysis & 212 \\
Dispersion Chemistry & 134 \\
Crystallography \& Structural Chemistry & 58 \\
\cmidrule(lr){1-2}
\subdomain{Physics}{71}
Physics & 71 \\

\startgraymidrule
\textbf{Social Sciences}~\iconsocial & \textbf{1395}\\
\stopgraymidrule

Sustainable Development & 600 \\
Public Health & 566 \\
Education & 229 \\

\startgraymidrule
\textbf{Humanities}~\iconhumanities & \textbf{362}\\
\stopgraymidrule

\subdomain{Cognitive Studies}{301}
Psychology & 250 \\
Language \& Linguistics & 51 \\

\cmidrule(lr){1-2}
\subdomain{Anthropology}{61}
Anthropology & 61 \\

\startgraymidrule
\textbf{Economics}~\iconbusiness & \textbf{55}\\
\stopgraymidrule
Economics & 55 \\
\bottomrule
\end{tabularx}
\end{minipage}%
\hfill
\caption{The number of queries for each of 75 fields, by research domain, in \dataset.}
\label{table:field-query-breakdown}
\end{table*}

\begin{table*}[!t]
\scriptsize
\centering
\setlength{\tabcolsep}{7pt}
\renewcommand{\arraystretch}{1.28}

\begin{tabular}{l p{6.9cm} p{6.9cm}}
\toprule
\textbf{Type} & \textbf{\gptfouronemini} & \textbf{\gptfourone} \\

\specialrule{0.06em}{\aboverulesep}{0pt} 
\rowcolor{lavender!20}
\rule[-2.8\aboverulesep]{0pt}{\dimexpr\aboverulesep + \normalbaselineskip + \belowrulesep\relax}\null
\query & What is the rule-based approach in machine translation, and what are its benefits and limitations?
 & 
How do rule-based machine translation systems use linguistic analysis and dictionaries to carry out translations?\\

\rowcolor{gray!20}
$R_1$ & 
Does the response provide a detailed explanation of the linguistic analysis process involved in the Rule-based Approach, including morphology, syntax, and lexical semantics? & Does the response explain how linguistic analysis at multiple levels (morphology, syntax, lexical semantics) is used in rule-based machine translation?\\

\rowcolor{gray!10}
$R_2$ & Does the response discuss the flexibility of the Rule-based Approach in adapting to new language constructs by updating the dictionary? & Does the response describe the use of bilingual dictionaries for mapping source language words to target language equivalents? \\

\rowcolor{gray!20}
$R_3$ & Does the response highlight the advantage of the Rule-based Approach in not requiring as many parallel sentence pairs as Neural Machine Translation? & Does the response discuss how RBMT systems handle word ambiguities and parts of speech during translation?
\\

\rowcolor{gray!10}
$R_4$ & Does the response mention the use of dictionaries and expert knowledge in establishing grammar rules for the Rule-based Approach? & Does the response explore in depth how linguistic analysis at multiple levels (morphology, syntax, lexical semantics) is performed and integrated in rule-based machine translation systems?\\

\rowcolor{gray!20}
$R_5$ & Does the response include a definition or explanation of what a Rule-based Approach in machine translation is? & 
Does the response provide a detailed explanation of how bilingual dictionaries are structured and utilized within the translation process?\\

\rowcolor{gray!10}
$R_6$ & Does the response mention specific benefits of using a Rule-based Approach in machine translation? & 
Does the response provide a detailed explanation of how linguistic analysis components are integrated into the translation process?\\

\rowcolor{gray!20}
$R_7$ & Does the response highlight any limitations or challenges associated with the Rule-based Approach in machine translation? & Does the response explore in depth the role and structure of dictionaries used in rule-based machine translation?\\

\rowcolor{gray!10}
$R_8$ & Does the response cite the paper by Scott and Barreiro (2009) (title: OpenLogos MT and the SAL representation language) that provides insights into the linguistic analysis and dictionary mapping process in Rule-based Machine Translation? & Does the response cite papers like Kay (1980) (title: The Proper Place of Men and Machines in Language Translation) that discuss the integration of dictionaries with linguistic rules for accurate translation?  \\

\arrayrulecolor{gray!10}\specialrule{\aboverulesep}{0pt}{0pt}%
\arrayrulecolor{black}%
\specialrule{\heavyrulewidth}{0pt}{\belowbottomsep} 
\end{tabular}

\caption{Comparisons between when pipeline models are used as generators. Generally, queries and rubrics are comparable in quality, where those from \gptfourone are slightly more specific than those from \gptfouronemini.}
\label{table:pipeline-model-comparison}
\end{table*}

\paragraph{Generating queries from survey sections.}
We extract a date cutoff~\datecutoff from each section's article metadata, and we use \pipelinemodel to parse the section content into a hierarchical summary to generate an initial query and initial reference answer \queryanswerinitialtuple.
To select high quality queries, all queries (total $=$ \statsectionsmincitation, yield $=$ \statqueriestruncated) are filtered:

\begin{itemize}
    \item \textbf{Initial query is self-contained}: \pipelinemodel assigns \queryinitial a self-containment score from 1 to 10, where higher scores indicate more self-containment. We remove queries with $<$~7. 
    \item \textbf{Initial query has low answer variability}: \pipelinemodel assigns \queryinitial an answer variability score from 1 to 10, where higher scores indicate likely answer variability (e.g., expert disagreement or subjectivity). We remove queries with $>$~4. 
    \item \textbf{(Rephrasing Step) Query and reference answer cohesion}: \pipelinemodel rephrases \queryanswerinitialtuple to queries~\query and reference answers~\mockanswer to improve their cohesion. We do not remove any queries at this step.
    \item \textbf{Query does not contain a citation}: \pipelinemodel detects whether a citation is in the query~\query.
    \item \textbf{Reference answer length is long enough ($\geq$~800 characters)}: Reference answers~\mockanswer substantiate a semi open-ended query. 
    \item \textbf{Final query is self-contained}: After rephrasing, some queries are still not standalone, so we apply a keyword-based method to remove non-standalone queries.
    Keywords include past tense auxiliary verbs (\keyword{did}, \keyword{was}, \keyword{were}) or words in referring expressions (\keyword{questionnaire}, \keyword{literature}).
    
    \item \textbf{Final query is aligned with field}:
    Venues can be multidisciplinary, and therefore some queries are unrelated to the field those venues are mapped to.
    To ensure field alignment, \pipelinemodel clasifies and removes queries that are unrelated to the field.
    \vspace{1em}
    
    \item \textbf{Queries are from a field with $\geq$~50 queries}: Queries should cover important aspects of a field, which is difficult to accomplish if there are not enough queries. We set 50 to be the minimum query count to call a field sufficiently covered by queries, which leaves \statqueriestruncated queries~\query that cover \statfields research fields.
\end{itemize}

\subsection{Rubric generation from survey articles}

We create rubrics with \pipelinemodel.
To diversify rubric item types, three generation prompts are used---information, depth, and citation.
These rubric types are intended to diversify the types of rubric items that are generated, whereas rubric items have been categorized post-hoc to more granular labels for analysis.
\begin{itemize}
\item \textbf{Information-based rubric item}: A binary yes/no question asking whether an answer addresses a specific statement, finding, opinion, or comparison. 
Rubric items of this category allow for broad analysis of quality.

\emph{Example: Does the response address the specific benchmarks where auto-regressive LMs outperform bi-directional encoders?}

\item \textbf{Depth-based rubric item}: 
A binary yes/no question asking whether an answer elaborates or explains a topic.
Rubric items of this category allow measuring analysis, discussion, and explanation in answers.

\emph{Example: Does the response elaborate on the hybrid approach of GRIT in unifying auto-regressive and bi-directional features?}

\item \textbf{Citation-based rubric item}: A binary yes/no question asking whether or not an answer cites a specific study. Rubric items of this category evaluate answer grounding in the literature.

\emph{Example: Does the response cite papers such as Wang et al. (2023) (title: Improving text embeddings with large language models) and the MTEB paper (title: Massive Text Embedding Benchmark) that show the performance of auto-regressive LMs surpassing bi-directional encoders in retrieval tasks?}
\end{itemize}

In addition to automatically generated rubrics, we explore generic rubrics using evaluation criteria from related work~\cite{openscholar,sciarena}.
Generic rubric items are in \Cref{table:generic-rubric}.

\paragraph{Remapping research fields and domains.}
For the purposes of evaluation, some fields are overly specific (e.g., Wood Science \& Technology) or overly general (e.g., Health \& Medical Sciences (general)).
Additionally, the Physics \& Mathematics branch does not yield enough queries to sufficiently represent a domain.
For these reasons, we redistribute data to match a more intuitive hierarchical structure.
In total, we merge 170 of 257 fields and introduce 7 more: Animal Biology \& Behavior, Biomedical Engineering, Electrical Engineering, Management, Mathematics, Physics, and Neuroscience.
These changes result in a hierarchy covering 94 fields and 7 domains.
Our pipeline is able to sufficiently cover 75 of 94 fields (where each of the 75 fields has $\geq$~50 queries) using the available datastores.
A full list of fields is in tabulated in \Cref{table:field-query-breakdown}.
\begin{table}[!t]
\centering
\footnotesize
\begin{tabular}{lrrr}
\toprule
\textbf{Model} & \textbf{Queries} & \textbf{Rubrics} & \textbf{Total} \\
\midrule
\gptfouronemini & \$ \phantom{0}349.65 & \$ 133.65  & \$ \phantom{0}483.30 \\
\gptfourone     & \$ 1748.25 & \$ 668.25 & \$ 2416.50 \\
\gptfouro       & \$ 2185.31 & \$ 835.31 & \$ 3020.62 \\
\bottomrule
\end{tabular}
\caption{Cost comparison for different pipeline models for the full generation of \dataset. All numbers are in batched inference pricing (2$\times$ cheaper).}
\label{table:model_costs}
\end{table}

\section{Pipeline Model Analysis}
\label{sec:pipeline-model-analysis}

In this section, we discuss the performances and potential costs of different models \pipelinemodel used in the pipeline to create \dataset.
We consider three state-of-the-art models at the time of consideration in \Cref{table:model_costs}.

\paragraph{\pipelinemodel as a classifier.}
We use \pipelinemodel as a classifier for quality filtering, such as removing non-literature reviews, ambiguous queries, open-ended queries, and removing queries with citations.
For each filtering task, we find that the smallest model, \gptfouronemini, obtains at least .80 F1 compared against author annotations.

\paragraph{\pipelinemodel as a generator.}
We show sample outputs from using \gptfouronemini and \gptfourone in \Cref{table:pipeline-model-comparison}.
We find that queries and rubrics generated by the two models are comparable in quality and content.
Queries generated by \gptfourone are more specific, such as asking explicitly how linguistics and dictionaries are applied in rule-based machine translation; by contrast, \gptfouronemini asks for an open-ended explanation of rule-based machine translation with benefits and limitations.
These differences propagate to the rubrics, with those from \gptfourone judging by discussion of word ambiguities and parts-of-speech whereas those from \gptfouronemini judge by data efficiency in rule-based methods ($R_3$).
Overall, we find content similarities between model outputs with small differences in the fine-grained evaluation criteria.

\paragraph{Costs.}
Costs for batched inference of \pipelinemodel are in \Cref{table:model_costs}.
Notably, \gptfourone is $\sim$\$2K more expensive than \gptfouronemini.
Given performance similarities, we choose \gptfouronemini for scaling out \dataset.

\begin{table}[!t]
\scriptsize
\centering
\setlength{\tabcolsep}{4pt}
\renewcommand{\arraystretch}{0.98}

\begin{tabular}{l p{5.5cm}}
\toprule
\textbf{Type} & \textbf{Generic Rubric Item} \\
\midrule
Correctness & Does the response make factually correct statements? \\
Citations & Does the response cite relevant papers? \\
Coverage & Does the response provide sufficient coverage and amount of information? \\
Relevance & Does the response stay on topic and maintain a clear focus? \\
Organization & Does the response have an organized structure? \\
Usefulness & Does the response fulfill information needs? \\
Attribution & Does the response make correct citation-to-claim attributions? \\
Examples & Does the response include relevant examples? \\
\bottomrule
\end{tabular}

\caption{The generic rubric baseline, sampled from ScholarQABench~\cite{openscholar} and SciArena~\cite{sciarena} criteria. ``Examples'' is manually written to fill the rubric to 8 items.}
\label{table:generic-rubric}
\end{table}
\begin{table*}[t]
\centering
\scriptsize
\setlength{\tabcolsep}{2.5pt}
\begin{tabular}{lllr}
\toprule
\textbf{System} &  \textbf{Model Card} & \textbf{Additional settings} & \textbf{Cost} \\
\startgraymidrule
Parametric & & &\\
\stopgraymidrule

{\scriptsize\iconmeta~\llamathreethreeseventyb} & \texttt{llama-3.3-70b-instruct} & & \$\phantom{000}5 \\
{\scriptsize\iconanthropic~\claudefoursonnet} & \texttt{claude-sonnet-4-20250514} & & \$\phantom{00}24 \\
{\scriptsize\iconopenai~\gptfourone} & \texttt{gpt-4.1} & & \$\phantom{00}15 \\
{\scriptsize\iconqwen~\qwenthreethirtytwob} & \texttt{qwen3-32b} & & \$\phantom{000}5 \\
{\scriptsize\icongoogle~\geminitwofivepro} & \texttt{gemini-2.5-pro-preview-06-05} & & \$\phantom{00}15 \\

\startgraymidrule
Retrieval (Naive) & & &\\
\stopgraymidrule
{\scriptsize\iconallenai~\openscholareightb} & \texttt{llama-3.1\_openscholar-8b} & \makecell[l]{\texttt{[Embed] openscholar-retriever}\\\texttt{[Reranker] openscholar-reranker}} & \$\phantom{000}0 \\
{\scriptsize\iconqwen~\qwenthreethirtytwob} & \texttt{qwen3-32b} & \texttt{[Embed] gte-qwen-2-7b-instruct} & \$\phantom{00}10 \\
{\scriptsize\icongoogle~\geminitwofivepro} & \texttt{gemini-2.5-pro-preview-06-05} & \texttt{[Embed] text-embedding-004} & \$\phantom{00}35 \\
{\scriptsize\iconanthropic~\claudefoursonnet} & \texttt{claude-sonnet-4-20250514} & \texttt{[Embed] voyage-3-large} & \$\phantom{0}218 \\
{\scriptsize\iconopenai~\gptfourone} & \texttt{gpt-4.1} & \texttt{[Embed] text-embedding-3-large} & \$\phantom{0}151 \\

\startgraymidrule
Retrieval (Production) & & &\\
\stopgraymidrule
{\scriptsize\iconperplexity~\sonar} & \texttt{sonar} & & \$\phantom{00}19 \\
{\scriptsize\iconallenai~\openscholareightbagentic} & \texttt{llama-3.1\_openscholar-8b} & & \$\phantom{0}100 \\
{\scriptsize\iconperplexity~\sonarreasoning} & \texttt{sonar-reasoning} & & \$\phantom{00}38 \\
{\scriptsize\iconopenai~\openaiwebsearch} & \texttt{gpt-4o-search-preview} & \texttt{search\_context\_size=medium} & \$\phantom{0}150 \\
{\scriptsize\icongoogle~\geminiwebsearch} & \texttt{gemini-2.5-pro-preview-06-05} & \texttt{dynamic\_threshold=0.3} & \$\phantom{00}30 \\
{\scriptsize\iconanthropic~\anthropicwebsearch} & \texttt{claude-sonnet-4-20250514} & \makecell[l]{\texttt{tools.type=web\_search\_20250305}\\\texttt{max\_uses=3}}  & \$\phantom{0}450 \\

\startgraymidrule
Deep Research & & &\\
\stopgraymidrule
{\scriptsize\iconopenai~\openaideepresearch} & \texttt{o4-mini-deep-research} & \makecell[l]{\texttt{tools.type=web\_search\_preview}\\\texttt{reasoning.summary=auto}} & \$4200 \\
{\scriptsize\iconperplexity~\sonardeepresearch} & \texttt{sonar-deep-research} & & \$1500 \\
\bottomrule
\end{tabular}
\caption{\textbf{LLM System Configuration and Cost.} Default setting: \texttt{temperature=0}. Cost is for running on the full test set of \dataset (3750 queries). 
Organizations and related works: \iconmeta~Meta~\cite{grattafiori2024llama}, \iconanthropic~Anthropic~\cite{claude}, \iconopenai~OpenAI~\cite{gpt4,openaiIntroducingDeep}, \iconqwen~Alibaba~\cite{yang2025qwen3}, \icongoogle~Google~\cite{geminitwofive,geminiGeminiDeep}, \iconallenai AI2~\cite{openscholar}, \iconperplexity~Perplexity~\cite{perplexity2025deepresearch}.
}
\label{table:system-config}
\end{table*}

\begin{figure}[!t]
\centering
\includegraphics[width=\columnwidth]{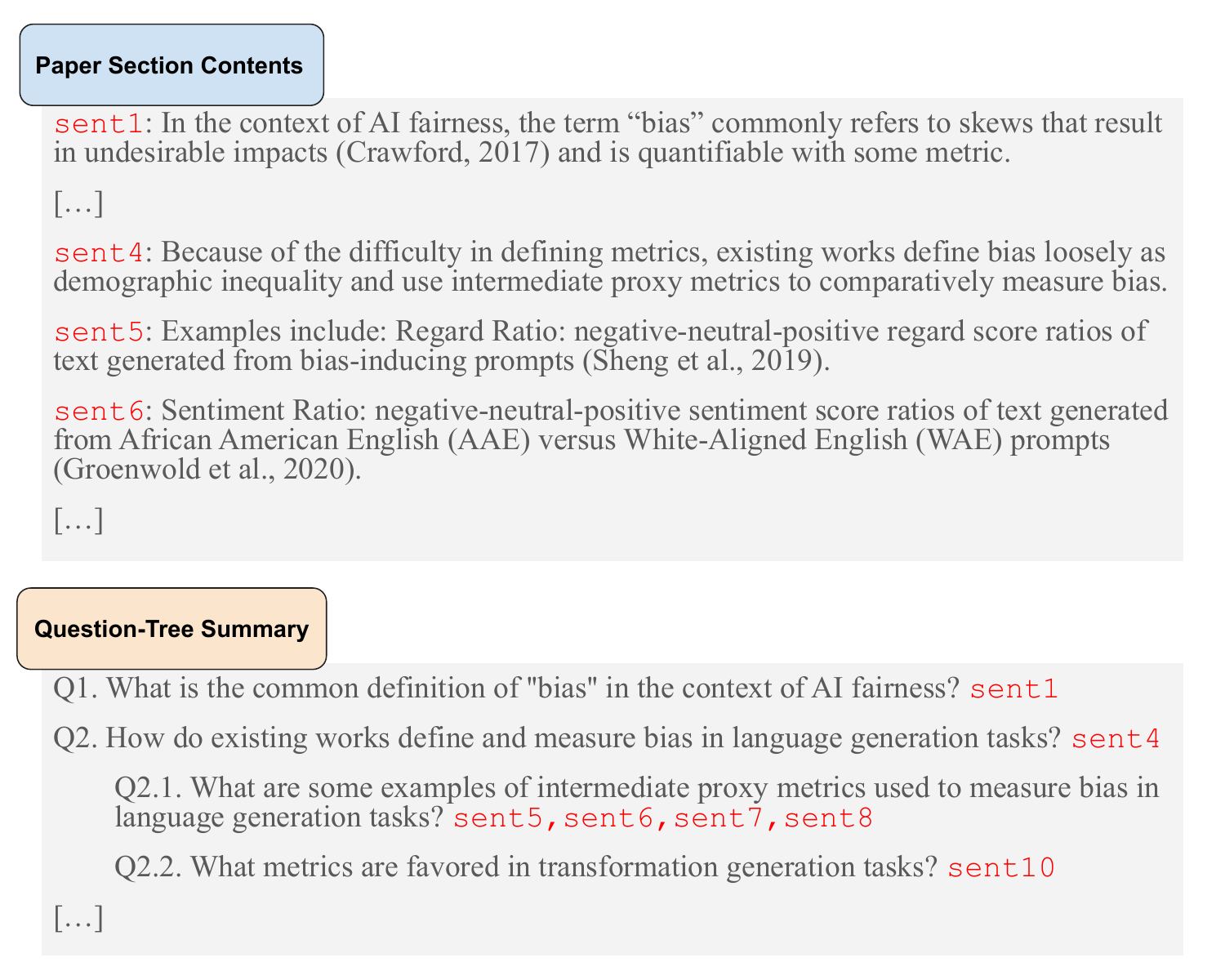}
\vspace{-1em}
\caption{Hierarchical summary of section content, motivated by hierarchical summarization~\cite{christensen2014hierarchical} and question-under-discussion parsing~\cite{benz2017questions,wu2023qudeval}.
}
\label{fig:question_tree}
\end{figure}
\begin{figure*}[!t]
\centering
\includegraphics[width=.82\textwidth]{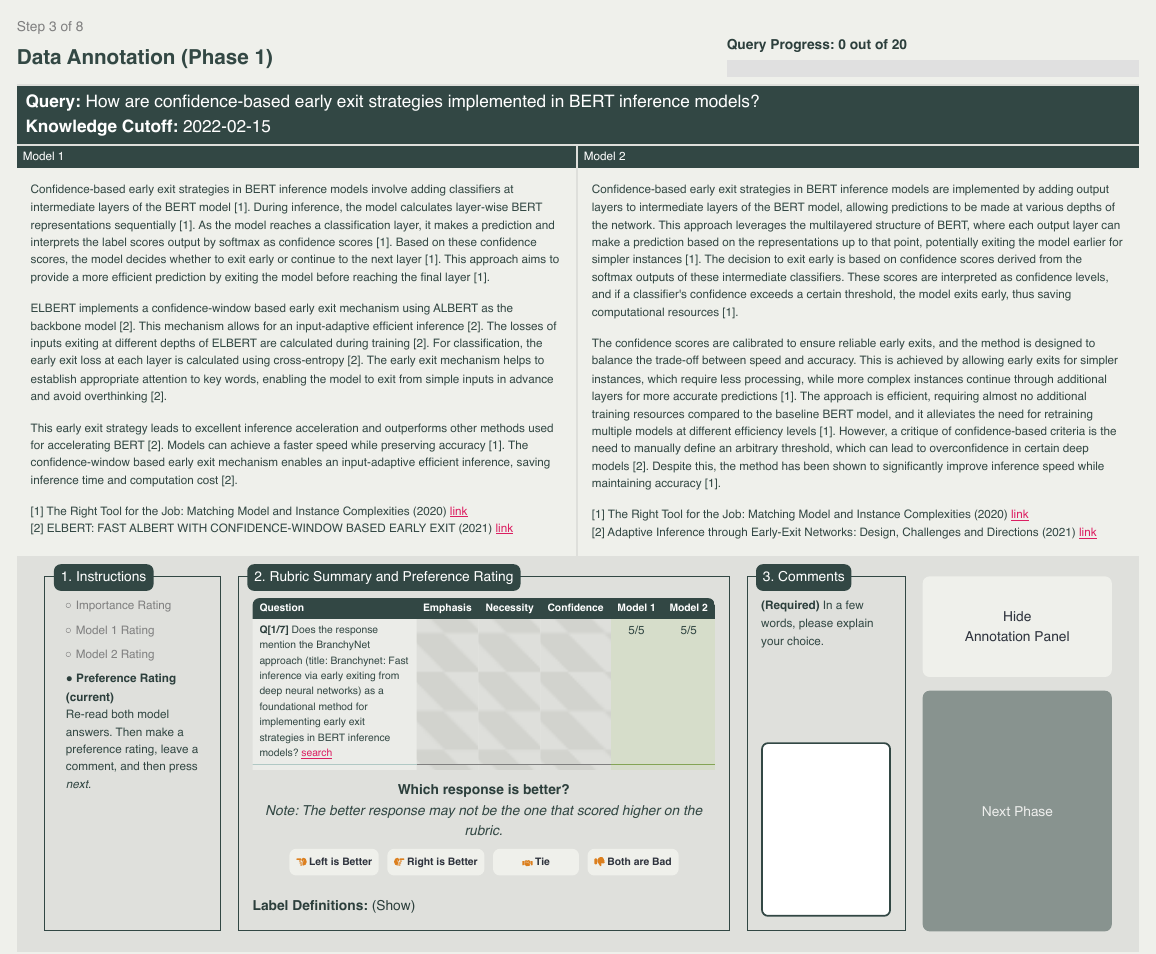}
\caption{\dataset human evaluation interface.}
\label{fig:eval_interface}
\end{figure*}
\definecolor{rubricgreen}{HTML}{297B00}
\definecolor{rubricorange}{HTML}{BD880C}

\begin{figure*}[!t]
\centering
\includegraphics[width=\textwidth]{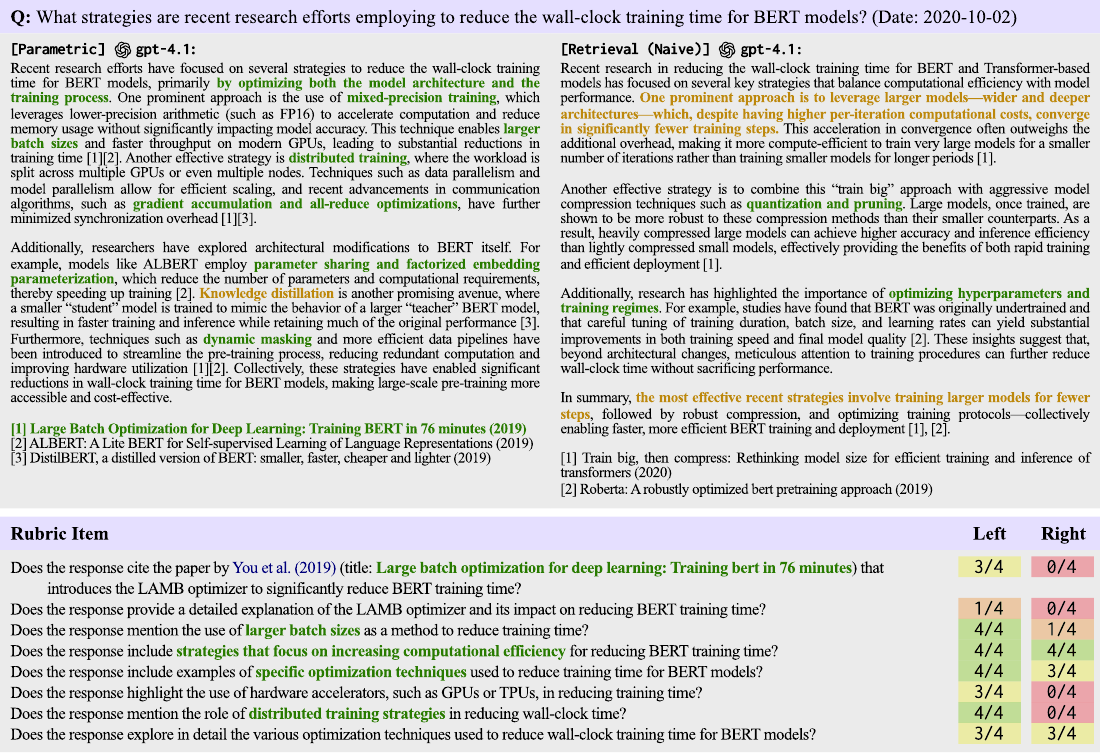}
\vspace{-1.5em}
\caption{Retrieval-augmented  responses can overly commit to orthogonal context, losing to parametric responses. {\color{rubricgreen} Green text} is content mentioned in the rubrics; {\color{rubricorange} Orange text} is content generally not in the rubrics.}
\label{fig:parametric_vs_rag}
\end{figure*}
\begin{figure*}[!t]
\centering
\includegraphics[width=\textwidth]{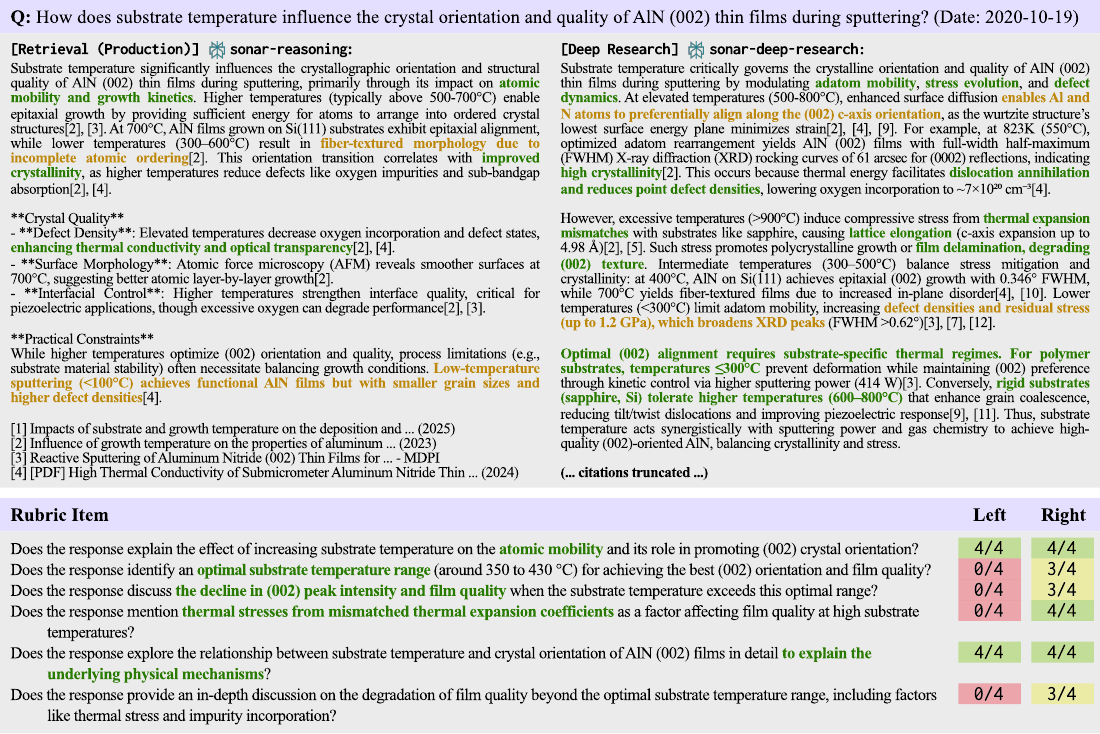}
\vspace{-1.5em}
\caption{Deep research responses cover more concepts in greater detail. {\color{rubricgreen} Green text} is content mentioned explicitly in the rubrics; {\color{rubricorange} Orange text} is content generally not explicit in the rubrics.}
\label{fig:rag_vs_dr}
\end{figure*}

\begin{figure*}[!t]
\centering
\includegraphics[width=0.8\textwidth]{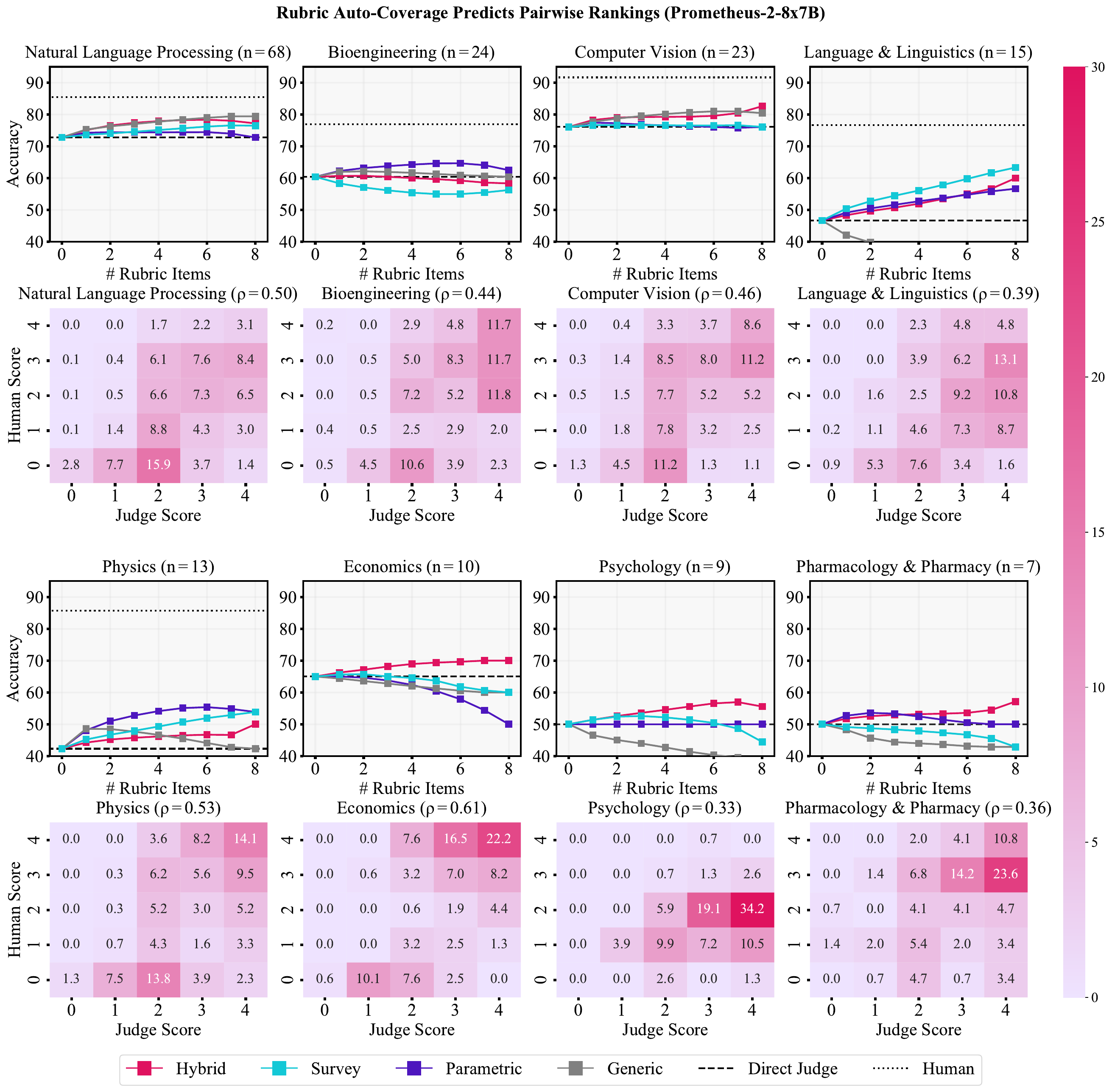}
\caption{\textbf{Rubrics used in ensemble judges can help to predict expert-labeled pairwise preference}. Each graph represents fields covered by expert annotators. Heatmaps show confusion matrices of LLM and human rubric coverage scores. $n$ indicates the count of binary majority labels. In 7 out of 8 fields, hybrid rubrics can improve accuracy to majority labels, with hybrid rubrics obtaining highest performance in 4 out of 8.}
\label{fig:auto-eval-by-field}
\end{figure*}

\begin{figure*}[!t]
\centering
\includegraphics[width=0.75\textwidth]{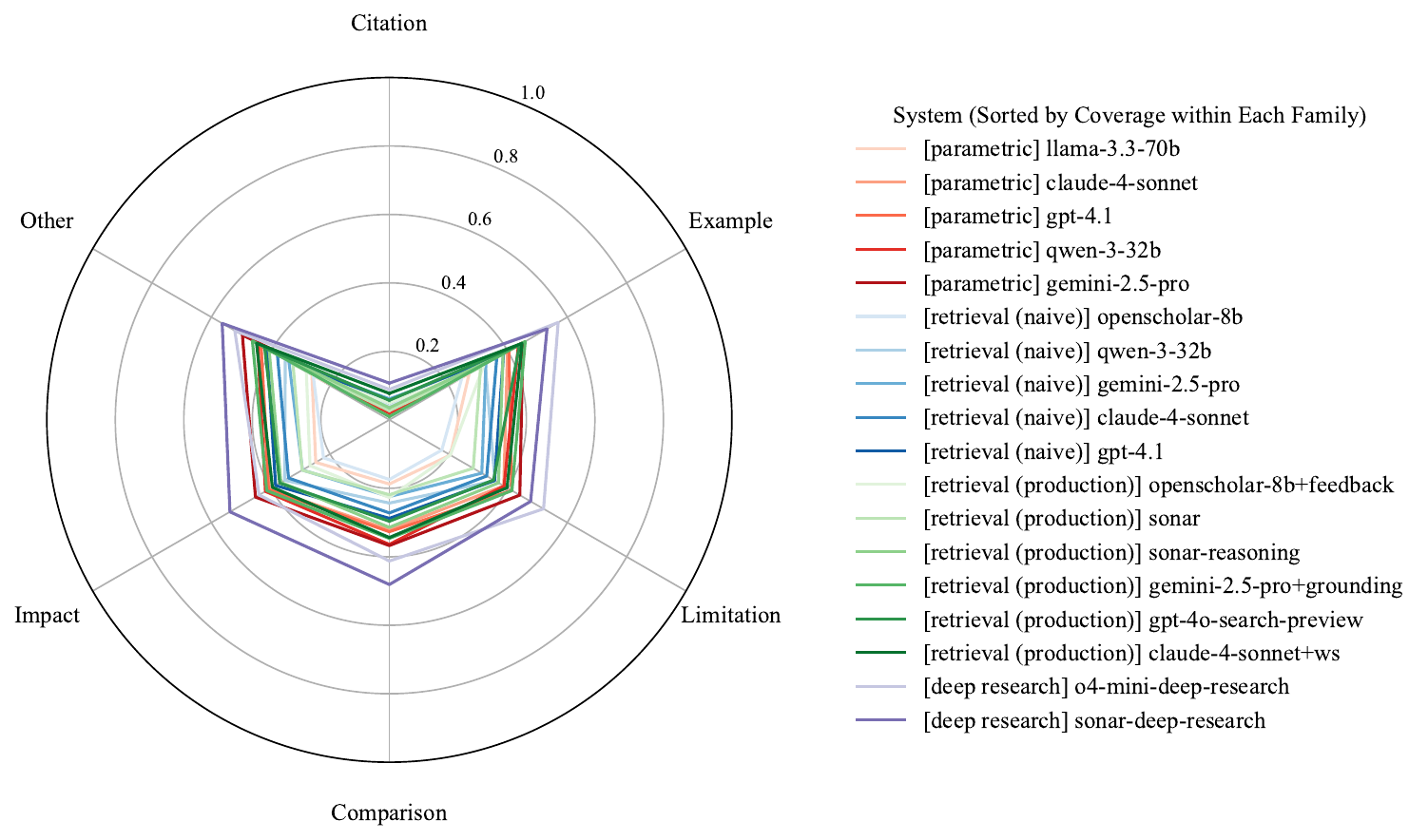}
\caption{
\textbf{Comparison of LLM System Performance by Rubric Type (Radar).} Performance is measured as the percentage of fully covered rubrics. Each rubric type is represented as an objective, with \sonardeepresearch forming the Pareto frontier. ``DR'' denotes ``Deep Research.''}
\label{fig:systems_rubric_radar}
\end{figure*}

\begin{figure*}[!t]
\centering
\includegraphics[width=0.9\textwidth]{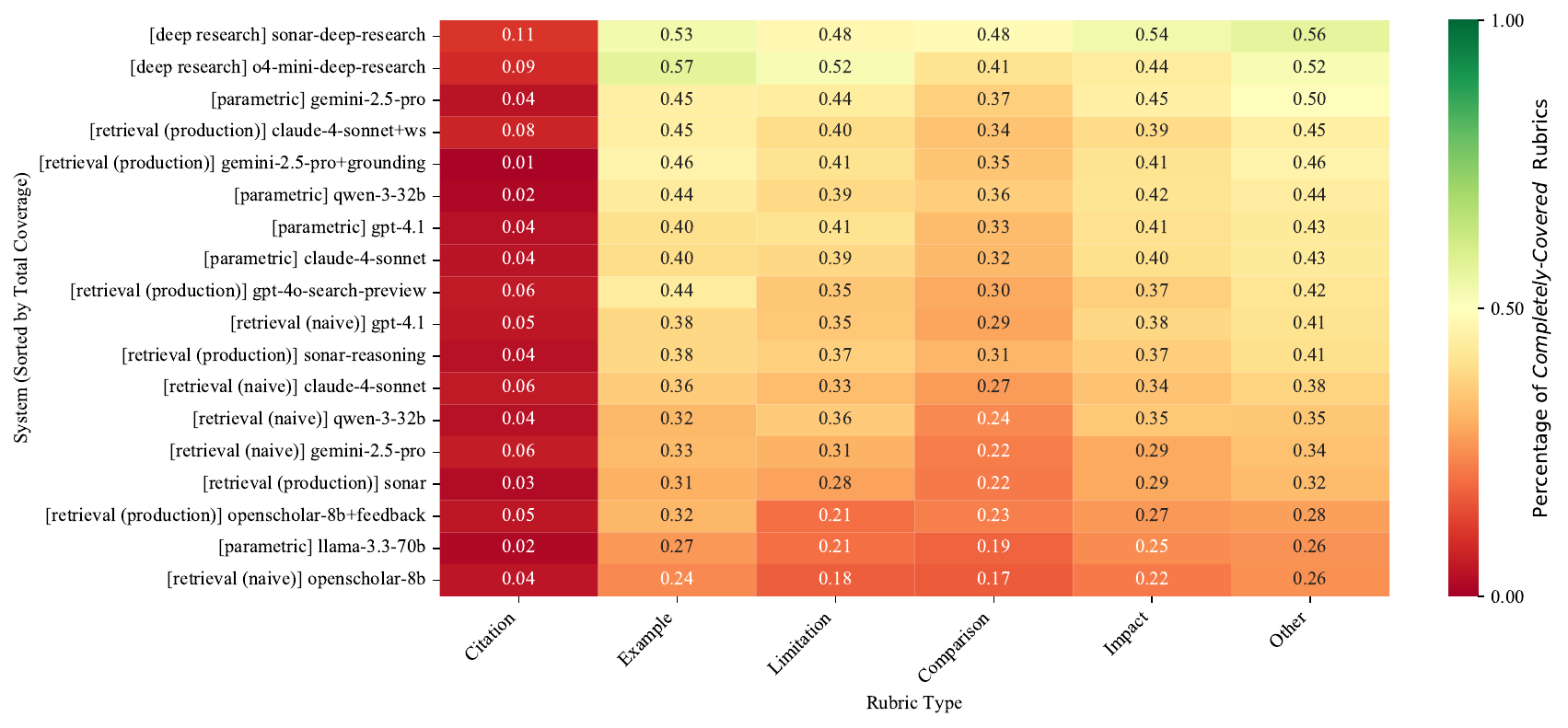}
\caption{
\textbf{Comparison of LLM System Performance by Rubric Type (Heatmap).} Performance is measured as the percentage of fully covered rubrics. Underperforming rubric types (x-axis) are shown for each system in the pairwise tournament (y-axis). ``DR'' denotes ``Deep Research.''}
\label{fig:systems_rubric_heatmap}
\end{figure*}

\begin{figure*}[!t]
\centering
\includegraphics[width=0.6\textwidth]{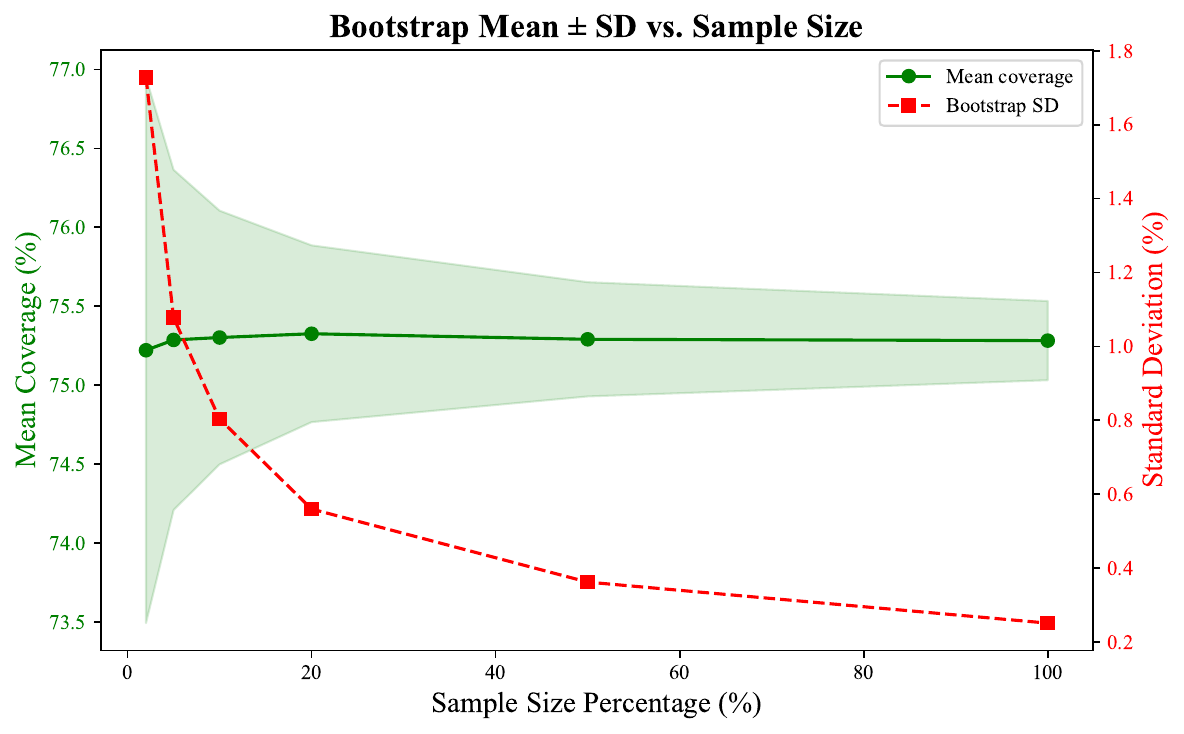}
\caption{\textbf{Coverage \% is reliable at 20\% sample size of \datatest.} Mean coverage and 95\% bootstrapped confidence interval (green) and standard deviation (red) across varying sample sizes for \sonardeepresearch.
A 20\% sample size (750 queries) can provide a stable estimate of rubric coverage with a standard deviation of approximately 0.6\%, suggesting reliable approximation under budget constraints.}
\label{fig:coverage-vs-size}
\end{figure*}

\begin{table*}[t]
\centering
\scriptsize
\setlength{\tabcolsep}{5pt}
\renewcommand{\arraystretch}{0.98}
\newcommand{\pair}[2]{#1\,/\,#2} 

\begin{tabular}{l r r r r}
\toprule
\textbf{System} & \textbf{Leakage \%} & \multicolumn{3}{c}{\textbf{Coverage \%}} \\
\cmidrule{3-5}
&& $\neg$~Leaked & Leaked & $\Delta$\\
\midrule
{\scriptsize Parametric} \\
\cmidrule(r){1-1}
\iconmeta~\llamathreethreeseventyb   & 3  & \textbf{53.49} & \diffannot{51.11}{$-$2.4} \\
\iconanthropic~\claudefoursonnet          & 4  & \textbf{64.44} & \diffannot{61.21}{$-$3.2} \\
\iconopenai~\gptfourone                & 8  & \textbf{65.91} & \diffannot{60.18}{$-$5.7} \\
\iconqwen~\qwenthreethirtytwob       & 3  & \textbf{66.69} & \diffannot{65.08}{$-$1.6} \\
\icongoogle~\geminitwofivepro          & 5  & \textbf{69.18} & \diffannot{62.04}{$-$7.1} \\

\cmidrule(r){1-1}
{\scriptsize Retrieval (Naive)} \\
\cmidrule(r){1-1}
\iconallenai~\openscholareightb         & 0  & 54.71 & \diffannot{---}{---} \\
\icongoogle~\geminitwofivepro          & 0  & 59.92 & \diffannot{---}{---} \\
\iconqwen~\qwenthreethirtytwob       & 0  & 60.90 & \diffannot{---}{---} \\
\iconanthropic~\claudefoursonnet          & 0  & 62.50 & \diffannot{---}{---} \\
\iconopenai~\gptfourone                & 0  & 64.80 & \diffannot{---}{---} \\

\cmidrule(r){1-1}
{\scriptsize Retrieval (Production)} \\
\cmidrule(r){1-1}
\iconperplexity~\sonar                     & 18 & \textbf{58.84} & \diffannot{57.50}{$-$1.3} \\
\iconallenai~\openscholareightbagentic  & 29 & 57.57 & \diffannot{\textbf{61.54}}{$+$4.0} \\
\iconperplexity~\sonarreasoning            & 22 & 64.21 & \diffannot{\textbf{64.77}}{$+$0.6} \\
\iconopenai~\openaiwebsearch           & 27 & 65.30 & \diffannot{\textbf{67.81}}{$+$2.5} \\
\icongoogle~\geminiwebsearch           & 2  & \textbf{68.54} & \diffannot{66.91}{$-$1.6} \\
\iconanthropic~\anthropicwebsearch        & 30 & 68.30 & \diffannot{\textbf{71.27}}{$+$3.0} \\

\cmidrule(r){1-1}
{\scriptsize Deep Research} \\
\cmidrule(r){1-1}
\iconopenai~\openaideepresearch        & 28 & 71.85 & \diffannot{\textbf{74.82}}{$+$3.0} \\
\iconperplexity~\sonardeepresearch         & 21 & \textbf{75.51} & \diffannot{74.46}{$-$1.1} \\
\bottomrule
\end{tabular}
\caption{Retrieving the distilled survey does not provide large advantages toward higher Coverage~\%.
Parametric systems have 3-8\% leakage (L\%); Retrieval (Naive) do not produce answers with leakage (likely because surveys are intentionally removed from their retrieval); Retrieval (Production) and Deep Research result in 20-30\% leakage.
Coverage \% roughly stays the same with leakage (Leaked) and without leakage ($\neg$~Leaked).}
\label{table:agentic-leakage-full}
\end{table*}


\begin{figure*}[!t]
\centering
\includegraphics[width=0.9\textwidth]{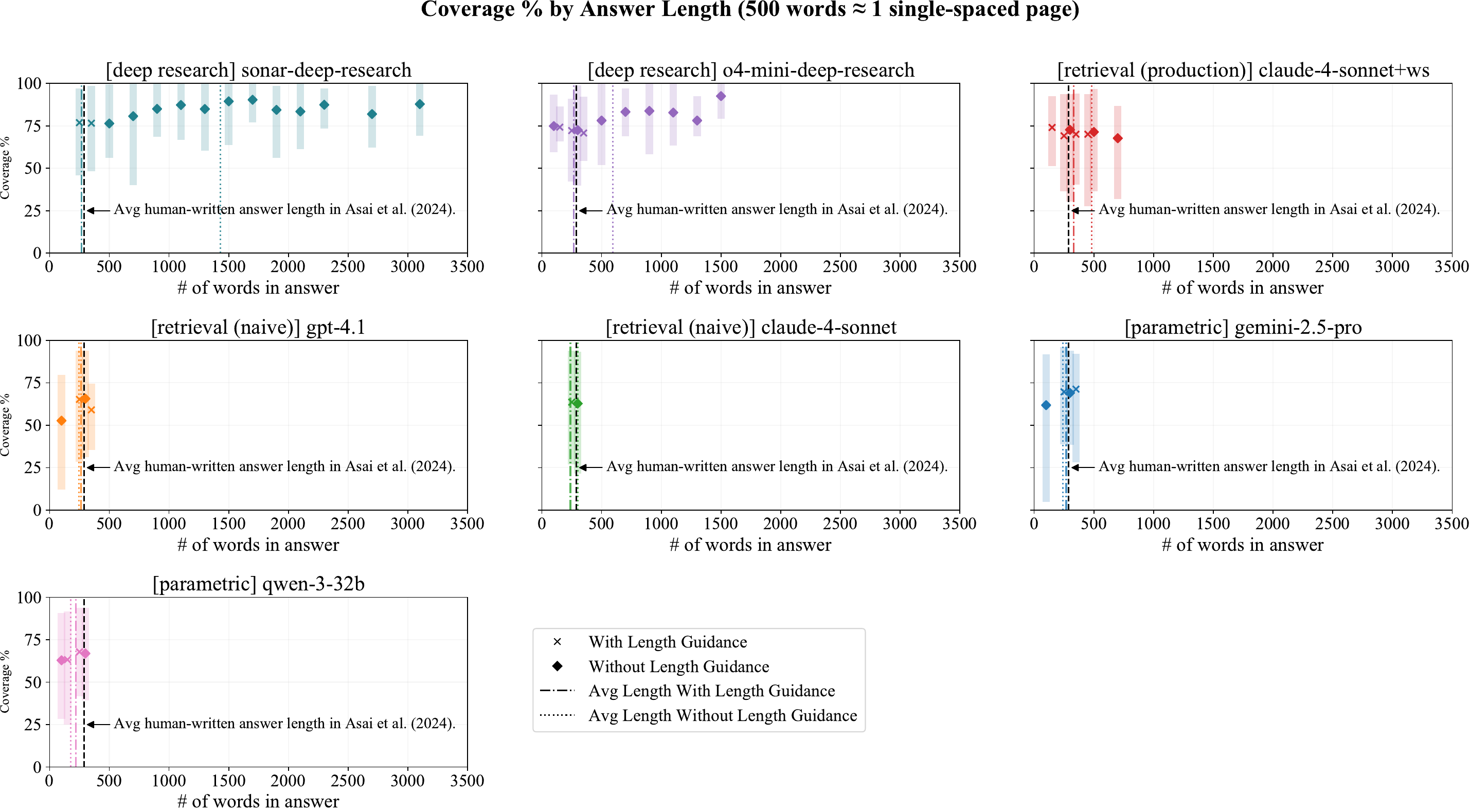}
\caption{\textbf{Coverage \% vs. Length.} 
The relation between Coverage \% and answer length for the top performers in parametric, naive retrieval, and deep research categories selected by Coverage \%, generating answers for 3 queries per field (225 total). Each subplot shows binned coverage across answer lengths, with markers distinguishing answers with vs. without length guidance. Vertical dashed lines indicate the average human-written answer length reported by \citet{openscholar}; and Error bars represent 95\% confidence intervals from bootstrap resampling.
}
\label{fig:coverage-vs-length}
\end{figure*}

\FloatBarrier
\clearpage
\section{Model Prompts}
\label{appendix:model_prompts}

\prompt
{Survey filtering prompt: is literature review}{
\systemprompt{
Determine if the academic paper title indicates a LITERATURE REVIEW.\\
\\
Answer ONLY ``Yes'' if the paper's PRIMARY PURPOSE is to analyze, synthesize, and evaluate research on a topic.\\
\\
Answer ONLY ``No'' for all other paper types, including:\\
\\
1. Papers describing user studies, surveys of people, or field observations\\
2. Papers that report on events, occurrences, or observations (even if they use ``review'' in the title)\\
3. Reports that summarize disease outbreaks, statistics, or time-period data\\
4. Papers that evaluate or compare methods for conducting literature reviews\\
\\
IMPORTANT DISTINCTIONS:\\
- A LITERATURE REVIEW is about analyzing research itself\\
- Having words like ``review,'' ``survey,'' ``meta-analysis,'' or ``overview'' in the title is NOT sufficient
}
\vspace{0.5em}
\userprompt{
Paper Title: \prompttag{\textbf{\color{insert} <PAPER\_TITLE>}}
}}
\prompt
{Query generation prompt: hierarchical summarization}{
\systemprompt{
In this task, you will be given a section from an academic paper, split into individual sentences. \\
You will also receive a set of questions based on a partial section in a hierarchical manner, where each question is clearly answered by the sentence indicated by its index. \\
You need to generate a question that, within the scope of the hierarchical question set, requires understanding the corresponding part of this section. \\
Ideally, this question should be based on the sentence, along with sufficient following sentences, as indicated in the hierarchical question set. \\
Also, provide the supporting sentence indices. \\
\\
\# Content\\
Paper Title: Societal Biases in Language Generation: Progress and Challenges\\
Section Title: Bias Definitions and Metrics\\
Content: \\
sent1: In the context of AI fairness, the term ``bias'' commonly refers to skews that result in undesirable impacts (Crawford, 2017) and is quantifiable with some metric.\\
sent2: There are relatively more existing studies on biases in NLU tasks, where it is arguably simpler to define bias metrics, since we can intuitively compare the accuracy of the task (e.g., coreference resolution, hate speech detection) for different demographics.\\
sent3: Language generation tasks often involve stochastic generation of open-ended and lengthy texts, traits that are not directly compatible with traditional algorithmic bias definitions (e.g., equalized odds, equal opportunity, demographic parity (Dwork et al., 2012; Hardt et al., 2016)).\\
sent4: Because of the difficulty in defining metrics, existing works define bias loosely as demographic inequality and use intermediate proxy metrics to comparatively measure bias.\\
sent5: Examples include: Regard Ratio: negative-neutral-positive regard score ratios of text generated from bias-inducing prompts (Sheng et al., 2019).\\
sent6: Sentiment Ratio: negative-neutral-positive sentiment score ratios of text generated from African American English (AAE) versus White-Aligned English (WAE) prompts (Groenwold et al., 2020).\\
sent7: Individual and Group Fairness through Sentiment: comparisons of the sentiment distributions of generated text across demographics and prompts (Huang et al., 2020).\\
sent8: Gendered Word Co-occurrence Score: mean and standard deviations of the absolute log ratio of probabilities: P(word|female terms) to P(word|male terms) across all words in generated text (Bordia and Bowman, 2019).\\
sent9: There are also metrics for other bias evaluation setups in continuation generation tasks involving sentiment (Shwartz et al., 2020), the ratio of gendered words (Solaiman et al., 2019; Vig et al., 2020; Dinan et al., 2020a), and other novel metrics (Peng et al., 2020; Yeo and Chen, 2020).\\
sent10: Studies of biases in transformation generation tasks favor metrics of accuracy in terms of successfully transforming text to have a desired property.\\
sent11: We present a more thorough comparison of metrics in Section 5.4.\\
sent12: Bias metrics can also be categorized by how they define associations between demographic group attributes and text.\\
sent13: Biases can be towards people described in text, people who produce the text, or people to whom the text is addressed (Dinan et al., 2020b).\\
sent14: Most existing works define bias metrics through the first association—these biases are relatively easier to analyze, since both the demographic and the textual signals of bias are encapsulated within the text.\\
sent15: There are also works that define biases towards people who produce the text (Groenwold et al., 2020) or people to whom the text is addressed (Sheng et al., 2021b), though there are relatively fewer works that study these latter associations.\\
Output:\\
1. What is the common definition of ``bias'' in the context of AI fairness? sent1\\
2. How are biases in NLU tasks typically defined and measured? sent2\\
3. Why are traditional algorithmic bias definitions not directly compatible with language generation tasks? sent3\\
4. How do existing works define and measure bias in language generation tasks? sent4\\
    4.1. What are some examples of intermediate proxy metrics used to measure bias in language generation tasks? sent5, sent6, sent7, sent8\\
    4.2. Are there other metrics for bias evaluation in continuation generation tasks? sent9\\
    4.3. What metrics are favored in transformation generation tasks? sent10\\
    4.4. Where can a more thorough comparison of metrics be found? sent11\\
5. How can bias metrics be categorized? sent12\\
6. What are the different associations between demographic group attributes and text in bias metrics? sent13\\
    6.1. Which association is most commonly used in existing works and why? sent14\\
    6.2. Are there works that define biases towards people who produce the text or people to whom the text is addressed? sent15\\
\\
\# Content\\
Paper Title: Modeling Language Variation and Universals: A Survey on Typological Linguistics for Natural Language Processing\\
Section Title: Hand-Crafted Documentation in Typological Databases\\
Content: \\
sent1: Typological databases are created manually by linguists.\\
sent2: They contain taxonomies of typological features, their possible values, as well as the documentation of feature values for the world’s languages.\\
sent3: Major typological databases, listed in Table 1, typically organize linguistic information in terms of universal features and language-specific values.\\
sent4: For example, Figure 3 presents language-specific values for the feature number of grammatical genders for nouns on a world map.\\
sent5: Note that each language is color-coded according to its value.\\
sent6: Further examples for each database can be found in the rightmost column of Table 1.\\
sent7: Some databases store information pertaining to multiple levels of linguistic description.\\
sent8: These include WALS (Dryer and Haspelmath 2013) and the Atlas of Pidgin and Creole Language Structures (APiCS) (Michaelis et al. 2013).\\
sent9: Among all presently available databases, WALS has been the most widely used in NLP.\\
sent10: In this resource, which has 142 typological features in total, features 1–19 deal with phonology, 20–29 with morphology, 30–57 with nominal categories, 58–64 with nominal syntax, 65–80 with verbal categories, 81–97 and 143–144 with word order, 98–121 with simple clauses, 122–128 with complex sentences, 129–138 with the lexicon, and 139–142 with other properties.\\
sent11: Other databases only cover features related to a specific level of linguistic description.\\
sent12: For example, both Syntactic Structures of the World’s Languages (SSWL) (Collins and Kayne 2009) and AUTOTYP (Bickel et al. 2017) focus on syntax.\\
sent13: SSWL features are manually crafted, whereas AUTOTYP features are derived automatically from primary linguistic data using scripts.\\
sent14: The Valency Patterns Leipzig (ValPaL) (Hartmann, Haspelmath, and Taylor 2013) provides verbs as attributes and predicate–argument structures as their values (including both valency and morphosyntactic constraints).\\
sent15: For example, in both Mandinka and Sliammon, the verb to laugh has a valency of 1; in other words, it requires only one mandatory argument, the subject.\\
sent16: In Mandinka the subject precedes the verb, but there is no agreement requirement; in Sliammon, on the other hand, the word order does not matter, but the verb is required to morphologically agree with the subject.\\
sent17: For phonology, the Phonetics Information Base and Lexicon (PHOIBLE) (Moran, McCloy, and Wright 2014) collates information on segments (binary phonetic features).\\
sent18: In the Lyon–Albuquerque Phonological Systems Database (LAPSyD) (Maddieson et al. 2013), attributes are articulatory traits, syllabic structures, or tonal systems.\\
sent19: Finally, StressTyp2 (Goedemans, Heinz, and der Hulst 2014) deals with stress and accent patterns.\\
sent20: For instance, in Koromfé each word’s first syllable has to be stressed, but not in Cubeo.\\
sent21: Other databases document various aspects of semantics.\\
sent22: The World Loanword Database (WOLD) (Haspelmath and Tadmor 2009) documents loanwords by identifying the donor languages and the source words.\\
sent23: The Automated Similarity Judgment Program (ASJP) (Wichmann, Holman, and Brown 2016) and the Intercontinental Dictionary Series (IDS) (Key and Comrie 2015) indicate how a meaning is lexicalized across languages:\\
sent24: For example, the concept of WORLD is expressed as mir in Russian, and as ārkiśos. i in Tocharian A.\\
sent25: Although typological databases store abundant information on many languages, they suffer from shortcomings that limit their usefulness.\\
sent26: Perhaps the most significant shortcoming of such resources is their limited coverage.\\
sent27: In fact, feature values are missing for most languages in most databases.\\
sent28: Other shortcomings are related to feature granularity.\\
sent29: In particular, most databases fail to account for feature value variation within each language: They report only majority value rather than the full range of possible values and their corresponding frequencies.\\
sent30: For example, the dominant adjective–noun word order in Italian is adjective before noun; however, the opposite order is also attested.\\
sent31: The latter information is often missing from typological databases.\\
sent32: Further challenges are posed by restricted feature applicability and feature hierarchies.\\
sent33: Firstly, some features apply, by definition, only to subsets of languages that share another feature value.\\
sent34: For instance, WALS feature 113A documents ``Symmetric and Asymmetric Standard Negation,'' whereas WALS feature 114A ``Subtypes of Asymmetric Standard Negation.''\\
sent35: Although a special NA value is assigned for symmetric-negation languages in the latter, there are cases where languages without the prerequisite feature are simply omitted from the sample.\\
sent36: Secondly, features can be partially redundant, and subsume other features.\\
sent37: For instance, WALS feature 81A ``Order of Subject, Object and Verb'' encodes the same information as WALS feature 82A ``Order of Subject and Verb'' and 83A ``Order of Object and Verb,'' with the addition of the order of subject and object.\\
Output:\\
1. What are typological databases and how are they created? sent1\\
    1.1. What do typological databases contain? sent2\\
    1.2. How is linguistic information organized in major typological databases? sent3\\
        1.2.1. Can you provide an example of how linguistic information is organized? sent4, sent5\\
        1.2.2. Where can further examples for each database be found? sent6\\
    1.3. Do some databases store information at multiple levels of linguistic description? sent7\\
        1.3.1. Which databases store information at multiple levels of linguistic description? sent8\\
        1.3.2. Which database is the most widely used in NLP? sent9\\
            1.3.2.1. What features does WALS cover? sent10\\
    1.4. Do other databases cover specific levels of linguistic description? sent11\\
        1.4.1. Can you provide examples of databases that focus on specific levels of linguistic description? sent12\\
            1.4.1.1. How are features in SSWL and AUTOTYP crafted? sent13\\
            1.4.1.2. What does the Valency Patterns Leipzig (ValPaL) provide? sent14\\
                1.4.1.2.1. Can you provide an example of valency in different languages? sent15, sent16\\
            1.4.1.3. What information does PHOIBLE collate? sent17\\
            1.4.1.4. What attributes are included in LAPSyD? sent18\\
            1.4.1.5. What does StressTyp2 deal with? sent19\\
                1.4.1.5.1. Can you provide an example of stress patterns in different languages? sent20\\
    1.5. Do other databases document aspects of semantics? sent21\\
        1.5.1. What does the World Loanword Database (WOLD) document? sent22\\
        1.5.2. What do ASJP and IDS indicate? sent23\\
            1.5.2.1. Can you provide an example of how a meaning is lexicalized across languages? sent24\\
2. What are the shortcomings of typological databases? sent25\\
    2.1. What is the most significant shortcoming? sent26\\
        2.1.1. What is the evidence for the limited coverage of typological databases? sent27\\
    2.2. What are other shortcomings related to? sent28\\
        2.2.1. How do most databases fail to account for feature value variation within each language? sent29\\
            2.2.1.1. Can you provide an example of missing information in typological databases? sent30, sent31\\
    2.3. What further challenges are posed by typological databases? sent32\\
        2.3.1. What is the first challenge related to? sent33\\
            2.3.1.1. Can you provide an example of restricted feature applicability? sent34, sent35\\
        2.3.2. What is the second challenge related to? sent36\\
            2.3.2.1. Can you provide an example of feature redundancy and subsumption? sent37\\
\\
\# Content\\
Paper Title: Efficient Methods for Natural Language Processing: A Survey\\
Section Title: Sparse Modeling\\
Content: \\
sent1: To leverage sparsity for efficiency, many models follow the mixture-of-experts (MoE) concept (Jacobs et al., 1991; Shazeer et al., 2017; Fedus et al., 2022a), which routes computation through small subnetworks instead of passing the input through the entire model.\\
sent2: Relevant works on this line include GShard (Lepikhin et al., 2021), Switch Transformer (Fedus et al., 2022b), and ST-MoE (Zoph et al., 2022), which replace the feed forward layers in transformers with MoE layers.\\
sent3: More recently, Rajbhandari et al. (2022) scaled transformers up by compressing and optimizing the usage of MoE.\\
sent4: Overall, MoE models have been shown to achieve strong performance across several NLP tasks while reducing the overall resource consumption (Sec. 8).\\
sent5: For instance, GLaM (Du et al., 2022) used only $\sim$1/3 of GPT-3’s energy consumption (with additional hardware-based optimization), while Rajbhandari et al. (2022) reached a 5x reduction in terms of training cost.\\
sent6: However, MoE models have also exhibited training instabilities in practice, and may require architecture-specific implementation (Zoph et al., 2022; Mustafa et al., 2022).\\
sent7: Another promising direction for exploiting sparse modeling is Sparsefinder (Treviso et al., 2022), which extends the Adaptively Sparse Transformer (Correia et al., 2019) to allow a more efficient attention mechanism by identifying beforehand the sparsity pattern returned by entmax attention—a sparse alternative to (dense) softmax attention (Peters et al., 2019).\\
sent8: Finally, sparsity can also be induced via modularity, e.g., by encapsulating task-specific parameters (Ponti et al., 2022).\\
Output:\\
1. How do models leverage sparsity for efficiency, such as MoE method? sent1\\
    1.1. What are some relevant works that follow the MoE concept? sent2\\
        1.1.1 How have recent works scaled transformers using MoE? sent3\\
    1.2. What is the overall performance and resource consumption of MoE models? sent4\\
        1.2.1. Can you provide specific examples of resource consumption reduction in MoE models? sent5\\
        1.2.2. What are some challenges associated with MoE models? sent6\\
    1.3. What is another promising direction for exploiting sparse modeling? sent7\\
    1.4. How can sparsity be induced via modularity? sent8\\
\\
\# Content\\
Paper Title: \prompttag{\textbf{\color{insert}<PAPER\_TITLE>}}\\
Section Title: \prompttag{\textbf{\color{insert}<SECTION\_TITLE>}}\\
Content: \\
\prompttag{\textbf{\color{insert}<SECTION\_SENTENCE\_PREFIXED>}}\\
Output:
}}%
\prompt
{Query generation prompt: generate initial query}{
\systemprompt{
In this task, you will be given a section from an academic paper, split up into individual sentences. \\
Also a set of questions based on a partial section in a hierarchical manner, where each question is clearly answered by the sentence indicated by the index.\\
You need to generate a question that within the scope of the hierarchical question set, requires understanding the corresponding part of this section.\\
Ideally, this question is based on the sentence that with sufficent following sentences, indicated in the hierarchical question set.\\
Also provide the suppprintg sentence indexs.\\
\\
This question needs to be:\\
1. Unambiguous: Clearly framed so it does not require follow-up questions for clarification. \\
    1.1 It should be understandable to any expert without needing specific context or jargon found in the given section.\\
2. Natural: Phrased as if it is asked by a domain expert conducting research.\\
3. Answerable: Should be entirely answerable based on the provided section.\\
4. Precise: Question should express a clear information need and not be vague. \\
    4.1 Multiple experts should answer such a question in a similar way.\\
    4.2 The question may need to mention the sub-area and domain specified in the section (especially titles) to be precise.\\
5. Requires a long-form, comprehensive answer: not simply extractive or yes-no questions.\\
6. The question should not specifically about a particular paper e.g. Artetxe and Schwenk (2019).\\
7. The question should be based on the sentence that with sufficent following sentences, indicated in the hierarchical question set.\\
8. Less than 20 words.\\
\\
\# Content: \\
Paper Title: Societal Biases in Language Generation: Progress and Challenges\\
Section Title: Bias Definitions and Metrics\\
Section content: \\
sent1: In the context of AI fairness, the term ``bias'' commonly refers to skews that result in undesirable impacts (Crawford, 2017) and is quantifiable with some metric.\\
sent2: There are relatively more existing studies on biases in NLU tasks, where it is arguably simpler to define bias metrics, since we can intuitively compare the accuracy of the task (e.g., coreference resolution, hate speech detection) for different demographics.\\
sent3: Language generation tasks often involve stochastic generation of open-ended and lengthy texts, traits that are not directly compatible with traditional algorithmic bias definitions (e.g., equalized odds, equal opportunity, demographic parity (Dwork et al., 2012; Hardt et al., 2016)).\\
sent4: Because of the difficulty in defining metrics, existing works define bias loosely as demographic inequality and use intermediate proxy metrics to comparatively measure bias.\\
sent5: Examples include: Regard Ratio: negative-neutral-positive regard score ratios of text generated from bias-inducing prompts (Sheng et al., 2019).\\
sent6: Sentiment Ratio: negative-neutral-positive sentiment score ratios of text generated from African American English (AAE) versus White-Aligned English (WAE) prompts (Groenwold et al., 2020).\\
sent7: Individual and Group Fairness through Sentiment: comparisons of the sentiment distributions of generated text across demographics and prompts (Huang et al., 2020).\\
sent8: Gendered Word Co-occurrence Score: mean and standard deviations of the absolute log ratio of probabilities: P(word|female terms) to P(word|male terms) across all words in generated text (Bordia and Bowman, 2019).\\
sent9: There are also metrics for other bias evaluation setups in continuation generation tasks involving sentiment (Shwartz et al., 2020), the ratio of gendered words (Solaiman et al., 2019; Vig et al., 2020; Dinan et al., 2020a), and other novel metrics (Peng et al., 2020; Yeo and Chen, 2020).\\
sent10: Studies of biases in transformation generation tasks favor metrics of accuracy in terms of successfully transforming text to have a desired property.\\
sent11: We present a more thorough comparison of metrics in Section 5.4.\\
sent12: Bias metrics can also be categorized by how they define associations between demographic group attributes and text.\\
sent13: Biases can be towards people described in text, people who produce the text, or people to whom the text is addressed (Dinan et al., 2020b).\\
sent14: Most existing works define bias metrics through the first association—these biases are relatively easier to analyze, since both the demographic and the textual signals of bias are encapsulated within the text.\\
sent15: There are also works that define biases towards people who produce the text (Groenwold et al., 2020) or people to whom the text is addressed (Sheng et al., 2021b), though there are relatively fewer works that study these latter associations.\\
Hierarchical question set:\\
1. What is the common definition of ``bias'' in the context of AI fairness? sent1\\
2. How are biases in NLU tasks typically defined and measured? sent2\\
3. Why are traditional algorithmic bias definitions not directly compatible with language generation tasks? sent3\\
4. How do existing works define and measure bias in language generation tasks? sent4\\
    4.1. What are some examples of intermediate proxy metrics used to measure bias in language generation tasks? sent5, sent6, sent7, sent8\\
    4.2. Are there other metrics for bias evaluation in continuation generation tasks? sent9\\
    4.3. What metrics are favored in transformation generation tasks? sent10\\
    4.4. Where can a more thorough comparison of metrics be found? sent11\\
5. How can bias metrics be categorized? sent12\\
6. What are the different associations between demographic group attributes and text in bias metrics? sent13\\
    6.1. Which association is most commonly used in existing works and why? sent14\\
    6.2. Are there works that define biases towards people who produce the text or people to whom the text is addressed? sent15\\
Output:\\
Question: When generating text, how are the major types of bias measures used to evaluate bias with respect to gender?\\
Supporting sentence: sent4, sent5, sent6, sent7, sent8, sent9\\
\\
\# Content\\
Paper Title: Modeling Language Variation and Universals: A Survey on Typological Linguistics for Natural Language Processing\\
Section Title: Hand-Crafted Documentation in Typological Databases\\
Section content: \\
sent1: Typological databases are created manually by linguists.\\
sent2: They contain taxonomies of typological features, their possible values, as well as the documentation of feature values for the world’s languages.\\
sent3: Major typological databases, listed in Table 1, typically organize linguistic information in terms of universal features and language-specific values.\\
sent4: For example, Figure 3 presents language-specific values for the feature number of grammatical genders for nouns on a world map.\\
sent5: Note that each language is color-coded according to its value.\\
sent6: Further examples for each database can be found in the rightmost column of Table 1.\\
sent7: Some databases store information pertaining to multiple levels of linguistic description.\\
sent8: These include WALS (Dryer and Haspelmath 2013) and the Atlas of Pidgin and Creole Language Structures (APiCS) (Michaelis et al. 2013).\\
sent9: Among all presently available databases, WALS has been the most widely used in NLP.\\
sent10: In this resource, which has 142 typological features in total, features 1–19 deal with phonology, 20–29 with morphology, 30–57 with nominal categories, 58–64 with nominal syntax, 65–80 with verbal categories, 81–97 and 143–144 with word order, 98–121 with simple clauses, 122–128 with complex sentences, 129–138 with the lexicon, and 139–142 with other properties.\\
sent11: Other databases only cover features related to a specific level of linguistic description.\\
sent12: For example, both Syntactic Structures of the World’s Languages (SSWL) (Collins and Kayne 2009) and AUTOTYP (Bickel et al. 2017) focus on syntax.\\
sent13: SSWL features are manually crafted, whereas AUTOTYP features are derived automatically from primary linguistic data using scripts.\\
sent14: The Valency Patterns Leipzig (ValPaL) (Hartmann, Haspelmath, and Taylor 2013) provides verbs as attributes and predicate–argument structures as their values (including both valency and morphosyntactic constraints).\\
sent15: For example, in both Mandinka and Sliammon, the verb to laugh has a valency of 1; in other words, it requires only one mandatory argument, the subject.\\
sent16: In Mandinka the subject precedes the verb, but there is no agreement requirement; in Sliammon, on the other hand, the word order does not matter, but the verb is required to morphologically agree with the subject.\\
sent17: For phonology, the Phonetics Information Base and Lexicon (PHOIBLE) (Moran, McCloy, and Wright 2014) collates information on segments (binary phonetic features).\\
sent18: In the Lyon–Albuquerque Phonological Systems Database (LAPSyD) (Maddieson et al. 2013), attributes are articulatory traits, syllabic structures, or tonal systems.\\
sent19: Finally, StressTyp2 (Goedemans, Heinz, and der Hulst 2014) deals with stress and accent patterns.\\
sent20: For instance, in Koromfé each word’s first syllable has to be stressed, but not in Cubeo.\\
sent21: Other databases document various aspects of semantics.\\
sent22: The World Loanword Database (WOLD) (Haspelmath and Tadmor 2009) documents loanwords by identifying the donor languages and the source words.\\
sent23: The Automated Similarity Judgment Program (ASJP) (Wichmann, Holman, and Brown 2016) and the Intercontinental Dictionary Series (IDS) (Key and Comrie 2015) indicate how a meaning is lexicalized across languages:\\
sent24: For example, the concept of WORLD is expressed as mir in Russian, and as ārkiśos. i in Tocharian A.\\
sent25: Although typological databases store abundant information on many languages, they suffer from shortcomings that limit their usefulness.\\
sent26: Perhaps the most significant shortcoming of such resources is their limited coverage.\\
sent27: In fact, feature values are missing for most languages in most databases.\\
sent28: Other shortcomings are related to feature granularity.\\
sent29: In particular, most databases fail to account for feature value variation within each language: They report only majority value rather than the full range of possible values and their corresponding frequencies.\\
sent30: For example, the dominant adjective–noun word order in Italian is adjective before noun; however, the opposite order is also attested.\\
sent31: The latter information is often missing from typological databases.\\
sent32: Further challenges are posed by restricted feature applicability and feature hierarchies.\\
sent33: Firstly, some features apply, by definition, only to subsets of languages that share another feature value.\\
sent34: For instance, WALS feature 113A documents ``Symmetric and Asymmetric Standard Negation,'' whereas WALS feature 114A ``Subtypes of Asymmetric Standard Negation.''\\
sent35: Although a special NA value is assigned for symmetric-negation languages in the latter, there are cases where languages without the prerequisite feature are simply omitted from the sample.\\
sent36: Secondly, features can be partially redundant, and subsume other features.\\
sent37: For instance, WALS feature 81A ``Order of Subject, Object and Verb'' encodes the same information as WALS feature 82A ``Order of Subject and Verb'' and 83A ``Order of Object and Verb,'' with the addition of the order of subject and object.\\
Hierarchical question set:\\
1. What are typological databases and how are they created? sent1\\
    1.1. What do typological databases contain? sent2\\
    1.2. How is linguistic information organized in major typological databases? sent3\\
        1.2.1. Can you provide an example of how linguistic information is organized? sent4, sent5\\
        1.2.2. Where can further examples for each database be found? sent6\\
    1.3. Do some databases store information at multiple levels of linguistic description? sent7\\
        1.3.1. Which databases store information at multiple levels of linguistic description? sent8\\
        1.3.2. Which database is the most widely used in NLP? sent9\\
            1.3.2.1. What features does WALS cover? sent10\\
    1.4. Do other databases cover specific levels of linguistic description? sent11\\
        1.4.1. Can you provide examples of databases that focus on specific levels of linguistic description? sent12\\
            1.4.1.1. How are features in SSWL and AUTOTYP crafted? sent13\\
            1.4.1.2. What does the Valency Patterns Leipzig (ValPaL) provide? sent14\\
                1.4.1.2.1. Can you provide an example of valency in different languages? sent15, sent16\\
            1.4.1.3. What information does PHOIBLE collate? sent17\\
            1.4.1.4. What attributes are included in LAPSyD? sent18\\
            1.4.1.5. What does StressTyp2 deal with? sent19\\
                1.4.1.5.1. Can you provide an example of stress patterns in different languages? sent20\\
    1.5. Do other databases document aspects of semantics? sent21\\
        1.5.1. What does the World Loanword Database (WOLD) document? sent22\\
        1.5.2. What do ASJP and IDS indicate? sent23\\
            1.5.2.1. Can you provide an example of how a meaning is lexicalized across languages? sent24\\
2. What are the shortcomings of typological databases? sent25\\
    2.1. What is the most significant shortcoming? sent26\\
        2.1.1. What is the evidence for the limited coverage of typological databases? sent27\\
    2.2. What are other shortcomings related to? sent28\\
        2.2.1. How do most databases fail to account for feature value variation within each language? sent29\\
            2.2.1.1. Can you provide an example of missing information in typological databases? sent30, sent31\\
    2.3. What further challenges are posed by typological databases? sent32\\
        2.3.1. What is the first challenge related to? sent33\\
            2.3.1.1. Can you provide an example of restricted feature applicability? sent34, sent35\\
        2.3.2. What is the second challenge related to? sent36\\
            2.3.2.1. Can you provide an example of feature redundancy and subsumption? sent37\\
Output:\\
Question: What are the differences between publicly available linguistic typology databases?\\
Supporting sentence: sent7, sent8, sent9, sent10, sent11, sent12, sent13, sent14, sent17, sent18, sent19, sent21, sent22, sent23 \\
\\
\# Content\\
Paper Title: Efficient Methods for Natural Language Processing: A Survey\\
Section Title: Sparse Modeling\\
Section content: \\
sent1: To leverage sparsity for efficiency, many models follow the mixture-of-experts (MoE) concept (Jacobs et al., 1991; Shazeer et al., 2017; Fedus et al., 2022a), which routes computation through small subnetworks instead of passing the input through the entire model.\\
sent2: Relevant works on this line include GShard (Lepikhin et al., 2021), Switch Transformer (Fedus et al., 2022b), and ST-MoE (Zoph et al., 2022), which replace the feed forward layers in transformers with MoE layers.\\
sent3: More recently, Rajbhandari et al. (2022) scaled transformers up by compressing and optimizing the usage of MoE.\\
sent4: Overall, MoE models have been shown to achieve strong performance across several NLP tasks while reducing the overall resource consumption (Sec. 8).\\
sent5: For instance, GLaM (Du et al., 2022) used only $\sim$1/3 of GPT-3’s energy consumption (with additional hardware-based optimization), while Rajbhandari et al. (2022) reached a 5x reduction in terms of training cost.\\
sent6: However, MoE models have also exhibited training instabilities in practice, and may require architecture-specific implementation (Zoph et al., 2022; Mustafa et al., 2022).\\
sent7: Another promising direction for exploiting sparse modeling is Sparsefinder (Treviso et al., 2022), which extends the Adaptively Sparse Transformer (Correia et al., 2019) to allow a more efficient attention mechanism by identifying beforehand the sparsity pattern returned by entmax attention—a sparse alternative to (dense) softmax attention (Peters et al., 2019).\\
sent8: Finally, sparsity can also be induced via modularity, e.g., by encapsulating task-specific parameters (Ponti et al., 2022).\\
Hierarchical question set:\\
1. How do models leverage sparsity for efficiency, such as MoE method? sent1\\
    1.1. What are some relevant works that follow the MoE concept? sent2\\
        1.1.1 How have recent works scaled transformers using MoE? sent3\\
    1.2. What is the overall performance and resource consumption of MoE models? sent4\\
        1.2.1. Can you provide specific examples of resource consumption reduction in MoE models? sent5\\
        1.2.2. What are some challenges associated with MoE models? sent6\\
    1.3. What is another promising direction for exploiting sparse modeling? sent7\\
    1.4. How can sparsity be induced via modularity? sent8\\
Output:\\
Question: How can we utilize sparsity to enhance efficiency in designing NLP models?\\
Supporting sentence: sent1, sent2, sent3, sent4, sent5, sent6, sent7, sent8\\
\\
\# Content: \\
Paper Title: \prompttag{\textbf{\color{insert}<PAPER\_TITLE>}}\\
Section Title: \prompttag{\textbf{\color{insert}<SECTION\_TITLE>}}\\
Section content: \\
\prompttag{\textbf{\color{insert}<SECTION\_CONTENT\_SENTENCE\_PREFIXED>}}\\
Hierarchical question set:\\
\prompttag{\textbf{\color{insert}<HIERARCHICAL\_SUMMARY>}}\\
Output:
}}%
\prompt
{Query filtering prompt: self-containment}{
\systemprompt{
You will be given a academic question.\\
Your task is to rate the question on one metric.\\
Please make sure you read and understand these instructions carefully.\\
\\
Evaluation Criteria:\\
Score -1: The question is hard to understand from your perspective and cannot be judged.\\
Score 1$\sim$2: Entirely inappropriate: The question assumes a lot of context or prior knowledge unlikely to be had by a junior researcher.\\
Score 3$\sim$4: Mostly inappropriate: The question mostly assumes context or prior knowledge unlikely to be had, but some elements are appropriate.\\
Score 5$\sim$6: Mixed appropriateness: There are elements that are appropriate and inappropriate, containing a roughly equal mix.\\
Score 7$\sim$8: Mostly appropriate: The question mostly assumes context and prior knowledge that is likely to be had by a junior researcher.\\
Score 9$\sim$10: Entirely appropriate: The question could definitely be asked by a junior researcher.\\
\\
You should score 1$\sim$4, including but not limited to the scenarios:\\
(1) the question contains undefined jargon or abbreviations that a junior researcher is unlikely to understand (excluding abbreviations that are considered common knowledge within this domain).\\
(2) the question covers advanced topics that a junior researcher is unlikely to be familiar with.\\
(3) the question is highly contextualized or includes unusual details or conclusions that would require specific context to make sense.\\
\\
You should score 7$\sim$10, including but not limited to the scenarios:\\
(1) the question is general and involves topics that a junior researcher is likely to be familiar with.\\
(2) the question is well-understood independently, without any additional contenxt provided.\\
\\
You should not take the length of question into account. The length of question has nothing to do with the metric.\\
\\
You should be more critical: try to make the best judgment of whether the question leans toward being appropriate or inappropriate for a junior researcher, instead of choosing a safe mediocre score 5$\sim$6.\\
You should be more critical: try to make the best judgment of whether the question leans toward being appropriate or inappropriate for a junior researcher, instead of choosing a safe mediocre score 5$\sim$6.\\
You should be more critical: try to make the best judgment of whether the question leans toward being appropriate or inappropriate for a junior researcher, instead of choosing a safe mediocre score 5$\sim$6.\\
\\
Question: What is the application of computer vision in the field of language processing?\\
Standalone (scores ONLY): -1\\
\\
Question: What are the characteristics and findings related to BERT's contextualized embeddings?\\
Standalone (scores ONLY): 10\\
\\
Question: What are the main differences between graph-based and transition-based dependency parsers?\\
Standalone (scores ONLY): 8\\
\\
Question: How do recent approaches for distant supervision in NLP handle the negative effects of noisy labels?\\
Standalone (scores ONLY): 6\\
\\
Question: How do existing dialogue models for social influence utilize templates, retrieval methods, and conditional generation to produce system responses?\\
Standalone (scores ONLY): 3\\
\\
Question: How do different assumptions affect the recognizability of PARITY by transformer encoders?\\
Standalone (scores ONLY): 1\\
\\
Question: \prompttag{\textbf{\color{insert}<QUESTION>}}\\
Standalone (scores ONLY): 
}}%
\prompt
{Query filtering prompt: answer variability}{
\systemprompt{
You will be given a academic question.\\
Your task is to rate the question on one metric.\\
Please make sure you read and understand these instructions carefully.\\
\\
Evaluation Criteria:\\
Please rate how likely it would be for experts in the area to provide very different responses to the question.\\
Please ignore possible phrasing differences, but instead focus on comparing what information and papers the experts are likely to mention in their response.\\
Assume they are both instructed to provide similar length responses.\\
Evaluate on a scale of 1 to 10, how many possible answers could be given by senior sub-area experts. Use the following definition for the scale.\\
\\
Score -1: The question is difficult to impossible to imagine an answer to, and so cannot be judged.\\
Score 1$\sim$2: Low variability: Most experts are likely to respond with very similar information.\\
Score 3$\sim$4: Moderate variability: Much of the information they respond with will be similar, but may have a few differences.\\
Score 5$\sim$6: Shared core, but variable: The responses are likely to have a shared core, but then each response is likely to have a piece unique to them.\\
Score 7$\sim$8: Mostly Variable: The responses are likely to have overlap in information but we expect each response to have large amounts of unique information.\\
Score 9$\sim$10: Highly Variable: It is likely that the responses will share little to no information in common.\\
\\
You should score 1$\sim$4, including but not limited to the scenarios:\\
(1) the question is asking about some specific techniques details in a narrow setting.\\
\\
You should score 7$\sim$10, including but not limited to the scenarios:\\
(1) the question merely requires to list a bunch of facts or resources, e.g. ``What are the datasets ...'', ``What are the benchmarks ...'', ``What are the applications of ...''.\\
(2) the question asks a overly general method, application, etc. in a broad setting.\\
\\
You should not take the length of question into account. The length of question has nothing to do with the metric.\\
\\
You should be more critical: trying to make the best judge of whether it leans to low or high variability, instead of choosing a safe mediocre score 5$\sim$6.\\
You should be more critical: trying to make the best judge of whether it leans to low or high variability, instead of choosing a safe mediocre score 5$\sim$6.\\
You should be more critical: trying to make the best judge of whether it leans to low or high variability, instead of choosing a safe mediocre score 5$\sim$6.\\
\\
Question: What is the application of computer vision in the field of language processing?\\
Answer Variability (scores ONLY): -1\\
\\
Question: What are the main differences between graph-based and transition-based dependency parsers?\\
Answer Variability (scores ONLY): 1\\
\\
Question: What are the methods for data augmentation in NLP at different levels of text granularity?\\
Answer Variability (scores ONLY): 3\\
\\
Question: What are the major challenges in addressing Conversational Machine Comprehension (CMC)?\\
Answer Variability (scores ONLY): 5\\
\\
Question: What are the characteristics and findings related to BERT's contextualized embeddings?\\
Answer Variability (scores ONLY): 8\\
\\
Question: What are the popular NLP tasks that utilize knowledge graphs (KGs)?\\
Answer Variability (scores ONLY): 10\\
\\
Question: \prompttag{\textbf{\color{insert}<INITIAL\_QUERY>}}\\
Answer Variability (scores ONLY): 
}}%
\prompt
{Query filtering prompt: contains citation}{
\systemprompt{
Does the question \prompttag{\textbf{\color{insert}<INITIAL\_QUERY>}} contain citation or author name? Only reply YES or NO:
}}%
\prompt
{Query generation prompt: generate initial reference answer}{
\systemprompt{
In this task, you will be given a section from an academic paper with a question, and you need to generate an answer using and only using the contents of the section w.r.t. the question.\\
\\
This answer needs to be:\\
1. Decontextualized: remove the contents anchored in external tables, figures, sections, etc., and rephrase the first-person expressions as third person.\\
2. Relevant: remove all other irrelevant details that do not answer the question\\
3. No new information: do not include any information that is not mentioned in the content, except for necessary connective words \\
4. Content preservative: retain the relevant information in the content as closely as possible to the original\\
5. Citation preservative: keep the relevant inline citations in the content exactly the same as in the original content, such as (Author, Year), (Author), Author (Year), or [1], etc.\\
6. Structure preservative: maintain the logical structure of the content such as the order, lists, steps, etc., as in the original text. \\
7. Display the answer sentence by sentence.\\
\\
\# Given extracted content:\\
Paper Title: Societal Biases in Language Generation: Progress and Challenges\\
Section Title: Bias Definitions and Metrics\\
Section Content:\\
In the context of AI fairness, the term ``bias'' commonly refers to skews that result in undesirable impacts (Crawford, 2017) and is quantifiable with some metric. There are relatively more existing studies on biases in NLU tasks, where it is arguably simpler to define bias metrics, since we can intuitively compare the accuracy of the task (e.g., coreference resolution, hate speech detection) for different demographics. Language generation tasks often involve stochastic generation of open-ended and lengthy texts, traits that are not directly compatible with traditional algorithmic bias definitions (e.g., equalized odds, equal opportunity, demographic parity (Dwork et al., 2012; Hardt et al., 2016)).\\
Because of the difficulty in defining metrics, existing works define bias loosely as demographic inequality and use intermediate proxy metrics to comparatively measure bias. Examples include:\\
• Regard Ratio: negative-neutral-positive regard score ratios of text generated from bias-inducing prompts (Sheng et al., 2019)\\
• Sentiment Ratio: negative-neutral-positive sentiment score ratios of text generated from African American English (AAE) versus White-Aligned English (WAE) prompts (Groenwold et al., 2020)\\
• Individual and Group Fairness through Sentiment: comparisons of the sentiment distributions of generated text across demographics and prompts (Huang et al., 2020)\\
• Gendered Word Co-occurrence Score: mean and standard deviations of the absolute log ratio of probabilities: P(word|female terms) to P(word|male terms) across all words in generated text (Bordia and Bowman, 2019)\\
There are also metrics for other bias evaluation setups in continuation generation tasks involving sentiment (Shwartz et al., 2020), the ratio of gendered words (Solaiman et al., 2019; Vig et al., 2020; Dinan et al., 2020a), and other novel metrics (Peng et al., 2020; Yeo and Chen, 2020). Studies of biases in transformation generation tasks favor metrics of accuracy in terms of successfully transforming text to have a desired property. We present a more thorough comparison of metrics in Section 5.4.\\
Bias metrics can also be categorized by how they define associations between demographic group attributes and text. Biases can be towards people described in text, people who produce the text, or people to whom the text is addressed (Dinan et al., 2020b). Most existing works define bias metrics through the first association—these biases are relatively easier to analyze, since both the demographic and the textual signals of bias are encapsulated within the text. There are also works that define biases towards people who produce the text (Groenwold et al., 2020) or people to whom the text is addressed (Sheng et al., 2021b), though there are relatively fewer works that study these latter associations.\\
Question: When generating text, how are the major types of bias measures used to evaluate bias with respect to gender?\\
Answer: \\
Bias metrics can be categorized by how they define associations between demographic group attributes and text. \\
Biases can be towards people described in text, people who produce the text, or people to whom the text is addressed (Dinan et al., 2020b). \\
Most existing works define bias metrics through the first association—these biases are relatively easier to analyze, since both the demographic and the textual signals of bias are encapsulated within the text. \\
There are also works that define biases towards people who produce the text (Groenwold et al., 2020) or people to whom the text is addressed (Sheng et al., 2021b), though there are relatively fewer works that study these latter associations.\\
Language generation tasks often involve stochastic generation of open-ended and lengthy texts, traits that are not directly compatible with traditional algorithmic bias definitions (e.g., equalized odds, equal opportunity, demographic parity (Dwork et al., 2012; Hardt et al., 2016)).\\
Because of the difficulty in defining metrics, existing works define bias loosely as demographic inequality and use intermediate proxy metrics to comparatively measure bias. Examples include:\\
• Regard Ratio: negative-neutral-positive regard score ratios of text generated from bias-inducing prompts (Sheng et al., 2019)\\
• Sentiment Ratio: negative-neutral-positive sentiment score ratios of text generated from African American English (AAE) versus White-Aligned English (WAE) prompts (Groenwold et al., 2020)\\
• Individual and Group Fairness through Sentiment: comparisons of the sentiment distributions of generated text across demographics and prompts (Huang et al., 2020)\\
• Gendered Word Co-occurrence Score: mean and standard deviations of the absolute log ratio of probabilities: P(word|female terms) to P(word|male terms) across all words in generated text (Bordia and Bowman, 2019)\\
There are also metrics for other bias evaluation setups in continuation generation tasks involving sentiment (Shwartz et al., 2020), the ratio of gendered words (Solaiman et al., 2019; Vig et al., 2020; Dinan et al., 2020a), and other novel metrics (Peng et al., 2020; Yeo and Chen, 2020). \\
\\
\# Given extracted content:\\
Paper Title: A Survey on Contextual Embeddings\\
Section Title: Cross-lingual Polyglot Pre-training for Contextual Embeddings\\
Section Content:\\
Cross-lingual polyglot pre-training aims to learn joint multi-lingual representations, enabling knowledge transfer from data-rich languages like English to data-scarce languages like Romanian. Based on whether joint training and a shared vocabulary are used, we divide previous work into three categories.\\
Joint training \& shared vocabulary. Artetxe and Schwenk (2019) use a BiLSTM encoder-decoder framework with a shared BPE vocabulary for 93 languages. The framework is pre-trained using parallel corpora, including as Europarl and Tanzil. The contextual embeddings from the encoder are used to train classifiers using English corpora for downstream tasks. As the embedding space and the encoder are shared, the resultant classifiers can be transferred to any of the 93 languages without further modification. Experiments show that these classifiers achieve competitive performance on cross-lingual natural language inference, cross-lingual document classification, and parallel corpus mining.\\
Rosita (Mulcaire et al., 2019) pre-trains a language model using text from different languages, showing the benefits of polyglot learning on low-resource languages.\\
Recently, the authors of BERT developed a multi-lingual BERT which is pre-trained using the Wikipedia dump with more than 100 languages.\\
XLM (Lample and Conneau, 2019) uses three pre-training methods for learning cross-lingual language models: (1) Causal language modelling, where the model is trained to predict p(ti|t1, t2, ..., ti-1), (2) Masked language modelling, and (3) Translation language modelling (TLM). Parallel corpora are used, and tokens in both source and target sequences are masked for learning cross-lingual association. XLM performs strongly on cross-lingual classification, unsupervised machine translation, and supervised machine translation. XLM-R (Conneau et al., 2019) scales up XLM by training a Transformer-based masked language model on one hundred languages, using more than two terabytes of filtered CommonCrawl data. XLM-R shows that large-scale multi-lingual pre-training leads to significant performance gains for a wide range of cross-lingual transfer tasks.\\
Joint training \& separate vocabularies. Wu et al. (2019) study the emergence of cross-lingual structures in pre-trained multi-lingual language models. It is found that cross-lingual transfer is possible even when there is no shared vocabulary across the monolingual corpora, and there are universal latent symmetries in the embedding spaces of different languages.\\
Separate training \& separate vocabularies. Artetxe et al. (2019) use a four-step method for obtaining multi-lingual embeddings. Suppose we have the monolingual sequences of two languages L1 and L2: (1) Pre-training BERT with the vocabulary of L1 using L1’s monolingual data. (2) Replacing the vocabulary of L1 with the vocabulary of L2 and training new vocabulary embeddings, while freezing the other parameters, using L2’s monolingual data. (3) Fine-tuning the BERT model for a downstream task using labeled data in L1, while freezing L1’s vocabulary embeddings. (4) Replacing the fine-tuned BERT with L2’s vocabulary embeddings for zero-shot transfer tasks.\\
Question: How do multilingual NLP models handle joint vocabularies during pretraining?\\
Answer: \\
Cross-lingual polyglot pre-training aims to learn joint multi-lingual representations, enabling knowledge transfer from data-rich languages like English to data-scarce languages like Romanian. \\
In some cases, shared vocabularies combined with multilingual pretraining are used to achieve transfer to lower resource languages.  \\
For example, Artetxe and Schwenk (2019) use a BiLSTM encoder-decoder framework with a shared BPE vocabulary for 93 languages. \\
The framework is pre-trained using parallel corpora, including as Europarl and Tanzil. \\
The contextual embeddings from the encoder are used to train classifiers using English corpora for downstream tasks. \\
As the embedding space and the encoder are shared, the resultant classifiers can be transferred to any of the 93 languages without further modification. \\
Experiments show that these classifiers achieve competitive performance on cross-lingual natural language inference, cross-lingual document classification, and parallel corpus mining.\\
In another examples, Rosita (Mulcaire et al., 2019) pre-trains a language model using text from different languages, showing the benefits of polyglot learning on low-resource languages.\\
Also, recently, the authors of BERT developed a multi-lingual BERT which is pre-trained using the Wikipedia dump with more than 100 languages.\\
XLM (Lample and Conneau, 2019) uses three pre-training methods for learning cross-lingual language models: (1) Causal language modelling, where the model is trained to predict p(ti|t1, t2, ..., ti-1), (2) Masked language modelling, and (3) Translation language modelling (TLM). \\
Parallel corpora are used, and tokens in both source and target sequences are masked for learning cross-lingual association. \\
XLM performs strongly on cross-lingual classification, unsupervised machine translation, and supervised machine translation. \\
XLM-R (Conneau et al., 2019) scales up XLM by training a Transformer-based masked language model on one hundred languages, using more than two terabytes of filtered CommonCrawl data. XLM-R shows that large-scale multi-lingual pre-training leads to significant performance gains for a wide range of cross-lingual transfer tasks.\\
\\
\# Given extracted content:\\
Paper Title: A Primer on Contrastive Pretraining in Language Processing: Methods, Lessons Learned and Perspectives\\
Section Title: Noise Contrastive Estimation (NCE)\\
Section Content:\\
Noise contrastive estimation is the objective used by most contrastive learning approaches within NLP. Thus, we briefly outline its main variants and the core ideas behind them, while pointing to (Ma and Collins, 2018)1 for detailed, yet readily understandable explanations of the two main NCE variants. Both variants can intuitively be understood as a sub-sampled softmax with K negative samples a-i and one positive sample a+i . The first variant expresses NCE as a binary objective (loss) in the form of maximum log likelihood, where only K negatives are considered.\\
LB($\theta$, $\gamma$) = log $\sigma$(s(xi, a+i,0; $\theta$), $\gamma$) + K$\Sigma$k=1 log(1- $\sigma$(s(xi, a-i,k; $\theta$), $\gamma$)\\
Here, s(xi, ai,◦; $\theta$) is a scoring or similarity function that measures the compatibility between a single text input xi and another sample ai,◦. As mentioned above, the sample can be another input text or an output label (text), thus modeling NLP tasks as ‘text-to-text’ prediction similar to language models. The similarity function is typically a cosine similarity, a dot product or a logit (unscaled activation) produced by a input-sample matcher sub-network (Rethmeier and Augenstein, 2020). The $\sigma$(z, $\gamma$) is a scaling function, which for use in eq. (1) is typically the sigmoid $\sigma$(z) = exp(z - $\gamma$)/(1 + exp(z - $\gamma$)) with a hyperparameter $\gamma$ $\geq$ 0 (temperature), that is tuned or omitted depending on the way that negative samples a-i are attained.\\
The other NCE objective learns to rank a single positive pair (xi, a+i,0) over K negative pairs (xi, a-i,k):\\
LR($\theta$) = log e$\bar{\text{s}}$(xi, a+i,0; $\theta$) e$\bar{\text{s}}$(xi, a+i,0; $\theta$) + $\Sigma$Kk=1 e$\bar{\text{s}}$(xi, a-i,k; $\theta$)\\
Here, to improve LR or LB performance, (Ma and Collins, 2018) propose a regularized scoring function $\bar{\text{s}}$(xi, ai,◦) = s(xi, ai,◦) - log pN (ai,◦) that subtracts the probability of the current sample ai,◦ under a chosen noise distribution pN (ai,◦). In practice, the noise distribution can be set to 0 (Mnih and Teh, 2012; Wu et al., 2020; Rethmeier and Augenstein, 2020) to save on computation. To robustly learn word embeddings, pN (ai,◦) can be set as the word probability pword in a corpus (Mikolov et al., 2013b), or as the probability of a sequence under a language model pLM (Deng et al., 2020), when learning contrastive sequence prediction.\\
Generalization to an arbitrary number of positives: As (Khosla et al., 2020) mention, original contrastive formulations use only one positive pair per text instance (see e.g. (Mikolov et al., 2013b; Logeswaran and Lee, 2018)), while more recent methods mine multiple positives or use multiple gold class annotation representations for contrastive learning (Rethmeier and Augenstein, 2020; Qu et al., 2021). This means that e.g. the positive term in eq. (1) becomes $\Sigma$Pp=1 log $\sigma$(s(xi, a+i,p; $\theta$, $\gamma$)) to consider P positives.\\
Importance of negative sampling semantics and lessons learned: How positive and negative samples are generated or sampled is a key component of effective contrastive learning. (Saunshi et al., 2019) prove and empirically validate that ``sampling more negatives improves performance, but only if they are sampled from the same context or block of information such as the same paragraph''. Such hard to contrast (classify) negatives, are sampled in most works (Mikolov et al., 2013b; Saunshi et al., 2019; Rethmeier and Augenstein, 2020; Iter et al., 2020). Otherwise, performance can deteriorate due to weak contrast learning of conceptually related classes. Additionally, (Rethmeier and Augenstein, 2020) find that both positive and negative contrastive samples from a long-tail distribution are essential in predicting rare classes and in substantially boosting zero-shot performance, especially over minority classes. (Mikolov et al., 2013b) undersample negatives of frequent words to stabilize pretraining of word embeddings to a similar effect. Additional practical advice for negative sampling is mentioned in 3.1.\\
Question: In contrastive models for nlp, how do the choices of negatives samples influence the quality of representions?\\
Answer: \\
Noise contrastive estimation is the objective used by most contrastive learning approaches within NLP. \\
How positive and negative samples are generated or sampled is a key component of effective contrastive learning. \\
(Saunshi et al., 2019) prove and empirically validate that ``sampling more negatives improves performance, but only if they are sampled from the same context or block of information such as the same paragraph''. \\
Such hard to contrast (classify) negatives, are sampled in most works (Mikolov et al., 2013b; Saunshi et al., 2019; Rethmeier and Augenstein, 2020; Iter et al., 2020). \\
Otherwise, performance can deteriorate due to weak contrast learning of conceptually related classes. \\
Additionally, (Rethmeier and Augenstein, 2020) find that both positive and negative contrastive samples from a long-tail distribution are essential in predicting rare classes and in substantially boosting zero-shot performance, especially over minority classes. \\
(Mikolov et al., 2013b) undersample negatives of frequent words to stabilize pretraining of word embeddings to a similar effect. \\
\\
\# Given extracted content:\\
Paper Title: Modeling Language Variation and Universals: A Survey on Typological Linguistics for Natural Language Processing\\
Section Title: Hand-Crafted Documentation in Typological Databases\\
Section Content:\\
Typological databases are created manually by linguists. They contain taxonomies of typological features, their possible values, as well as the documentation of feature values for the world’s languages. Major typological databases, listed in Table 1, typically organize linguistic information in terms of universal features and language-specific values. For example, Figure 3 presents language-specific values for the feature number of grammatical genders for nouns on a world map. Note that each language is color-coded according to its value. Further examples for each database can be found in the rightmost column of Table 1.\\
Some databases store information pertaining to multiple levels of linguistic description. These include WALS (Dryer and Haspelmath 2013) and the Atlas of Pidgin and Creole Language Structures (APiCS) (Michaelis et al. 2013). Among all presently available databases, WALS has been the most widely used in NLP. In this resource, which has 142 typological features in total, features 1–19 deal with phonology, 20–29 with morphology, 30–57 with nominal categories, 58–64 with nominal syntax, 65–80 with verbal categories, 81–97 and 143–144 with word order, 98–121 with simple clauses, 122–128 with complex sentences, 129–138 with the lexicon, and 139–142 with other properties.\\
Other databases only cover features related to a specific level of linguistic description. For example, both Syntactic Structures of the World’s Languages (SSWL) (Collins and Kayne 2009) and AUTOTYP (Bickel et al. 2017) focus on syntax. SSWL features are manually crafted, whereas AUTOTYP features are derived automatically from primary linguistic data using scripts. The Valency Patterns Leipzig (ValPaL) (Hartmann, Haspelmath, and Taylor 2013) provides verbs as attributes and predicate–argument structures as their values (including both valency and morphosyntactic constraints). For example, in both Mandinka and Sliammon, the verb to laugh has a valency of 1; in other words, it requires only one mandatory argument, the subject. In Mandinka the subject precedes the verb, but there is no agreement requirement; in Sliammon, on the other hand, the word order does not matter, but the verb is required to morphologically agree with the subject.\\
For phonology, the Phonetics Information Base and Lexicon (PHOIBLE) (Moran, McCloy, and Wright 2014) collates information on segments (binary phonetic features). In the Lyon–Albuquerque Phonological Systems Database (LAPSyD) (Maddieson et al. 2013), attributes are articulatory traits, syllabic structures, or tonal systems. Finally, StressTyp2 (Goedemans, Heinz, and der Hulst 2014) deals with stress and accent patterns. For instance, in Koromfé each word’s first syllable has to be stressed, but not in Cubeo.\\
Other databases document various aspects of semantics. The World Loanword Database (WOLD) (Haspelmath and Tadmor 2009) documents loanwords by identifying the donor languages and the source words. The Automated Similarity Judgment Program (ASJP) (Wichmann, Holman, and Brown 2016) and the Intercontinental Dictionary Series (IDS) (Key and Comrie 2015) indicate how a meaning is lexicalized across languages: For example, the concept of WORLD is expressed as mir in Russian, and as ārkiśos. i in Tocharian A.\\
Although typological databases store abundant information on many languages, they suffer from shortcomings that limit their usefulness. Perhaps the most significant shortcoming of such resources is their limited coverage. In fact, feature values are missing for most languages in most databases. Other shortcomings are related to feature granularity. In particular, most databases fail to account for feature value variation within each language: They report only majority value rather than the full range of possible values and their corresponding frequencies. For example, the dominant adjective–noun word order in Italian is adjective before noun; however, the opposite order is also attested. The latter information is often missing from typological databases.\\
Further challenges are posed by restricted feature applicability and feature hierarchies. Firstly, some features apply, by definition, only to subsets of languages that share another feature value. For instance, WALS feature 113A documents ``Symmetric and Asymmetric Standard Negation,'' whereas WALS feature 114A ``Subtypes of Asymmetric Standard Negation.'' Although a special NA value is assigned for symmetric-negation languages in the latter, there are cases where languages without the prerequisite feature are simply omitted from the sample. Secondly, features can be partially redundant, and subsume other features. For instance, WALS feature 81A ``Order of Subject, Object and Verb'' encodes the same information as WALS feature 82A ``Order of Subject and Verb'' and 83A ``Order of Object and Verb,'' with the addition of the order of subject and object.\\
Question: What are the differences between publicly available linguistic typology databases?\\
Answer: \\
Some typological databases store information pertaining to multiple levels of linguistic description. \\
These include WALS (Dryer and Haspelmath 2013) and the Atlas of Pidgin and Creole Language Structures (APiCS) (Michaelis et al. 2013). \\
Among all presently available databases, WALS has been the most widely used in NLP. Other databases only cover features related to a specific level of linguistic description. \\
For example, both Syntactic Structures of the World’s Languages (SSWL) (Collins and Kayne 2009) and AUTOTYP (Bickel et al. 2017) focus on syntax. \\
SSWL features are manually crafted, whereas AUTOTYP features are derived automatically from primary linguistic data using scripts. \\
The Valency Patterns Leipzig (ValPaL) (Hartmann, Haspelmath, and Taylor 2013) provides verbs as attributes and predicate–argument structures as their values (including both valency and morphosyntactic constraints). \\
For phonology, the Phonetics Information Base and Lexicon (PHOIBLE) (Moran, McCloy, and Wright 2014) collates information on segments (binary phonetic features). \\
In the Lyon–Albuquerque Phonological Systems Database (LAPSyD) (Maddieson et al. 2013), attributes are articulatory traits, syllabic structures, or tonal systems. \\
Finally, StressTyp2 (Goedemans, Heinz, and der Hulst 2014) deals with stress and accent patterns.\\
Other databases document various aspects of semantics. The World Loanword Database (WOLD) (Haspelmath and Tadmor 2009) documents loanwords by identifying the donor languages and the source words. \\
The Automated Similarity Judgment Program (ASJP) (Wichmann, Holman, and Brown 2016) and the Intercontinental Dictionary Series (IDS) (Key and Comrie 2015) indicate how a meaning is lexicalized across languages.\\
\\
\# Given extracted content:\\
Paper Title: Efficient Methods for Natural Language Processing: A Survey\\
Section Title: Sparse Modeling\\
Section Content:\\
To leverage sparsity for efficiency, many models follow the mixture-of-experts (MoE) concept (Jacobs et al., 1991; Shazeer et al., 2017; Fedus et al., 2022a), which routes computation through small subnetworks instead of passing the input through the entire model. Relevant works on this line include GShard (Lepikhin et al., 2021), Switch Transformer (Fedus et al., 2022b), and ST-MoE (Zoph et al., 2022), which replace the feed forward layers in transformers with MoE layers. More recently, Rajbhandari et al. (2022) scaled transformers up by compressing and optimizing the usage of MoE. Overall, MoE models have been shown to achieve strong performance across several NLP tasks while reducing the overall resource consumption (Sec. 8). For instance, GLaM (Du et al., 2022) used only $\sim$1 3 of GPT-3’s energy consumption (with additional hardware-based optimization), while Rajbhandari et al. (2022) reached a 5x reduction in terms of training cost. However, MoE models have also exhibited training instabilities in practice, and may require architecture-specific implementation (Zoph et al., 2022; Mustafa et al., 2022).\\
Another promising direction for exploiting sparse modeling is Sparsefinder (Treviso et al., 2022), which extends the Adaptively Sparse Transformer (Correia et al., 2019) to allow a more efficient attention mechanism by identifying beforehand the sparsity pattern returned by entmax attention—a sparse alternative to (dense) softmax attention (Peters et al., 2019). Finally, sparsity can also be induced via modularity, e.g., by encapsulating task-specific parameters (Ponti et al., 2022).\\
Question: How can we utilize sparsity to enhance efficiency in designing NLP models?\\
Answer:\\
To leverage sparsity for efficiency, many models follow the mixture-of-experts (MoE) concept (Jacobs et al., 1991; Shazeer et al., 2017; Fedus et al., 2022a), which routes computation through small subnetworks instead of passing the input through the entire model. \\
Relevant works on this line include GShard (Lepikhin et al., 2021), Switch Transformer (Fedus et al., 2022b), and ST-MoE (Zoph et al., 2022), which replace the feed forward layers in transformers with MoE layers. \\
More recently, Rajbhandari et al. (2022) scaled transformers up by compressing and optimizing the usage of MoE. \\
Overall, MoE models have been shown to achieve strong performance across several NLP tasks while reducing the overall resource consumption. \\
For instance, GLaM (Du et al., 2022) used only $\sim$1/3 of GPT-3’s energy consumption (with additional hardware-based optimization), while Rajbhandari et al. (2022) reached a 5x reduction in terms of training cost. \\
However, MoE models have also exhibited training instabilities in practice, and may require architecture-specific implementation (Zoph et al., 2022; Mustafa et al., 2022).\\
Another promising direction for exploiting sparse modeling is Sparsefinder (Treviso et al., 2022), which extends the Adaptively Sparse Transformer (Correia et al., 2019) to allow a more efficient attention mechanism by identifying beforehand the sparsity pattern returned by entmax attention—a sparse alternative to (dense) softmax attention (Peters et al., 2019). \\
Finally, sparsity can also be induced via modularity, e.g., by encapsulating task-specific parameters (Ponti et al., 2022).\\
\\
\# Given extracted content:\\
Paper Title: Neural Approaches to Conversational AI\\
Section Title: Speaker Consistency\\
Section Content:\\
It has been shown that the popular seq2seq approach often produces conversations that are incoherent (Li et al., 2016b), where the system may for instance contradict what it had just said in the previous turn (or sometimes even in the same turn). While some of this effect can be attributed to the limitation of the learning algorithms, Li et al. (2016b) suggested that the main cause of this inconsistency is probably due to the training data itself. Indeed, conversational datasets (see Sec. 5.5) feature multiple speakers, which often have different or conflicting personas and backgrounds. For example, to the question ``how old are you?'', a seq2seq model may give valid responses such as ``23'', ``27'', or ``40'', all of which are represented in the training data.\\
This sets apart the response generation task from more traditional NLP tasks: While models for other tasks such as machine translation are trained on data that is mostly one-to-one semantically, conversational data is often one-to-many or many-to-many as the above example implies.5 As one-to-many training instances are akin to noise to any learning algorithm, one needs more expressive models that exploits a richer input to better account for such diverse responses.\\
To do this, Li et al. (2016b) proposed a persona-based response generation system, which is an extension of the LSTM model of Sec. 5.1.1 that uses speaker embeddings in addition to word embeddings. Intuitively, these two types of embeddings work similarly: while word embeddings form a latent space in which spacial proximity (i.e., low Euclidean distance) means two words are semantically or functionally close, speaker embeddings also constitute a latent space in which two nearby speakers tend to converse in the same way, e.g., having similar speaking styles (e.g., British English) or often talking about the same topic (e.g., sports).\\
Like word embeddings, speaker embedding parameters are learned jointly with all other parameters of the model from their one-hot representations. At inference time, one just needs to specify the one-hot encoding of the desired speaker to produce a response that reflects her speaking style. The global architecture of the model is displayed in Fig. 5.2, which shows that each target hidden state is conditioned not only on the previous hidden state and the current word embedding (e.g., ``England''), but also on the speaker embedding (e.g., of ``Rob''). This model not only helps generate more personalized responses, but also alleviates the one-to-many modeling problem mentioned earlier.\\
Other approaches also utilized personalized information. For example, Al-Rfou et al. (2016) presented a persona-based response generation model, but geared for retrieval using an extremely large dataset consisting of 2.1 billion responses. Their retrieval model is implemented as a binary classifier (i.e., good response or not) using a deep neural network. The distinctive feature of their model is a multi-loss objective, which augments a single-loss model p(R|I, A,C) of the response R, input I, speaker (``author'') A, and context C, by adding auxiliary losses that, e.g., model the probability of the response given the author p(R|A). This multi-loss model was shown to be quite helpful (Al-Rfou et al., 2016), as the multiple losses help cope with the fact that certain traits of the author are often correlated with the context or input, which makes it difficult to learn good speaker embedding representation. By adding a loss for p(R|A), the model is able to learn a more distinctive speaker embedding representation for the author.\\
More recently, Luan et al. (2017) presented an extension of the speaker embedding model of Li et al. (2016b), which combines a seq2seq model trained on conversational datasets with an autoencoder trained on non-conversational data, where the seq2seq and autoencoder are combined in a multi-task learning setup (Caruana, 1998). The tying of the decoder parameters of both seq2seq and autoencoder enables Luan et al. (2017) to train a response generation system for a given persona without actually requiring any conversational data available for that persona. This is an advantage of their approach, as conversational data for a given user or persona might not always be available. In (Bhatia et al., 2017), the idea of (Li et al., 2016b) is extended to a social-graph embedding model.\\
While (Serban et al., 2017) is not a persona-based response generation model per se, their work shares some similarities with speaker embedding models such as (Li et al., 2016b). Indeed, both Li et al. (2016b) and Serban et al. (2017) introduced a continuous high-dimensional variable in the target side of the model in order to bias the response towards information encoded in a vector. In the case of (Serban et al., 2017), that variable is latent, and is trained by maximizing a variational lower-bound on the log-likelihood. In the case of (Li et al., 2016b), the variable (i.e., the speaker embedding) is technically also latent, although it is a direct function of the one-hot representation of speaker. (Li et al., 2016b) might be a good fit when utterance-level information (e.g., speaker ID or topic) is available. On the other hand, the strength of (Serban et al., 2017) is that it learns a latent variable that best ``explains'' the data, and may learn a representation that is more optimal than the one based strictly on speaker or topic information.\\
Question: Why do conversation models often produce responses that are inconsistent with previous turns?\\
Answer: \\
It has been shown that the popular seq2seq approach often produces conversations that are incoherent (Li et al., 2016b), where the system may for instance contradict what it had just said in the previous turn (or sometimes even in the same turn). \\
While some of this effect can be attributed to the limitation of the learning algorithms, Li et al. (2016b) suggested that the main cause of this inconsistency is probably due to the training data itself. \\
Conversational datasets feature multiple speakers, which often have different or conflicting personas and backgrounds. \\
For example, to the question ``how old are you?'', a seq2seq model may give valid responses such as ``23'', ``27'', or ``40'', all of which are represented in the training data.\\
This sets apart the response generation task from more traditional NLP tasks: While models for other tasks such as machine translation are trained on data that is mostly one-to-one semantically, conversational data is often one-to-many or many-to-many as the above example implies.\\
As one-to-many training instances are akin to noise to any learning algorithm, one needs more expressive models that exploits a richer input to better account for such diverse responses.\\
\\
\# Given extracted content:\\
Paper Title: \prompttag{\textbf{\color{insert}<PAPER\_TITLE>}}\\
Section Title: \prompttag{\textbf{\color{insert}<SECTION\_TITLE>}}\\
Section Content:\\
\prompttag{\textbf{\color{insert}<SECTION\_CONTENT>}}\\
Question: \prompttag{\textbf{\color{insert}<INITIAL\_QUERY>}}\\
Answer: 
}}%
\prompt
{Query generation prompt: rephrase query and reference answer for cohesion}{
\systemprompt{
Original question and answer: \\
Question: \prompttag{\textbf{\color{insert}<INITIAL\_QUERY>}}\\
Answer:\\
\prompttag{\textbf{\color{insert}<INITIAL\_REF\_ANSWER>}}\\
\\
---\\
Rewrite the question and answer to make them more coherent so that it sounds like an answer that an human expert would answer when being asked this question. \\
Try to use the contents of the answer as much as possible, and keep the citations. \\
Focus on getting a better organization and transition of the contents, without adding new sentences that are not in the answer.\\
Output in a format of ``Question:\textbackslash{n}\textbackslash{n}{question}\textbackslash{n}\textbackslash{n}Answer:\textbackslash{n}\textbackslash{n}{answer}'' is required.\\
\\
Rephrased question and answer: 
}}%
\prompt
{Parametric rubric generation prompt: information-based item}{
\systemprompt{
You will be shown a query asked by a junior PhD researcher and a referral response to this query.\\
Imagine that you are required to write another response to this open-ended query,\\
you need to ask some follow-up questions to know the preferred content in the response or resolve ambiguity in the query.\\
\\
Here are criteria that individual questions need to satisfy:\\
- relevant: The question should ask about information that is related to the query, and would lower the answer variability of the query's response.\\
- salient: The question should address important piece to guide the writing of query's response in a way that an experienced research would like to.\\
- binary: All questions should be yes-no binary questions.\\
- qualitative: The question should be qualitative and focus on the big picture and important aspects, but not the nonessentials or specific numbers.\\
\\
Here are the criteria that the list of questions needs to satisfy:\\
- knowledge-cutoff: The question should not ask for or rely on any information that would violate the specified knowledge cutoff date\\
- sufficient: There should be enough important questions to cover a large space of possible contexts for the query.\\
- coverage: The questions in combinations should cover every potential important aspects.\\
\\
Specifically, here, you are required to generate the question of information type: \\
For instance, Does the response include key findings A and B? ... Example E?\\
It is to ask whether an important information exist in the response, such as a statement, a finding, an opinion, a comparison, etc..\\
It is not to ask a depth question (how to demonstrate an important point in deep), nor a citation question (whether a citation exists).\\
You should first think about what information is necessary to be included in a good response to the query, and then ask the corresponding questions.\\
\\
Generate up to 10 questions (but no need to) and they should all meet the above criteria.\\
You should generate questions that are important and useful, and address a *necessary* information perspective of the query's response.\\
Please make sure that your questions are relevant, salient, binary, qualitative, and grounded. \\
Please list the questions from the most to the least necessary to be in the query's response.\\
\\
Query: What are the latest works on finetuning an auto-regressive LM for dense passage retrieval and how are their performance compared with bi-directional encoders?\\
Date: 2024-11-21 \\
Questions:\\
1. Does the response to include examples of state-of-the-art auto-regressive LMs that outperform bi-directional encoders in retrieval tasks?\\
2. Dose the response highlight hybrid approaches that combine auto-regressive and bi-directional features?\\
3. Does the response emphasize the specific benchmarks (e.g., MTEB) where auto-regressive LMs outperform bi-directional encoders? \\
4. Does the response mention the importance to achieve full potential of the decoder models by appropriate optimization? \\
\\
Query: \prompttag{\textbf{\color{insert}<QUERY>}}\\
Date: \prompttag{\textbf{\color{insert}<DATE\_CUTOFF>}}\\
Questions:
}}%
\prompt
{Parametric rubric generation prompt: depth-based item}{
\systemprompt{
You will be shown a query asked by a junior PhD researcher and a referral response to this query.\\
Imagine that you are required to write another response to this open-ended query,\\
you need to ask some follow-up questions to know the preferred content in the response or resolve ambiguity in the query.\\
\\
Here are criteria that individual questions need to satisfy:\\
- relevant: The question should ask about information that is related to the query, and would lower the answer variability of the query's response.\\
- salient: The question should address important pieces to guide the writing of the query's response in a way that an experienced researcher would like to.\\
- binary: All questions should be yes-no binary questions.\\
- general: The question should not mention a specific paper's name (e.g. Author and year), nor narrow examples or demonstrations.\\
\\
Here are the criteria that the list of questions needs to satisfy:\\
- knowledge-cutoff: The question should not ask for or rely on any information that would violate the specified knowledge cutoff date\\
- sufficient: There should be enough important questions to cover a large space of possible contexts for the query.\\
- coverage: The questions in combinations should cover every potential important aspect.\\
\\
Specifically, here, you are required to generate the question of depth type (e.g., Does the response explore area C1 and C2 in detail so that to better support the claim C?).\\
It is to ask whether information exists in the response that is explored in depth, such as detailed extention or explanation. Generally it means the response takes a longer-than-average length to explore a topic.\\
It is not to ask an information question (merely whether a statement, example, finding, opinion etc. exists), instead it requires depth in the contents.\\
You should first find the depth information that was actually included in the given referral response, and then ask the corresponding questions.\\
\\
Generate up to 5 questions (but no need to) and they should all meet the above criteria.\\
You should generate questions that are important and useful.\\
\\
Query: What are the latest works on fine-tuning an auto-regressive LM for dense passage retrieval and how are their performance compared with bi-directional encoders?\\
Date: 2024-11-21 \\
Questions:\\
1. Does the response compare auto-regressive model and bi-directional model on the dense passage retrieval in detail?\\
2. Does the response elaborate on the hybrid approach of GRIT in unifying auto-regressive and bi-directional features?\\
\\
Query: \prompttag{\textbf{\color{insert}<QUERY>}}\\
Date: \prompttag{\textbf{\color{insert}<DATE\_CUTOFF>}}\\
Questions:
}}%
\prompt
{Parametric rubric generation prompt: citation-based item}{
\systemprompt{
You will be shown a query asked by a junior PhD researcher and a referral response to this query.\\
Imagine that you are required to wrtien another response to this open-ended query,\\
you need to ask some follow-up questions to know the preferred content in the response or resolve ambiguity in the query.\\
\\
Here are criteria that individual questions need to satisfy:\\
- relevant: The question should ask about information that is related to the query, and would lower the answer variability of the query's response.\\
- salient: The question should address important piece to guide the writing of query's response in a way that an experienced research would like to.\\
- binary: All questions should be yes-no binary questions.\\
- qualitative: The question should be qualitative and focus on the big picture and important aspects, but not the nonessentials or specific numbers.\\
\\
Here are the criteria that the list of questions needs to satisfy:\\
- knowledge-cutoff: The question should not ask for or rely on any information that would violate the specified knowledge cutoff date\\
- sufficient: There should be enough important questions to cover a large space of possible contexts for the query.\\
- coverage: The questions in combinations should cover every potential important aspects.\\
\\
Specifically, here, you are required to generate the question of citation type.\\
For instance, Does the response cite the paper by {author} (year) (title: {title}) that back the statement A by explaining the detail A1 and A2?\\
It is to ask whether an important citation exist in the response, and how it is used in the response. \\
It must mention the title of the paper in a way ``(title: {title})'', and the details of the statement that the citation is used to back up.\\
You should first think of the important citations that are necessary for the query's response, and then generate the questions that ask about them.\\
\\
Generate up to 5 questions (but no need to) and they should all meet the above criteria.\\
You should generate questions that are important and useful, and address a *necessary* citation perspective of the query's response.\\
Please make sure that your questions are relevant, salient, binary, and qualitative. \\
Please list the questions from the most to the least necessary to be in the query's response.\\
\\
Query: What are the latest works on finetuning an auto-regressive LM for dense passage retrieval and how are their performance compared with bi-directional encoders?\\
Date: 2024-11-21 \\
Questions:\\
1. Does the response cite some papers such as Wang et al. (2023) (title: Improving text embeddings with large language models) and the MTEB paper (title: Massive Text Embedding Benchmark) that show the performance of auto-regressive LMs surpassing bi-directional encoders in retrieval tasks?\\
2. Does the response cite the GRIT model paper (title: Generative Representational Instruction Tuning) that shows how to leverage the strengths of both auto-regressive and bi-directional model by incorporating two distinct language modeling heads atop the Transformer layers?\\
\\
Query: \prompttag{\textbf{\color{insert}<QUERY>}}\\
Date: \prompttag{\textbf{\color{insert}<DATE\_CUTOFF>}}\\
Questions:
}}%
\prompt
{Survey rubric generation prompt: information-based item}{
\systemprompt{
You will be shown a query asked by a junior PhD researcher and a referral response to this query.\\
Imagine that you are required to write another response to this open-ended query,\\
you need to ask some follow-up questions to know the preferred content in the response or resolve ambiguity in the query.\\
The referral response is written by experienced research, and is perfect for generating these follow-up questions to the query.\\
\\
Here are criteria that individual questions need to satisfy:\\
- relevant: The question should ask about information that is related to the query, and would lower the answer variability of the query's response.\\
- salient: The question should address important pieces to guide the writing of the query's response in a way that an experienced researcher would like to.\\
- binary: All questions should be yes-no binary questions.\\
- general: The question should not mention a specific paper's name (e.g. Author and year), nor narrow examples or demonstrations.\\
- grounded: These questions should be strictly based on the referral response.\\
\\
Here are the criteria that the list of questions needs to satisfy:\\
- sufficient: There should be enough important questions to cover a large space of possible contexts for the query.\\
- coverage: The questions in combinations should cover every important aspect in the referral response.\\
\\
Specifically, here, you are required to generate the question of information type (e.g., Does the response include key findings A and B? ... Example E?).\\
It is to ask whether important information exists in the response, such as a statement, a finding, an example, an opinion, a comparison, etc..\\
It is not to ask a depth question (whether a detailed extension or explanation exists), nor a citation question (whether a citation exists).\\
You should first find the information that was actually included in the given referral response, and then ask the corresponding questions.\\
\\
Generate up to 10 questions (but no need to) and they should all meet the above criteria.\\
You should generate questions that are important and useful.\\
\\
Query-Response Pair:\\
Query: \\
\\
What are the latest works on finetuning an auto-regressive LM for dense passage retrieval and how are their performance compared with bi-directional encoders?\\
\\
Response:\\
\\
It was traditionally assumed that decoder models would underperform compared to bi-directional autoregressive models, as referenced in {[9]}. However, recent advancements have challenged this notion, demonstrating that finetuned auto-regressive language models (LMs) can indeed surpass the capabilities of bi-directional encoder models in retrieval embedding tasks {[1]}{[2]}{[3]}{[4]}.\\
\\
One notable example is the E5-Mixtral model {[1]}, which employs a novel approach by generating a diverse set of synthetic data to finetune Mistral, a pre-trained large language model (LLM). This model has shown superior performance over the state-of-the-art (SOTA) bi-directional embedding models of that time, including OpenAI's text-embedding-3-large, Cohere-embed-english-v3.0, and voyage-lite-01-instruct on the MTEB benchmark {[8]}.\\
\\
Another study LLM2Vec {[2]} demonstrates the adaptation of a decoder model solely using public data for embedding tasks, which also achieved remarkable results on the MTEB benchmark {[8]}, surpassing other encoder models by a significant margin. This highlights the potential of decoder models when optimized appropriately.\\
\\
Furthermore, the GRIT model {[3]} unifies auto-regressive and bi-directional model designs. It incorporates two distinct language modeling heads atop the Transformer layers: one auto-regressive head, designed for generative tasks with a causal attention mask, and another bi-directional head, tailored for embedding tasks. This dual-head approach allows GRIT to leverage the strengths of both modeling techniques.\\
\\
Lastly, the NV-Embed model {[4]} adapts the decoder model architecture to enhance finetuning effectiveness using public datasets. This model not only ranks at the top on the MTEB benchmark {[8]} but also outperforms all existing baselines on the information retrieval benchmark {[5]}, underscoring the robustness and versatility of decoder models in handling complex language tasks.\\
\\
In short, E5-Mixtral {[1]}, LLM2Vec {[2]}, GRIT {[3]}, and NV-Embed {[4]} are good representations of the SOTA auto-regressive LMs on retrieval tasks whose performance is superior than traditional bi-directional encoders {[5]}{[6]}{[7]}{[8]}.\\
\\
{[1]} Improving text embeddings with large language models. (2023)\\
{[2]} LLM2Vec: Large Language Models Are Secretly Powerful Text Encoders (2024)\\
{[3]} Generative Representational Instruction Tuning (2024)\\
{[4]} NV-Embed: Improved Techniques for Training LLMs as Generalist Embedding Models (2024)\\
{[5]} NV-Embed: Improved Techniques for Training LLMs as Generalist Embedding Models (2024)\\
{[6]} Improving text embeddings with large language models. (2023)\\
{[7]} Generative Representational Instruction Tuning (2024)\\
{[8]} MTEB: Massive Text Embedding Benchmark (2022)\\
{[9]} NV-Embed: Improved Techniques for Training LLMs as Generalist Embedding Models (2024)\\
Questions:\\
1. Does the response include examples of state-of-the-art auto-regressive LMs such as E5-Mixtral, LLM2Vec, GRIT, and NV-Embed? \\
2. Does the response highlight hybrid approaches like GRIT that combine auto-regressive and bi-directional features? \\
3. Does the response discuss details about the novel techniques used in these models, such as synthetic data generation? \\
4. Does the response emphasize the specific benchmarks (e.g., MTEB) where auto-regressive LMs outperform bi-directional encoders? \\
5. Does the response address the role of public datasets in training models like LLM2Vec and NV-Embed? \\
6. Does the response mention the importance to achieve full potential of the decoder models by appropriate optimization? \\
\\
Query-Response Pair:\\
Query:\\
\prompttag{\textbf{\color{insert}<QUERY>}}\\
Response:\\
\prompttag{\textbf{\color{insert}<REF\_ANSWER>}}\\
Questions:
}}%
\prompt
{Survey rubric generation prompt: depth-based item}{
\systemprompt{
You will be shown a query asked by a junior PhD researcher and a referral response to this query.\\
Imagine that you are required to write another response to this open-ended query,\\
you need to ask some follow-up questions to know the preferred content in the response or resolve ambiguity in the query.\\
The referral response is written by experienced research, and is perfect for generating these follow-up questions to the query.\\
\\
Here are criteria that individual questions need to satisfy:\\
- relevant: The question should ask about information that is related to the query, and would lower the answer variability of the query's response.\\
- salient: The question should address important pieces to guide the writing of the query's response in a way that an experienced researcher would like to.\\
- binary: All questions should be yes-no binary questions.\\
- general: The question should not mention a specific paper's name (e.g. Author and year), nor narrow examples or demonstrations.\\
- grounded: These questions should be strictly based on the referral response.\\
\\
Here are the criteria that the list of questions needs to satisfy:\\
- sufficient: There should be enough important questions to cover a large space of possible contexts for the query.\\
- coverage: The questions in combinations should cover every important aspect in the referral response.\\
\\
Specifically, here, you are required to generate the question of depth type (e.g., Does the response explore area C1 and C2 in detail so that to better support the claim C?).\\
It is to ask whether information exists in the response that is explored in depth, such as detailed extention or explanation. Generally it means the response takes a longer-than-average length to explore a topic.\\
It is not to ask an information question (merely whether a statement, example, finding, opinion etc. exists), instead it requires depth in the contents.\\
You should first find the depth information that was actually included in the given referral response, and then ask the corresponding questions.\\
\\
Generate up to 5 questions (but no need to) and they should all meet the above criteria.\\
You should generate questions that are important and useful.\\
\\
Query: \\
\\
What are the latest works on finetuning an auto-regressive LM for dense passage retrieval and how are their performance compared with bi-directional encoders?\\
\\
Response:\\
It was traditionally assumed that decoder models would underperform compared to bi-directional autoregressive models, as referenced in {[9]}. However, recent advancements have challenged this notion, demonstrating that finetuned auto-regressive language models (LMs) can indeed surpass the capabilities of bi-directional encoder models in retrieval embedding tasks {[1]}{[2]}{[3]}{[4]}.\\
\\
One notable example is the E5-Mixtral model {[1]}, which employs a novel approach by generating a diverse set of synthetic data to finetune Mistral, a pre-trained large language model (LLM). This model has shown superior performance over the state-of-the-art (SOTA) bi-directional embedding models of that time, including OpenAI's text-embedding-3-large, Cohere-embed-english-v3.0, and voyage-lite-01-instruct on the MTEB benchmark {[8]}.\\
\\
Another study LLM2Vec {[2]} demonstrates the adaptation of a decoder model solely using public data for embedding tasks, which also achieved remarkable results on the MTEB benchmark {[8]}, surpassing other encoder models by a significant margin. This highlights the potential of decoder models when optimized appropriately.\\
\\
Furthermore, the GRIT model {[3]} unifies auto-regressive and bi-directional model designs. It incorporates two distinct language modeling heads atop the Transformer layers: one auto-regressive head, designed for generative tasks with a causal attention mask, and another bi-directional head, tailored for embedding tasks. This dual-head approach allows GRIT to leverage the strengths of both modeling techniques.\\
\\
Lastly, the NV-Embed model {[4]} adapts the decoder model architecture to enhance finetuning effectiveness using public datasets. This model not only ranks at the top on the MTEB benchmark {[8]} but also outperforms all existing baselines on the information retrieval benchmark {[5]}, underscoring the robustness and versatility of decoder models in handling complex language tasks.\\
\\
In short, E5-Mixtral {[1]}, LLM2Vec {[2]}, GRIT {[3]}, and NV-Embed {[4]} are good representations of the SOTA auto-regressive LMs on retrieval tasks whose performance is superior than traditional bi-directional encoders {[5]}{[6]}{[7]}{[8]}.\\
\\
{[1]} Improving text embeddings with large language models. (2023)\\
{[2]} LLM2Vec: Large Language Models Are Secretly Powerful Text Encoders (2024)\\
{[3]} Generative Representational Instruction Tuning (2024)\\
{[4]} NV-Embed: Improved Techniques for Training LLMs as Generalist Embedding Models (2024)\\
{[5]} NV-Embed: Improved Techniques for Training LLMs as Generalist Embedding Models (2024)\\
{[6]} Improving text embeddings with large language models. (2023)\\
{[7]} Generative Representational Instruction Tuning (2024)\\
{[8]} MTEB: Massive Text Embedding Benchmark (2022)\\
{[9]} NV-Embed: Improved Techniques for Training LLMs as Generalist Embedding Models (2024)\\
Questions:\\
1. Does the response compare auto-regressive model and bi-directional model on the dense passage retrieval in detail?\\
2. Does the response elaborate on the hybrid approach of GRIT in unifying auto-regressive and bi-directional features?\\
\\
Query-Response Pair:\\
Query:\\
\prompttag{\textbf{\color{insert}<QUERY>}}\\
Response:\\
\prompttag{\textbf{\color{insert}<REF\_ANSWER>}}\\
Questions:
}}%
\prompt
{Survey rubric generation prompt: citation-based-item}{
\systemprompt{
You will be shown a query asked by a junior PhD researcher and a referral response to this query.\\
Imagine that you are required to write another good response to this open-ended query,\\
you need to ask some follow-up questions to know the preferred content in the response or resolve ambiguity in the query.\\
The referral response is written by experienced research, and is perfect for generating these follow-up questions to the query.\\
\\
Here are criteria that individual questions need to satisfy:\\
- relevant: The question should ask about information that is related to the query, and would lower the answer variability of the query's response.\\
- salient: The question should address important piece to guide the writing of query's response in a way that an experienced research would like to.\\
- binary: All questions should be yes-no binary questions.\\
- qualitative: The question should be qualitative and focus on the big picture and important aspects, but not the nonessentials or specific numbers.\\
- grounded: These questions should be strictly based on the referral response.\\
\\
Here are the criteria that the list of questions needs to satisfy:\\
- sufficient: There should be enough important questions to cover a large space of possible contexts for the query.\\
- coverage: The questions in combinations should cover every important aspects in the referral response.\\
\\
Specifically, here, you are required to generate the question of citation type.\\
For instance, Does the response cite the paper by {author} (year) (title: {title}) that back the statement A by explaining the detail A1 and A2?\\
It is to ask whether an important citation exist in the response, and how it is used in the response. \\
It must mention the title of the paper in a way ``(title: {title})'' as long as the title is provided, and the details of the statement that the citation is used to back up.\\
You should first find the citation(s) that were actually included in the given referral response, and then ask the corresponding questions.\\
\\
Generate up to 5 questions (but no need to) and they should all meet the above criteria.\\
You should generate questions that are important and useful, and address a *necessary* citation perspective of the query's response.\\
Please make sure that your questions are relevant, salient, binary, qualitative, and grounded. \\
Please list the questions from the most to the least necessary to be in the query's response.\\
\\
Query-Response Pair:\\
Query:\\
\\
What are the latest works on finetuning an auto-regressive LM for dense passage retrieval and how are their performance compared with bi-directional encoders?\\
\\
Response: \\
\\
It was traditionally assumed that decoder models would underperform compared to bi-directional autoregressive models, as referenced in {[9]}. However, recent advancements have challenged this notion, demonstrating that finetuned auto-regressive language models (LMs) can indeed surpass the capabilities of bi-directional encoder models in retrieval embedding tasks {[1]}{[2]}{[3]}{[4]}.\\
\\
One notable example is the E5-Mixtral model {[1]}, which employs a novel approach by generating a diverse set of synthetic data to finetune Mistral, a pre-trained large language model (LLM). This model has shown superior performance over the state-of-the-art (SOTA) bi-directional embedding models of that time, including OpenAI's text-embedding-3-large, Cohere-embed-english-v3.0, and voyage-lite-01-instruct on the MTEB benchmark {[8]}.\\
\\
Another study LLM2Vec {[2]} demonstrates the adaptation of a decoder model solely using public data for embedding tasks, which also achieved remarkable results on the MTEB benchmark {[8]}, surpassing other encoder models by a significant margin. This highlights the potential of decoder models when optimized appropriately.\\
\\
Furthermore, the GRIT model {[3]} unifies auto-regressive and bi-directional model designs. It incorporates two distinct language modeling heads atop the Transformer layers: one auto-regressive head, designed for generative tasks with a causal attention mask, and another bi-directional head, tailored for embedding tasks. This dual-head approach allows GRIT to leverage the strengths of both modeling techniques.\\
\\
Lastly, the NV-Embed model {[4]} adapts the decoder model architecture to enhance finetuning effectiveness using public datasets. This model not only ranks at the top on the MTEB benchmark {[8]} but also outperforms all existing baselines on the information retrieval benchmark {[5]}, underscoring the robustness and versatility of decoder models in handling complex language tasks.\\
\\
In short, E5-Mixtral {[1]}, LLM2Vec {[2]}, GRIT {[3]}, and NV-Embed {[4]} are good representations of the SOTA auto-regressive LMs on retrieval tasks whose performance is superior than traditional bi-directional encoders {[5]}{[6]}{[7]}{[8]}.\\
\\
{[1]} Improving text embeddings with large language models. Liang Wang, et al. (2023)\\
{[2]} LLM2Vec: Large Language Models Are Secretly Powerful Text Encoders. Parishad BehnamGhader, et al. (2024)\\
{[3]} Generative Representational Instruction Tuning. Niklas Muennighoff, et al. (2024)\\
{[4]} NV-Embed: Improved Techniques for Training LLMs as Generalist Embedding Models. Chankyu Lee, et al. (2024)\\
{[5]} NV-Embed: Improved Techniques for Training LLMs as Generalist Embedding Models. Chankyu Lee, et al. (2024)\\
{[6]} Improving text embeddings with large language models. Liang Wang, et al. (2023)\\
{[7]} Generative Representational Instruction Tuning. Niklas Muennighoff, et al. (2024)\\
{[8]} MTEB: Massive Text Embedding Benchmark. Niklas Muennighoff, et al. (2022)\\
{[9]} NV-Embed: Improved Techniques for Training LLMs as Generalist Embedding Models. Chankyu Lee, et al. (2024)\\
Questions:\\
1. Does the response cite some papers such as Wang et al. (2023) (title: Improving text embeddings with large language models) and the MTEB paper (title: Massive Text Embedding Benchmark) that show the performance of auto-regressive LMs surpassing bi-directional encoders in retrieval tasks?\\
2. Does the response cite the GRIT model paper (title: Generative Representational Instruction Tuning) that shows how to leverage the strengths of both auto-regressive and bi-directional model by incorporating two distinct language modeling heads atop the Transformer layers?\\
\\
Query-Response Pair:\\
Query:\\
\prompttag{\textbf{\color{insert}<QUERY>}}\\
Response:\\
\prompttag{\textbf{\color{insert}<REF\_ANSWER>}}\\
Questions:
}}%
\prompt
{Rubric deduplication prompt}{
\systemprompt{
The rubric items below intend to evaluate answers to the following query:
\prompttag{\textbf{\color{insert}<QUERY>}} (Date Cutoff: \prompttag{\textbf{\color{insert}<DATE>}}).

Rubric: 
\prompttag{\textbf{\color{insert}<CURRENT\_RUBRIC>}}

The user wants to add to the rubric, but only if the new rubric item is not a rephrasing or restatement of an existing rubric item. In other words, no two rubric items should not repeat another. Answer only `new item' if the user should add the item to the rubric, and answer only `repeated item' if the item is already represented in the rubric. Do not elaborate.
}
\vspace{0.5em}
\userprompt{
Here is the proposed new rubric item:
\prompttag{\textbf{\color{insert}<NEW\_RUBRIC\_ITEM>}}
}}
\prompt
{Hybrid rubric reranker prompt}{
\systemprompt{
Select the \prompttag{\textbf{\color{insert}<RUBRIC\_SIZE>}} most important questions that should be addressed in order to answer the following query:\\
\prompttag{\textbf{\color{insert}<QUERY>}} (Date Cutoff: \prompttag{\textbf{\color{insert}<DATE>}}).\\
\\
Format: List the questions in a 0-indexed python list, e.g., [0, 3, ...].
}
\vspace{0.5em}
\userprompt{
Here are the following questions to choose from:
\prompttag{\textbf{\color{insert}<SURVEY\_RUBRIC>}}
\prompttag{\textbf{\color{insert}<PARAMETRIC\_RUBRIC>}}
}}
\end{document}